\documentclass{article} 
\usepackage{iclr2016_conference,times}
\usepackage{hyperref}
\usepackage{url}

\usepackage{epsfig}
\usepackage{graphicx,dblfloatfix}
\usepackage{amsmath}
\usepackage{amssymb}

\usepackage{etex}
\usepackage{subfig}
\usepackage{amsfonts}
\usepackage{amsmath}
\usepackage{tikz}
\usetikzlibrary{shapes,arrows,positioning,calc,matrix,3d}
\usepackage{booktabs}

\usepackage{rotating}
\usepackage{capt-of}

\def\eg{\emph{e.g.,}} 
\def\ie{\emph{i.e.,\;}} 
\def\cf{\emph{c.f.}}

\def\Mat#1{{\bf #1}}
\def\Vec#1{{\bf #1}}

\newcommand{\fig}{Fig.}

\newtheorem{hypothesis}{Hypothesis}

\makeatletter
\newcommand{\gettikzxy}[3]{%
	\tikz@scan@one@point\pgfutil@firstofone#1\relax
	\edef#2{\the\pgf@x}%
	\edef#3{\the\pgf@y}%
}
\makeatother

\definecolor{colorSepia}{RGB}{103,24,0}
\definecolor{colorPurple}{RGB}{128,0,255}

\title{Foveation-based Mechanisms \\ Alleviate  Adversarial Examples}

\author{Yan Luo$^1$ \;\; \;\;  Xavier Boix$^{1,2}$ \;\;\;\;  Gemma Roig$^2$  \;\;\;\; Tomaso Poggio$^2$ \;\;\;\; Qi Zhao$^1$ \\
$^1$ Department of Electrical and Computer Engineering, National University of Singapore, Singapore\\
$^2$ CBMM, Massachusetts Institute of Technology \& Istituto Italiano di Tecnologia, Cambridge, MA  \\
}

\begin{document}

\maketitle

\graphicspath{ {./fig/} }

\begin{abstract}
  
  We show that adversarial examples,~\ie the visually imperceptible perturbations that result in Convolutional Neural Networks (CNNs) fail, can be alleviated with a mechanism based on foveations---applying the CNN in different image regions. To see this, first, we report results in ImageNet that lead to a revision of the hypothesis that adversarial perturbations are a consequence of CNNs acting as a linear classifier: CNNs act locally linearly to changes in the image regions with objects recognized by the CNN, and in other regions the CNN may act non-linearly. Then, we corroborate that when the neural responses are linear, applying the foveation mechanism to the adversarial example tends to significantly reduce the effect of the perturbation. This is because, hypothetically, the CNNs for ImageNet are robust to changes of scale and translation of the object produced by the foveation, but this property does not generalize to transformations of the perturbation. As a result, the accuracy after a foveation is almost the same as the accuracy of the CNN without the adversarial perturbation, even if the adversarial perturbation is calculated taking into account a foveation.
  
\end{abstract}

\section{Introduction}

The generalization properties of CNNs have been recently questioned. \cite{SzegedyZSBEGF_CoRR_2013} showed that there are perturbations that when added to an image make CNNs fail to predict the object category in the image. Surprisingly, these perturbations are usually so small that they are imperceptible to humans. The perturbed images are the so-called \emph{adversarial examples}.

Adversarial examples exist because, hypothetically, CNNs act as a linear classifier of the image~\citep{GoodfellowSS_CoRR_2014}. The perturbation may be aligned with the linear classifier to counteract the correct prediction of the CNN, and since the dimensionality of the classifier is high, as it is equal to the number of pixels in the image, the perturbation can be spread among the many dimensions and make the perturbation of each pixel small,~\cf~\citep{GoodfellowSS_CoRR_2014,fawzi15}. This hypothesis of CNNs acting too linearly is supported by quantitative experiments in datasets where the target object is always centered and fixed to the same scale and orientation, namely, MNIST~\citep{Lecun_PIEEE_1998} and CIFAR~\citep{Krizhevsky2009cifar}.

In this paper, we challenge the current hypothesis by analyzing adversarial examples using several CNN architectures for ImageNet~\citep{Krizhevsky_NIPS_2012,Simonyan_CoRR_2014,Szegedy_CoRR_2014}, which do not assume a positioning and scale of the object as in previous works. Our experiments show that the CNN does not act as a linear classifier, but the CNN acts locally linearly to changes in the image regions where there is an object recognized by the CNN, and in other regions the CNN may act non-linearly (Hypothesis~\ref{hyp:linearity}). Also, the experiments show that the adversarial perturbations that are generated with a CNN   different from the CNN used for evaluation, produce a much smaller decrease of the accuracy compared to the CNN architectures used for the datasets in previous works~\citep{SzegedyZSBEGF_CoRR_2013,GoodfellowSS_CoRR_2014}.

This leads to the main hypothesis of this paper, which can be used to show that foveation mechanisms can substantially alleviate adversarial examples. A foveation mechanism  selects a region of the image to apply the CNN, discarding  information from the other regions. From the aforementioned local linearity hypothesis, it follows that the neural responses of a foveated object in the adversarial example can be decomposed into two independent terms, one that comes from the foveated clean object without perturbation, and another from the foveated perturbation. The hypothesis---supported by a series of experiments---says that the term of the classification score of the clean image without perturbation is not negatively affected by a foveation that contains the target object in the image, as CNNs for ImageNet are trained to be robust to scale and position transformations of the object. Yet, this property does not generalize to transformations of the perturbation, and the effect of perturbation on the classification score after the foveation is lower than before (Hypothesis~\ref{hyp:foveation}).

In a series of experiments under different conditions of clutter and using different kinds of foveations, we show that the foveation mechanisms on adversarial examples almost recover the accuracy performance of the CNN without the adversarial perturbation. This is also in the case of adversarial perturbations calculated taking into account a foveation, as we can always use a different foveation to exploit the Hypothesis~\ref{hyp:foveation}.

\input{tex/example}

\section{Experimental Set-up}
\label{noise}

In this section we introduce the experimental set-up that we use in the rest of the paper. 

{\bf Dataset \;} We use ImageNet (ILSVRC)~\citep{Deng_CVPR_2009}, which is a large-scale dataset for object classification and detection. We report our results on the validation set of ILSVRC 2012 dataset, which contains $50'000$ images. Each image is labeled with one of the $1'000$ possible object categories, and the ground truth bounding box around the object is also provided.  We evaluate the accuracy performance with the standard protocol~\citep{Russakovsky_CoRR_2014}, taking the top $5$ prediction error.

{\bf CNN Architectures \;} We report results using $3$ CNN which are representative of the state-of-the-art, namely AlexNet~\citep{Krizhevsky_NIPS_2012},   GoogLeNet~\citep{Szegedy_CoRR_2014} and VGG~\citep{Simonyan_CoRR_2014}. 
We use the publicly available pre-trained models provided by the Caffe library~\citep{Jia_arXiv_2014}. We use the whole image as input to the CNN, resized to~$227\times 227$ pixels in AlexNet, and $224\times 224$ in VGG and GoogLeNet. In order to improve the accuracy at test time, a commonly used strategy in the literature is to use multiple crops of the input image and then combine their classification scores. We do not use this strategy unless we explicitly mention it.

{\bf Generation of Adversarial Perturbations \;} The adversarial  perturbation consist on finding the perturbation with minimum norm that produces misclassification, under the top-$5$ criterion. This is an NP-hard problem, and the algorithms to compute the minimum perturbation are approximate.
To generate the adversarial examples we use the two algorithms available in the literature. Namely, the algorithm by \cite{SzegedyZSBEGF_CoRR_2013}, that is based on an optimization with L-BFGS, and we also use the algorithm by~\cite{GoodfellowSS_CoRR_2014}, that uses the sign of the gradient of the loss function. We call them \emph{BFGS} and \emph{Sign}, respectively. For more details about the generation of the perturbation we refer to Sec.~\ref{app:NoiseGeneration}, and to \citep{SzegedyZSBEGF_CoRR_2013,GoodfellowSS_CoRR_2014}.

Several qualitative examples of the perturbations are shown in Fig.~\ref{fig:examples}. Observe that for \emph{BFGS}, the perturbation is concentrated on the position of the object, while  for \emph{Sign}, the perturbation is spread through the image. Also, note that for each CNN architecture  the perturbation is different.

{\bf When are Adversarial Perturbations Perceptible? \;}
The perceptibility of adversarial examples is subjective. We show examples of the same adversarial perturbation varying the $L_1$ norm per pixel, Fig.~\ref{fig:l1example},~\ref{fig:l1example2},~\ref{fig:l1example3} in Sec.~\ref{app:Figures}, and  the $L_\infty$ norm, Fig.~\ref{fig:linfexample},~\ref{fig:linfexample2},~\ref{fig:linfexample3} in Sec.~\ref{app:Figures}. We can see that after the norm of the perturbation reaches a certain value, the perturbation becomes perceptible, and it starts occluding  parts of the target object. We give an estimate of the minimum value of the norm that makes the adversarial perturbations perceptible by visually analyzing few hundreds of adversarial examples. For \emph{BFGS},  we can see that when the $L_1$ norm per pixel is higher than $15$ the perturbation has become slightly visible in almost all cases, and for $L_{\infty}$ this value is $100$. This difference between the two values is because the density of \emph{BFGS} is not the same through all image, as commented before. For \emph{Sign}, the threshold for both norms to make the perturbation visible is about $15$ (here the value coincides for both because this perturbation is spread evenly through all the image). Note that this values are a rough estimation from qualitative observations and highly depend on the image and CNN used.

{\bf  Positioning of the Target Object \;} We use the positioning of the target object in the image as a tool  to take into account the perturbation's position in the image. We use the ground truth bounding box around the object provided in the ILSVRC 2012 dataset, which is manually annotated by a human.

\section{Review of the Linearity of CNNs for ImageNet}
\label{sec:exp}

We now introduce the analysis of adversarial examples in CNNs for ImageNet. In this dataset, objects can be at different scales and positions, and as we will see, this difference with previous works leads to a revision of the hypotheses introduced in these works.

Let $\Mat{x}$ be an image that contains an object whose object category is of class label $\ell$, and we denote as  $f(\Mat{x})$ the mapping of the CNN from the input image to the classification scores. We use $\Mat{\epsilon}$ to denote a perturbation that produces a misclassification when added to the input image,~\ie~$\ell \not\in C(f(\Mat{x}+\Mat{\epsilon}))$ in which $C(\cdot)$ maps the classification scores to class labels. The set of all perturbations that produce a misclassification is denoted as $ \mathcal{E}_{\Mat{x}}  = \left\{ \Mat{\epsilon}  |\;\ell \not\in C(f(\Mat{x}+\Mat{\epsilon}))  \right\}$. Also, we define $\Mat{\epsilon}^\star$  as the perturbation in $\mathcal{E}_{\Mat{x}}$ with minimal norm,~\ie~$\Mat{\epsilon}^\star = \arg\min_{\Mat{\epsilon}\in \mathcal{E}_{\Mat{x}}} \|\Mat{\epsilon}\|$, which may be imperceptible.

The causes of adversarial examples have been studied by \cite{GoodfellowSS_CoRR_2014,fawzi15}, that showed that small perturbations can affect linear classifiers, and CNNs may act as a high-dimensional linear classifier (of the same dimensionality as the image size). The learned parameters of the CNN make that the CNN acts as a linear classifier, bypassing the non-linearities in the architecture. In this way, $f(\Mat{x}+\Mat{\epsilon}^\star)$ can be rewritten as $\Mat{w}^\prime\Mat{x}+\Mat{w}^\prime\Mat{\epsilon}^\star$, where $\Mat{w}^\prime$ is the transpose of a high-dimensional linear classifier, and it yields a classifier approximately equivalent to $f(\cdot)$. Observe that the high-dimensionality of the classifier allows that even when $\Mat{w}^\prime\Mat{\epsilon}^\star$ is a high value, the average per pixel value of $\Mat{\epsilon}^\star$ may be low, since the perturbation can be spread among all the pixels, and make it imperceptible.

A direct consequence of the linearity of the CNN is that multiplying the adversarial perturbation by a constant value larger than $1$ also produces misclassification, since $c\Mat{w}^\prime\Mat{\epsilon}^\star$ is a factor $c$ times  larger than before. Thus, the set of adversarial examples, $\mathcal{E}_{\Mat{x}}$, is not only $\Mat{\epsilon}^\star$, but rather a dense set of perturbations around $\Mat{\epsilon}^\star$. Another phenomenon that can be explained through the linearity of the CNN is that adversarial examples generated for a specific CNN model produce misclassification in other CNNs~\citep{SzegedyZSBEGF_CoRR_2013}. This may be because the different CNNs act as a linear classifier, that are very similar among them.

In the following, we review the linearity of CNNs in ImageNet. Then, we analyze the influence of the position of the perturbation in the image, which leads to a new hypothesis.


{\bf Review of the Properties of Adversarial Examples \;}
We now revisit the aforementioned properties of the set of adversarial examples, $\mathcal{E}_{\Mat{x}}$, that can be derived from CNNs acting too linearly. To analyze that multiplying adversarial perturbations by a factor also leads to misclassification, we  measure the classification accuracy when we vary the norm of the perturbation. We use the $L_1$ norm per pixel and also the $L_{\infty}$ norm, as \emph{BFGS} and \emph{Sign} perturbations are optimized for each of these norms respectively. At the same time, we also analyze the effect of adversarial perturbations generated on a CNN architecture different from the CNN we evaluate.

\begin{figure*}[t!]
	\centering
	\vspace{-2ex}
	\raisebox{0\height}{\subfloat{\includegraphics[width=0.8\textwidth]{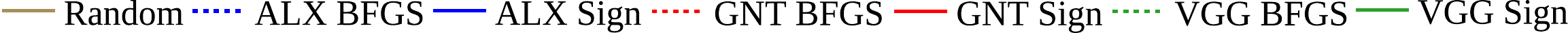}}}
	\vspace{-2ex}
	{\renewcommand{\arraystretch}{0.1}%
		\setlength{\tabcolsep}{2pt}
		\begin{tabular}{ccc}
			\subfloat{\includegraphics[width=.305\textwidth]{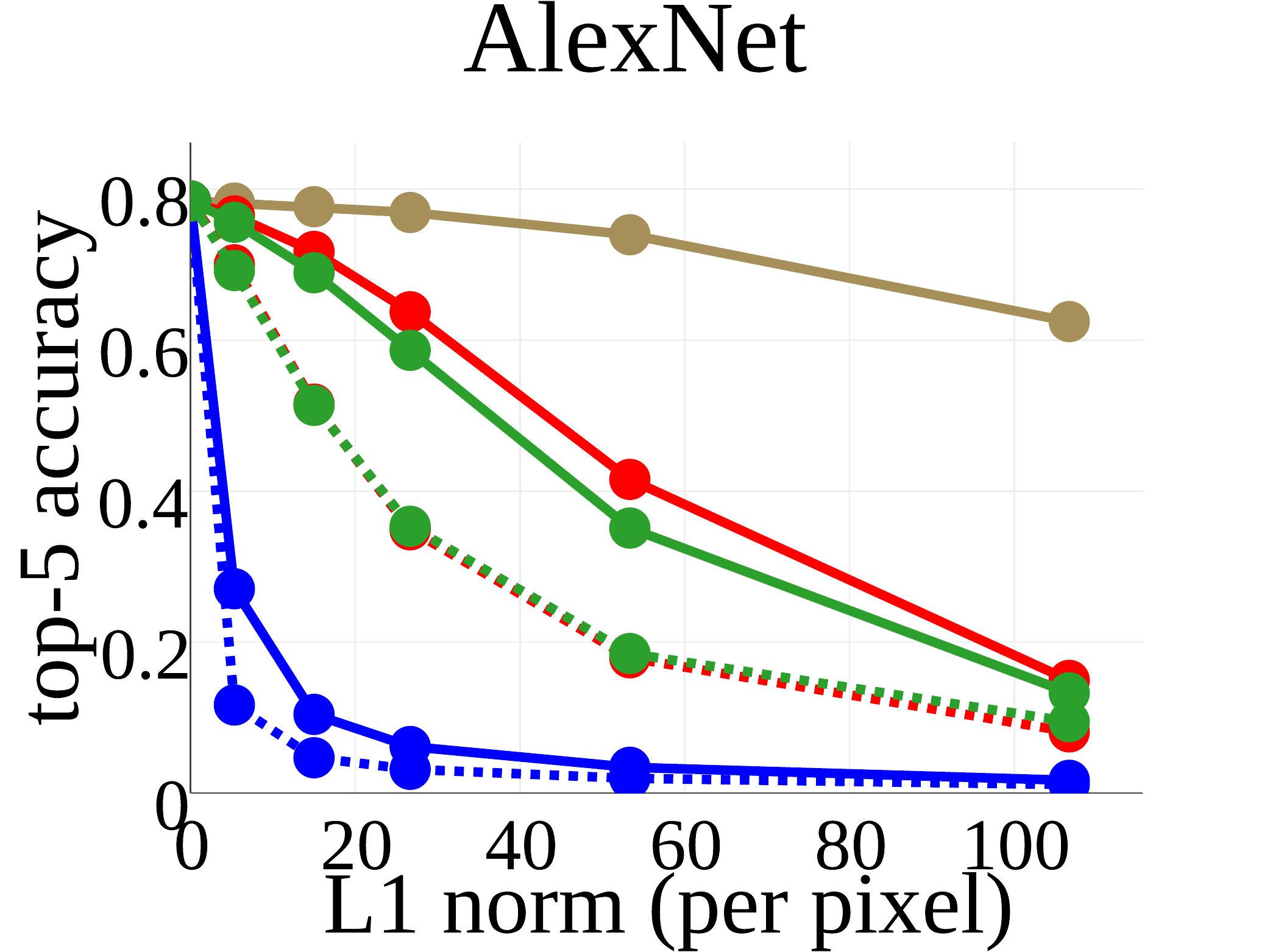}} &
			\subfloat{\includegraphics[width=.305\textwidth]{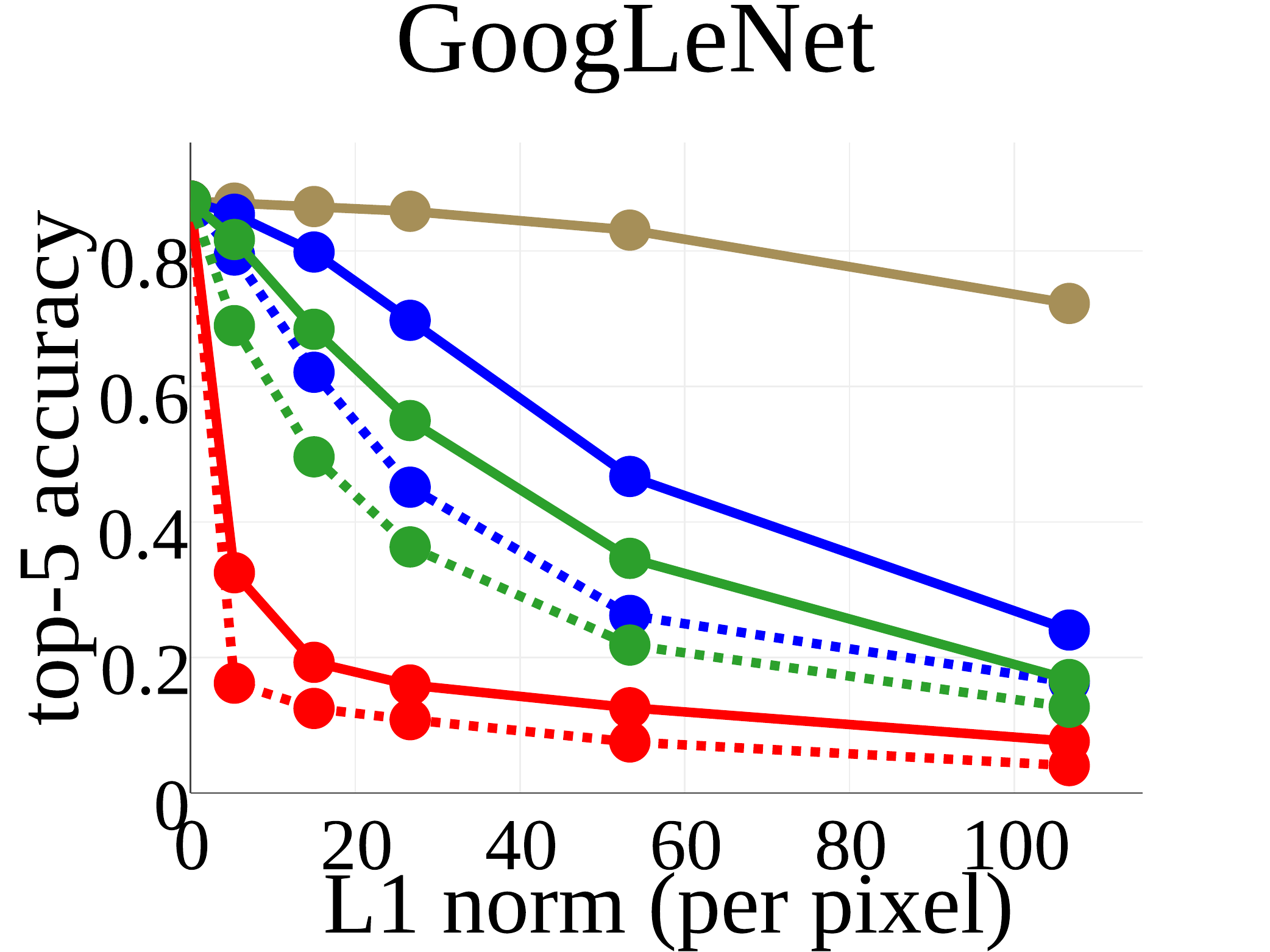}} &
			\subfloat{\includegraphics[width=.305\textwidth]{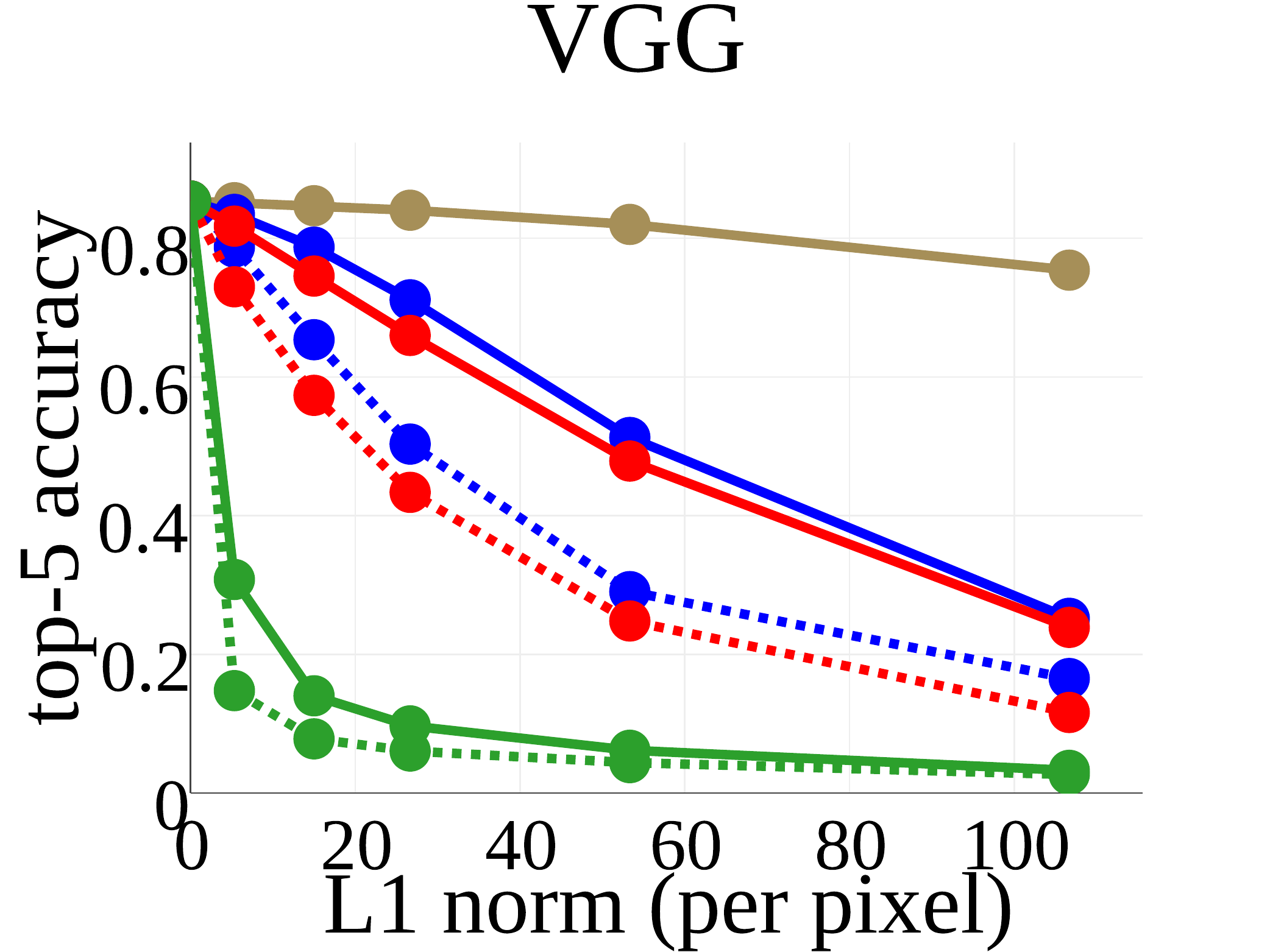}} \vspace{1ex}\\
		\end{tabular}
	}
	\caption
	{
		\textit{Accuracy when Varying the $L_1$ Norm per Pixel of the Perturbation.} Accuracy for the three CNNs we evaluate. We denote the perturbation as $X$ $Y$, where $X$ is the network that generated the perturbation --- ALX (AlexNet), GNT (GoogLeNet), VGG --- and $Y$ indicates the \emph{BFGS perturbation} or \emph{Sign}. See Fig.~\ref{fig:cross2} in~\ref{app:Results} for the same results using $L_{\infty}$.
	}
	\label{fig:cross}
\end{figure*}

Results can be seen in Fig.~\ref{fig:cross} for $L_1$ norm per pixel. We additionally provide the accuracy for a perturbation generated by adding to each pixel a random value uniformly distributed in the range $[-1,1]$. We observe a clear tendency that when we increase the norm of the perturbation, it results in more  misclassification on average, which is in accordance to the hypothesis that CNNs are linear~\citep{GoodfellowSS_CoRR_2014}. In Fig.~\ref{fig:cross2} in Sec.~\ref{app:Results}, the same conclusions can be extracted for $L_{\infty}$. Also, we observe that for the $L_1$ norm per pixel,  \emph{BFGS} produces more  misclassification on average than \emph{Sign}, and the opposite is true if we evaluate with $L_{\infty}$. This is not surprising as the \emph{BFGS} optimizes the  $L_1$ norm of the perturbation, and the \emph{Sign} optimizes the maximum. Thus, in the rest of the experiments of the paper we only evaluate \emph{BFGS} with the $L_1$ norm per pixel, and \emph{Sign} with the $L_{\infty}$ norm.

Interestingly, we observe that there is not an imperceptible adversarial perturbation for every image. Recall that an $L_1$ norm per pixel roughly higher than  $15$, and $L_{\infty}$ also roughly higher than $15$ produce a perturbation slightly perceptible (Sec.~\ref{noise}). Thus, the results in Fig.~\ref{fig:cross} show that the accuracy is between $10\%$ and $20\%$, depending on the CNN, when the adversarial perturbations are constrained to be no more than slightly perceptible.

Finally, we see that for a particular CNN architecture, the perturbation that results in more misclassification  with lower norm is in all cases the one that is generated using the same CNN architecture. Observe that the perturbation generated for a CNN architecture, does not affect so severely the other CNN architectures as reported in MNIST,~\cf~\citep{SzegedyZSBEGF_CoRR_2013}. When the adversarial perturbations are constrained to be no more than a bit perceptible, the accuracy is between $40\%$ and $60\%$ in the worst case, depending on the CNN. This may be because the differences among the learned CNNs in ImageNet are bigger than the differences among  the CNNs learned in MNIST.

{\bf Role of the Target Object Position in the Perturbation \;} 
We now analyze the effect of the perturbation at different regions of the image. We report the accuracy performance that yields the perturbation computed for the whole image when it is masked to be positioned only on the target object, or when it masked to be only on the background.
Specifically, we create an image mask using the ground truth bounding box (denoted as \emph{Object Masked} in the figures), where the mask's pixels are $1$ inside the bounding box, and $0$ otherwise. Before the perturbation  computed for the whole image is added to the image, we  multiply (pixel-wise) the perturbation by the mask, such that the resulting perturbation is only positioned inside the bounding box. To analyze the perturbation that is in the background, we also create  the inverted mask, denoted as \emph{Background Masked}, in which the pixels outside the bounding box are $1$, and $0$ for the pixels inside.

\begin{figure*}[t!]
	\centering
	\vspace{-3ex}
	{\renewcommand{\arraystretch}{0.1}
		\setlength{\tabcolsep}{2pt}
		\begin{tabular}{ccc}
			\multicolumn{3}{l}{\hspace{1ex}\scriptsize{legend (a):} \subfloat{\includegraphics[width=0.6\textwidth]{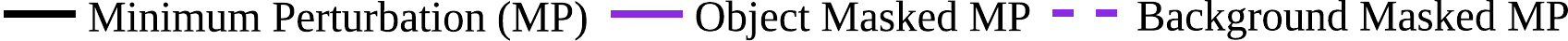}}} \vspace{-1.5ex}\\
			\multicolumn{3}{l}{\hspace{1ex}\scriptsize{legend (b):} \subfloat{\includegraphics[width=0.5\textwidth]{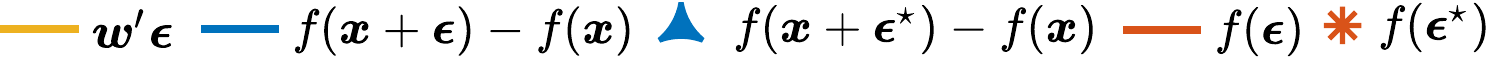}}{\hspace{10ex}\scriptsize{legend (c):} \subfloat{\includegraphics[width=0.15\textwidth]{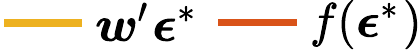}}}} \\
			
			\subfloat{\includegraphics[width=.305\textwidth]{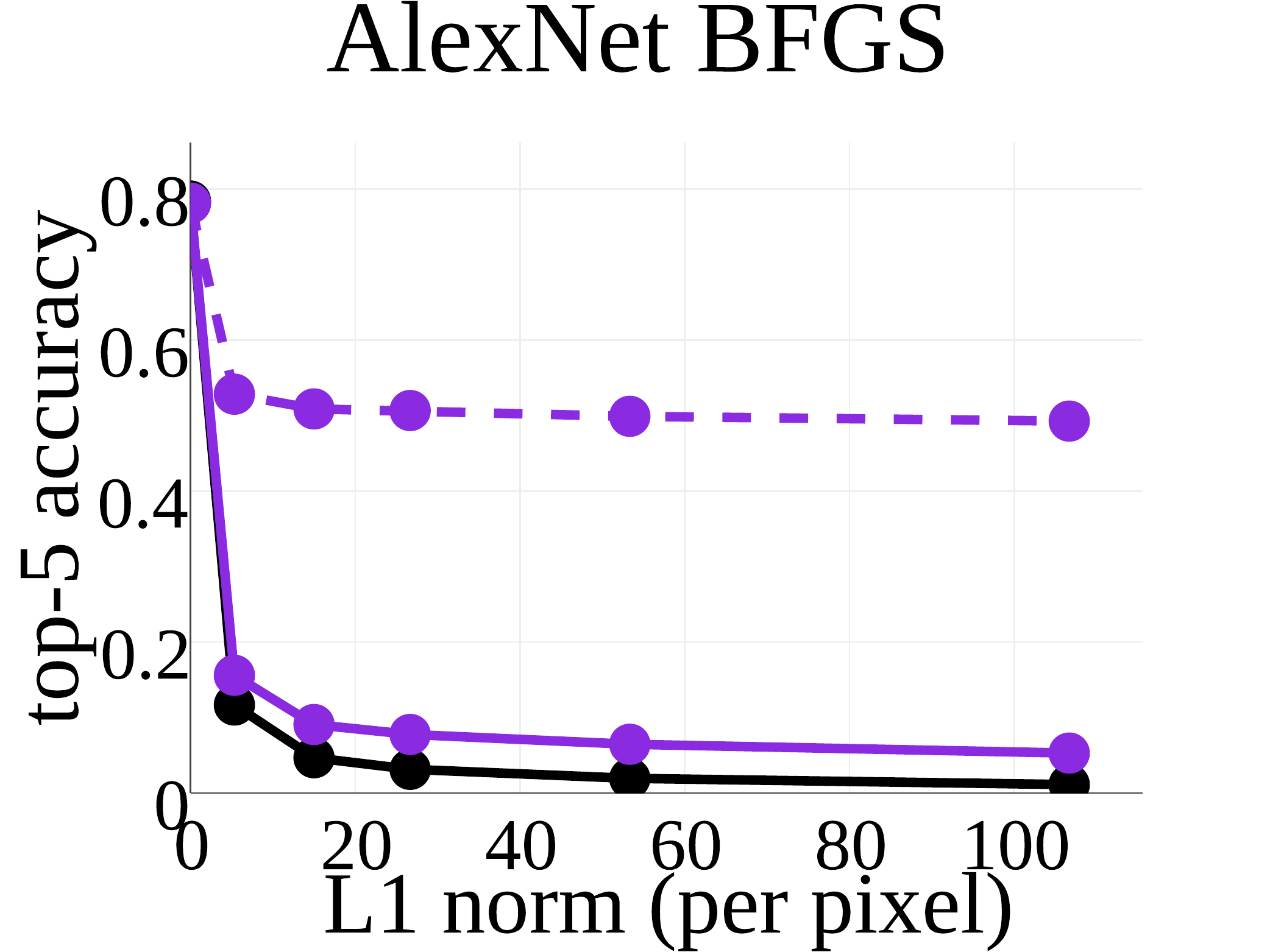}} &
			\subfloat{\includegraphics[width=.305\textwidth]{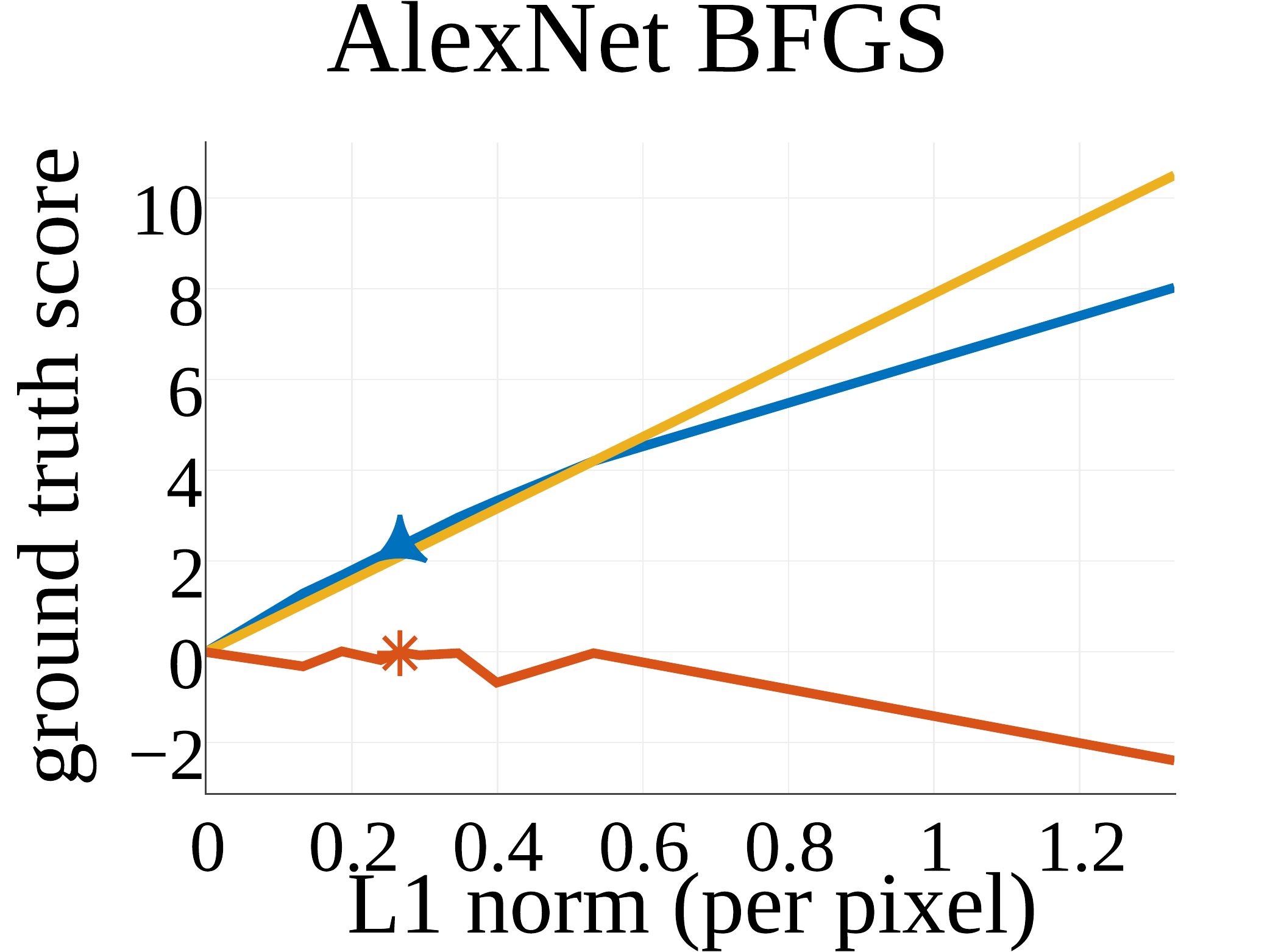}} &
			\subfloat{\includegraphics[width=.305\textwidth]{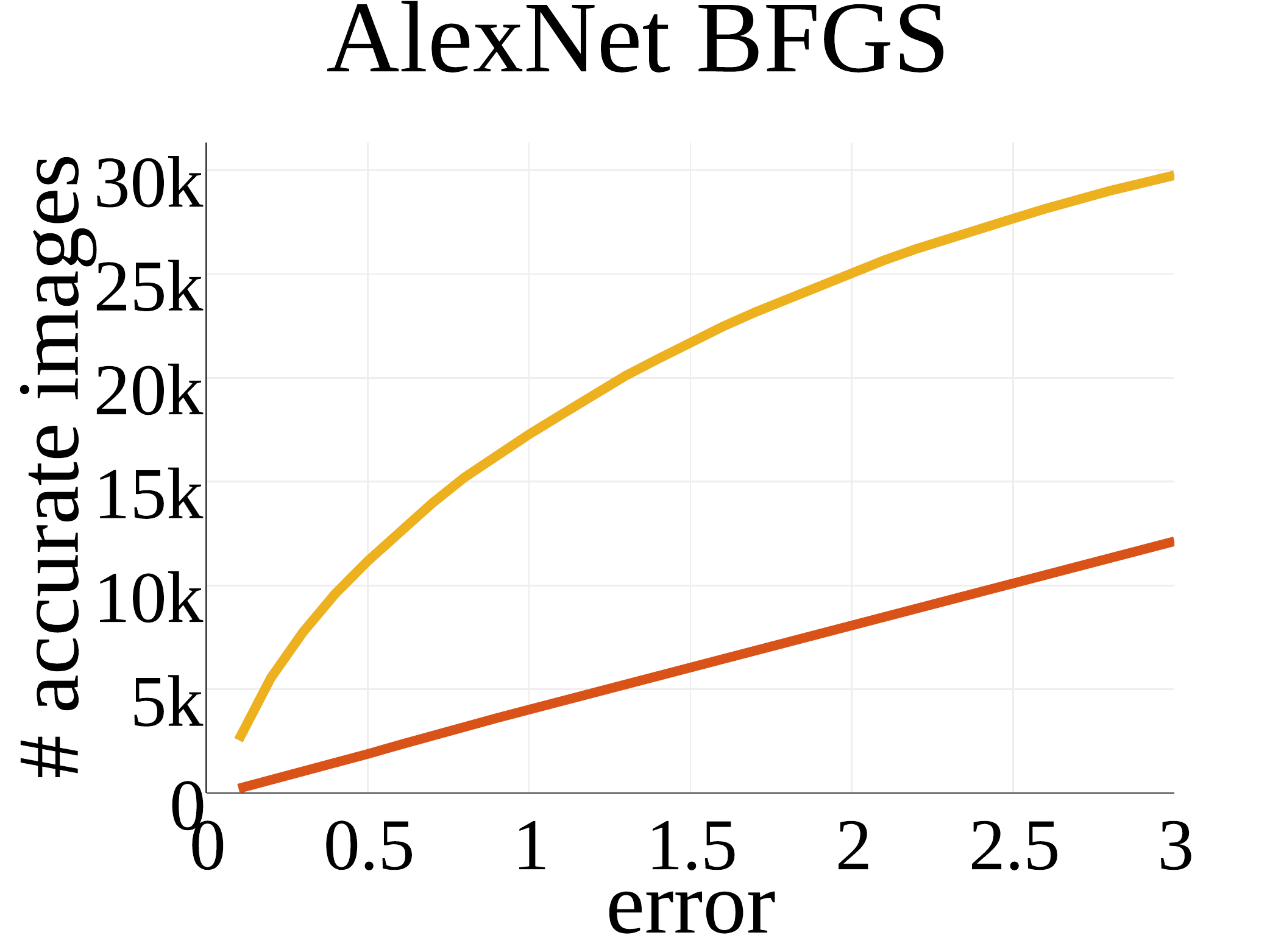}} \\
			{\small (a)}  &
			{\small (b)} &
			{\small (c)} \\
		\end{tabular}
	}
	\caption
	{\emph{Experiments for the Local Linearity Hypothesis.} 
		(a) Accuracy of the masked perturbations for the AlexNet CNN, when varying the norm of \emph{BFGS}. See Fig.~\ref{fig:removenoise2} in~\ref{app:Results} for the results with \emph{Sign} and $L_\infty$ norm for all CNNs we evaluate;
		(b) Classification score of the ground truth object category for the image in Fig.~\ref{fig:examples}, when varying the $L_1$ norm  per pixel of \emph{BFGS}. See Fig.~\ref{fig:linearexample18},~\ref{fig:linearexample75},~\ref{fig:linearexample174} in~\ref{app:Results} for more examples, and also \emph{Sign} with $L_\infty$ norm; (c) Cumulative histogram of the number of images with an $L_1$ error smaller than the value in the horizontal axis for \emph{BFGS} computed for AlexNet. Only the images that are correctly classified by the CNN are included. See Fig.\ref{fig:linearity2} in~\ref{app:Results} for the results of all CNNs.
	}
	\label{fig:removenoise}
\end{figure*}

In Fig.~\ref{fig:removenoise}a  we report the accuracy performance for \emph{BFGS} for \emph{Object Masked} and \emph{Background Masked}, and also  for the perturbation applied in the full image generated for the AlexNet CNN (\emph{minimum perturbation (MP)}). See Fig.~\ref{fig:removenoise2} in~\ref{app:Results} for the results for each CNN architecture and \emph{Sign} perturbation.
We observe that the accuracy performance for the perturbation positioned only on the object decreases  in a similar way as \emph{MP}. When adding the perturbation only on the background, the accuracy decreases significantly less than for \emph{Object Masked} (there is a difference between them of about $40\%$ of accuracy).
Note that this result is also clear for small values of the perturbation norm,~\ie the values that are imperceptible.

From the hypothesis that the CNN acts as a linear classifier, the result of this experiment suggests that the perturbation, $\Mat{\epsilon}^\star$, is aligned with the classifier, $\Mat{w}$, on the same position of the target object, otherwise there is not an alignment between them. However, if the CNN is linear, to produce misclassification, $\Mat{\epsilon}^\star$ could be aligned with  $\Mat{w}$ in any position, because $\Mat{w}^\prime\Mat{\epsilon}^\star$ only needs to surpass the value of $\Mat{w}^\prime\Mat{x}$, independently of the image, $\Mat{x}$.
This begs the question why there is such a clear relationship between  $\Mat{\epsilon}^\star$  and $\Mat{x}$.
An answer could be that for the algorithms that generate the adversarial perturbation (\emph{BFGS} and \emph{Sign}), it is easier to find perturbations on the position of the object than in other positions, and hypothetically, we could also find adversarial perturbations that are aligned with the classifier in positions different from the object. Yet, this needs to be verified. The next experiment clarifies this point.

\begin{figure*}[t!]
	\centering
	\vspace{-2ex}
	\raisebox{0\height}{\subfloat{\includegraphics[width=0.55\textwidth]{linearity_example/linearexample_legend}}}
	{\renewcommand{\arraystretch}{0.1}
		\setlength{\tabcolsep}{2pt}
		\begin{tabular}{ccc}
			\subfloat{\includegraphics[width=.305\textwidth]{linearity_example/alx_bfgs_18}} &
			\subfloat{\includegraphics[width=.305\textwidth]{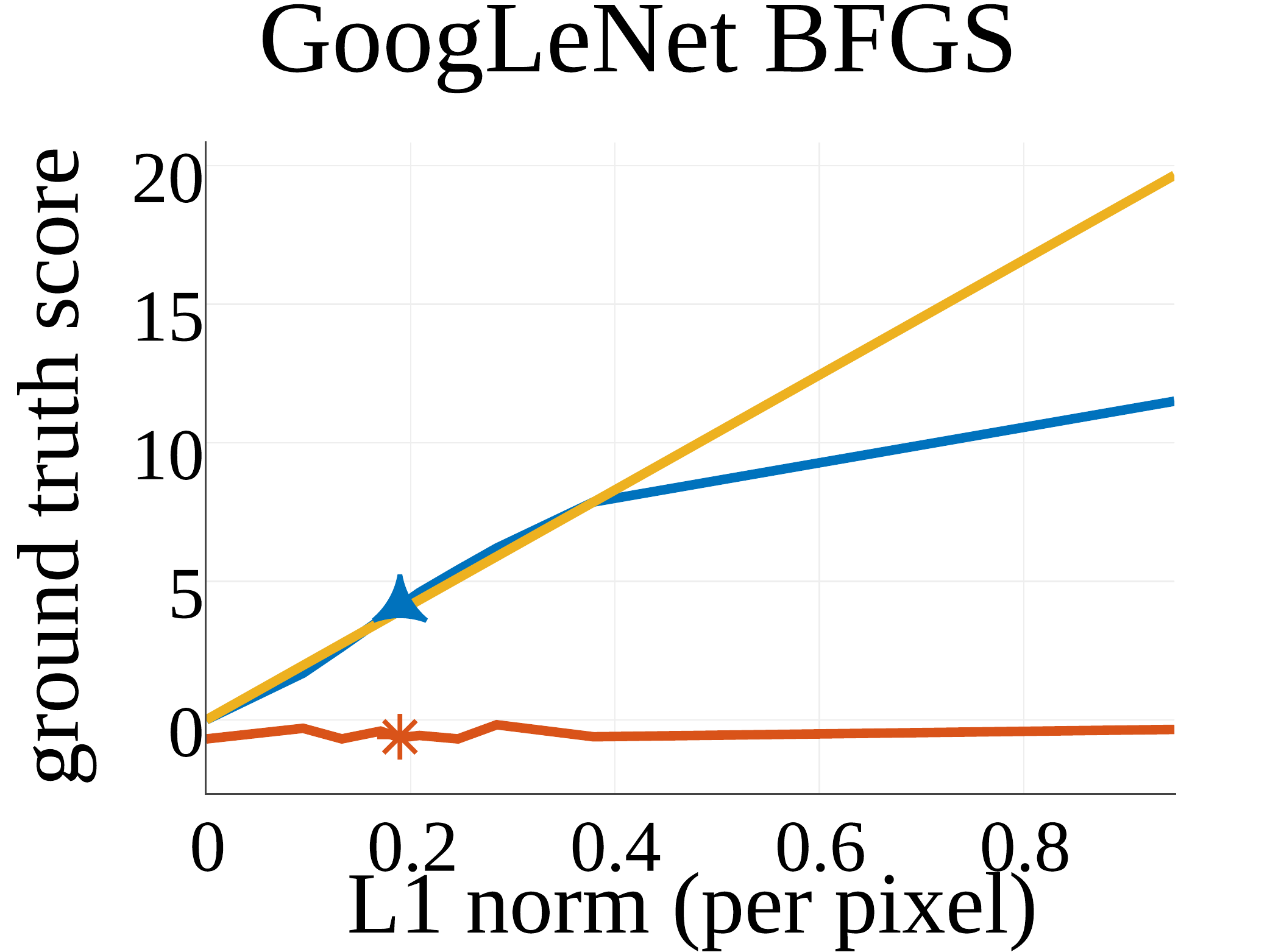}} &
			\subfloat{\includegraphics[width=.305\textwidth]{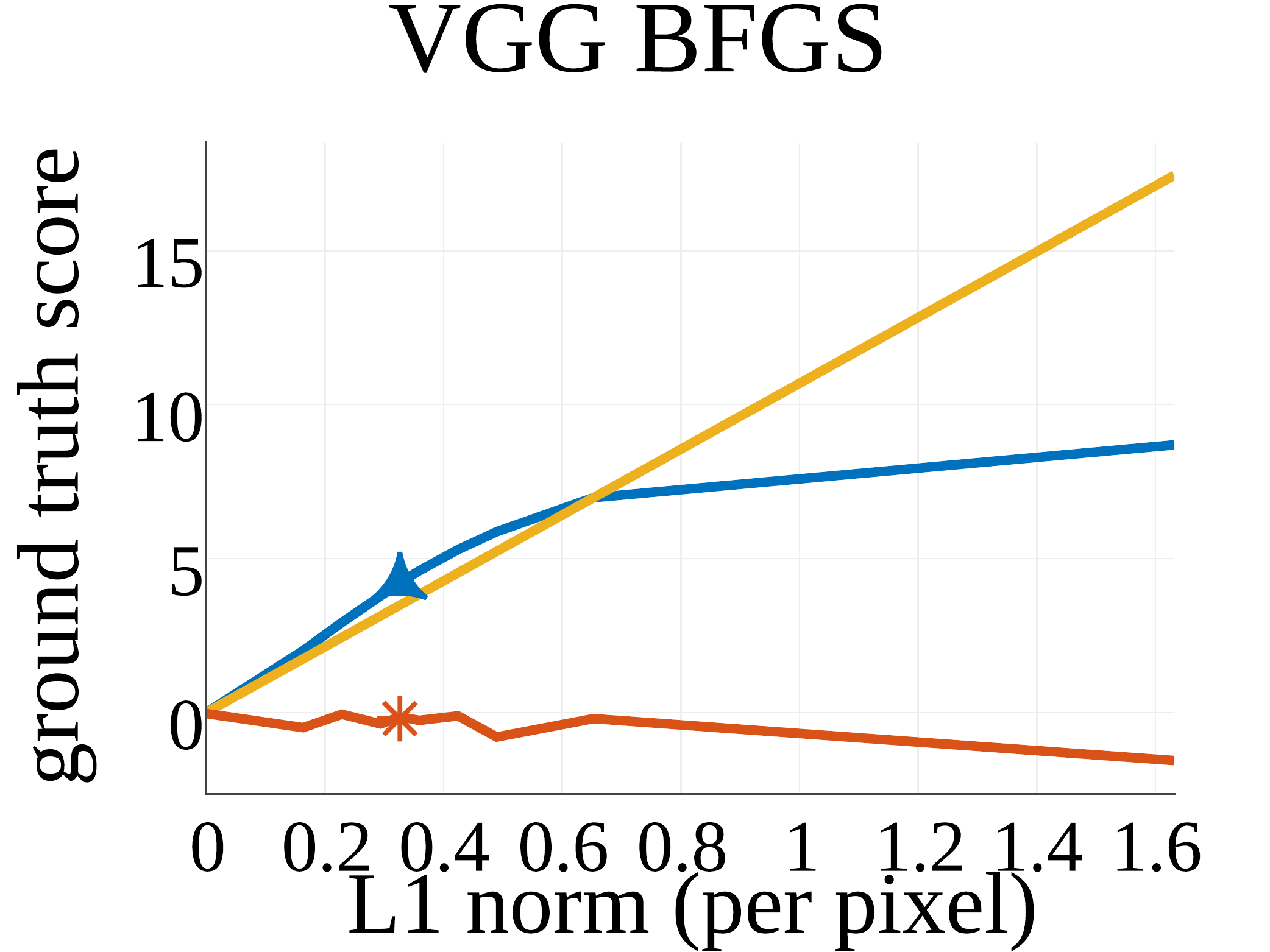}} 	\vspace{-1ex}\\
			\subfloat{\includegraphics[width=.305\textwidth]{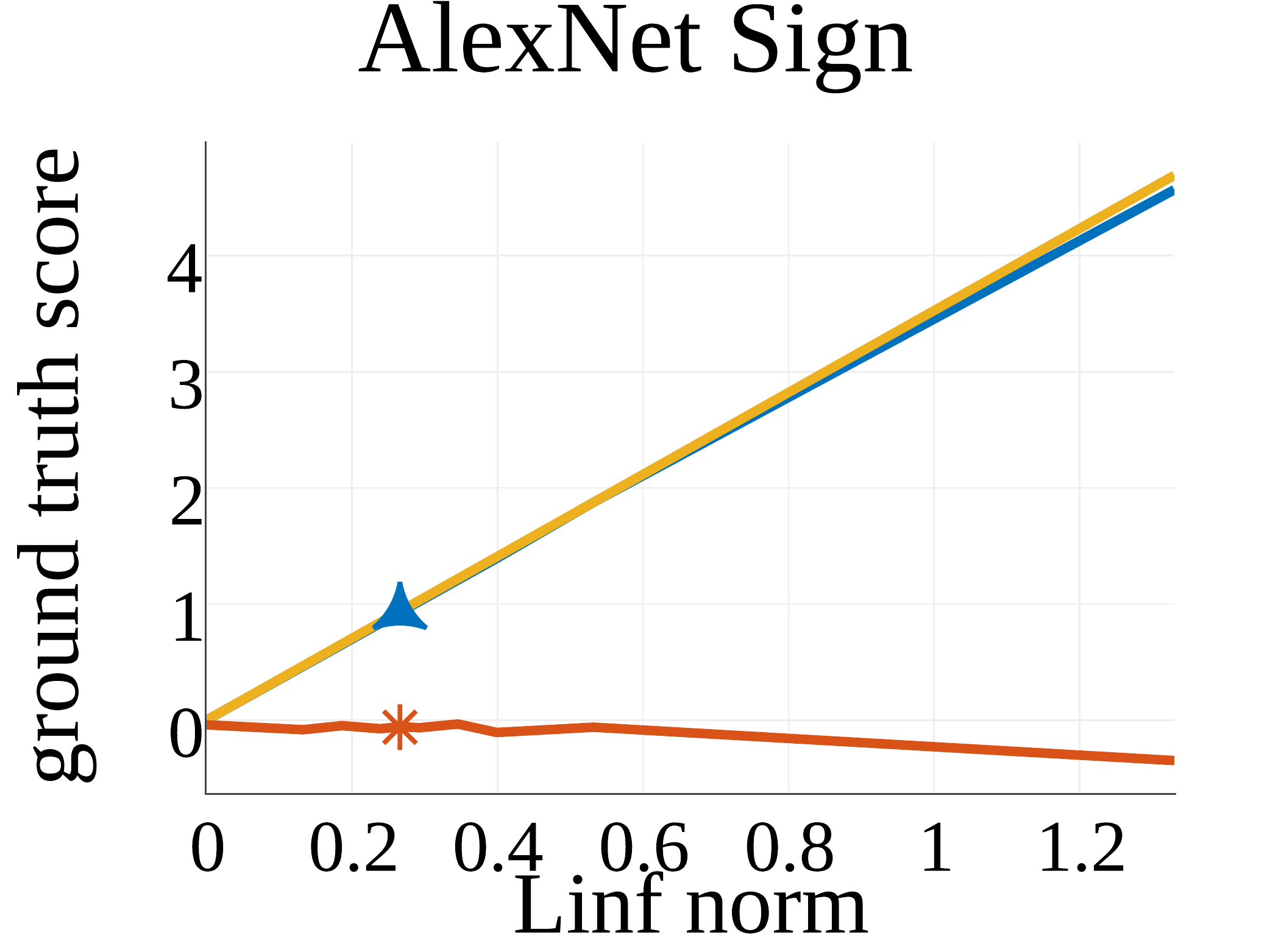}} &
			\subfloat{\includegraphics[width=.305\textwidth]{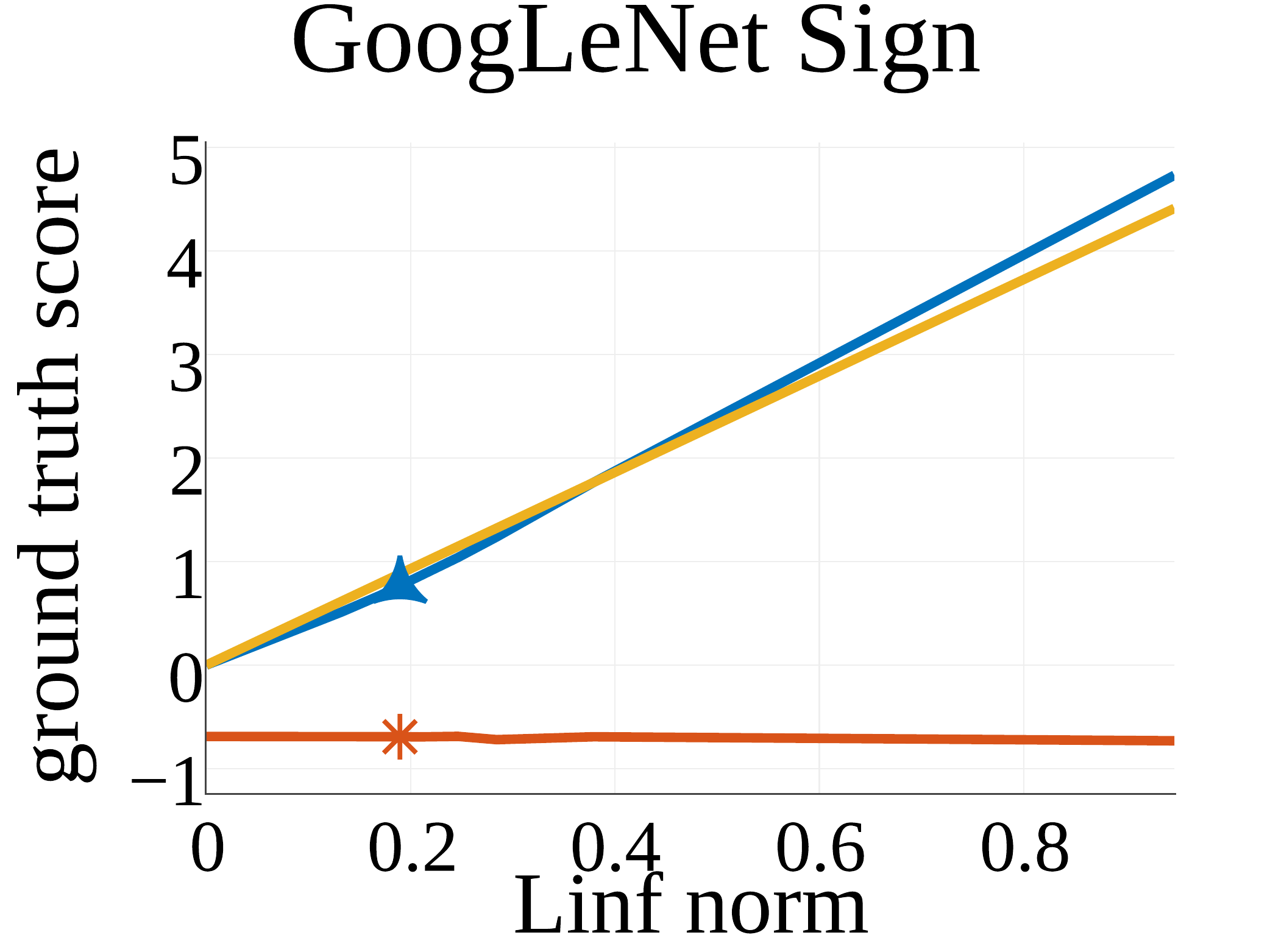}} &
			\subfloat{\includegraphics[width=.305\textwidth]{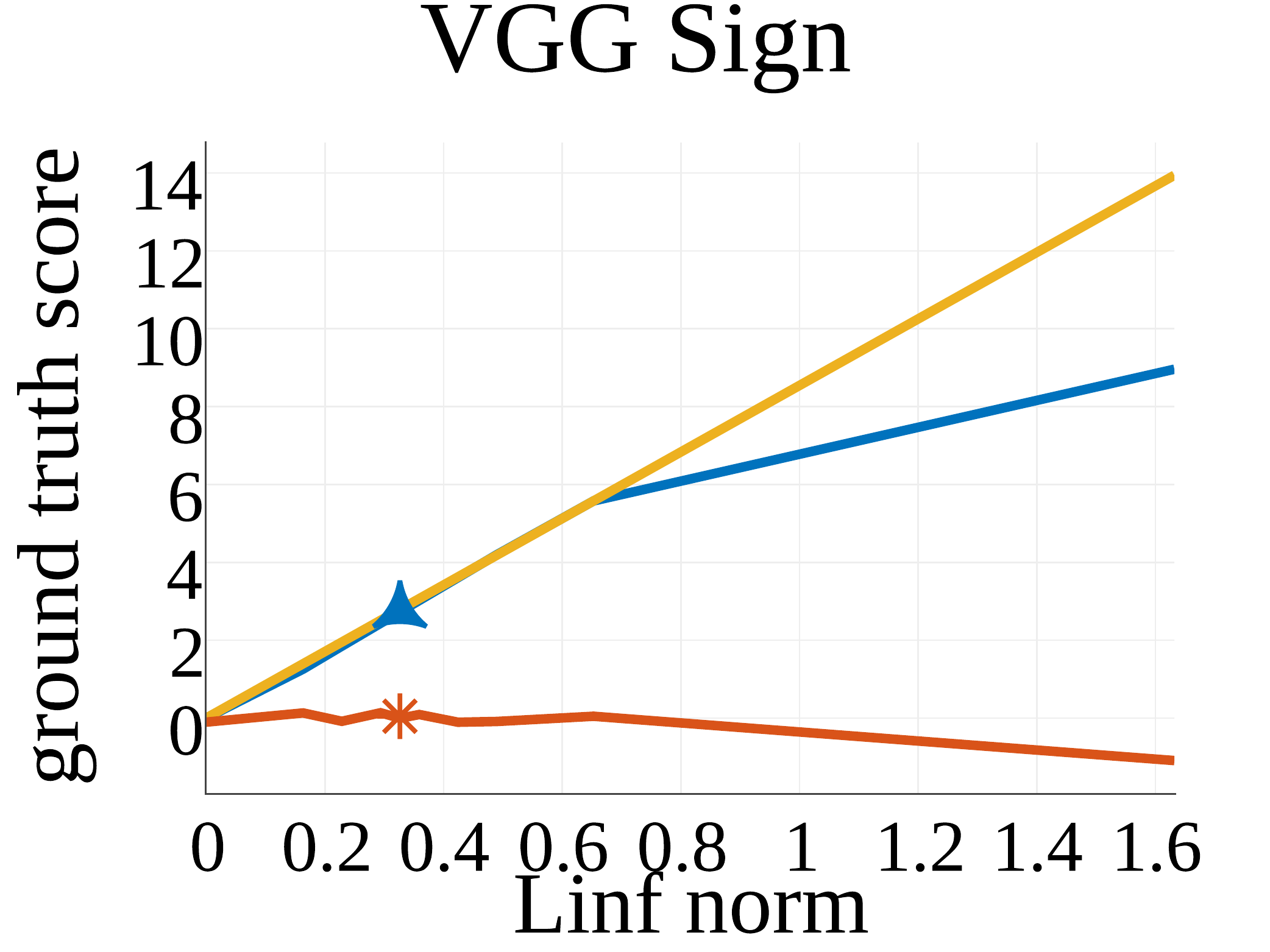}} \\
		\end{tabular}
	}
	\caption
	{
	\textit{Local Linearity of the CNNs.} 
		 Classification score of the ground truth object category for the image in Fig.~\ref{fig:l1example}, when varying the $L_1$ norm  per pixel of \emph{BFGS} and $L_\infty$ for \emph{Sign}.
		}
	\label{fig:linearexample18}
\end{figure*}
\begin{figure*}[t!]
	\centering
	\vspace{-2ex}
	\raisebox{0\height}{\subfloat{\includegraphics[width=0.55\textwidth]{linearity_example/linearexample_legend}}}
	{\renewcommand{\arraystretch}{0.1}
		\setlength{\tabcolsep}{2pt}
		\begin{tabular}{ccc}
			\subfloat{\includegraphics[width=.305\textwidth]{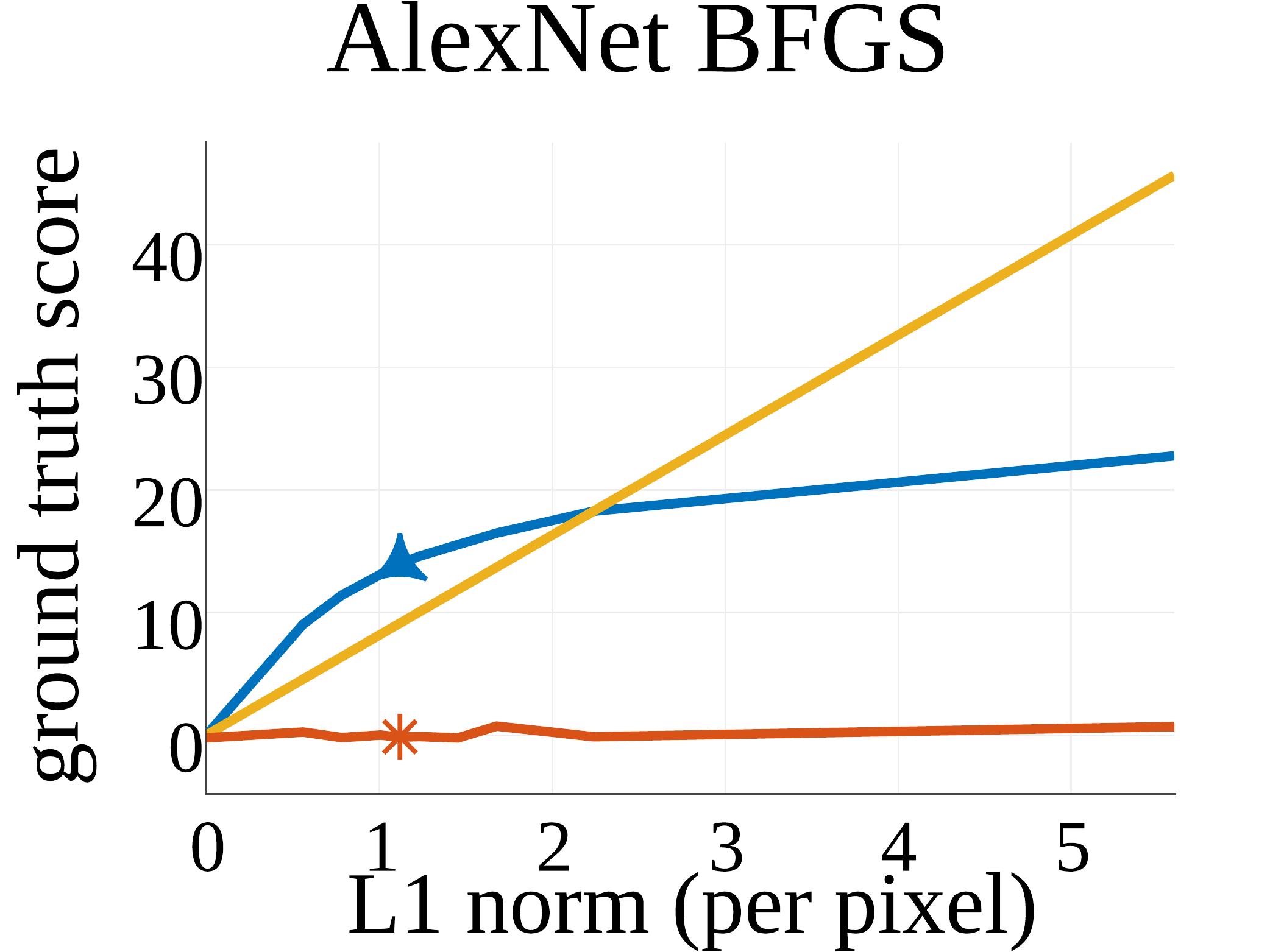}} &
			\subfloat{\includegraphics[width=.305\textwidth]{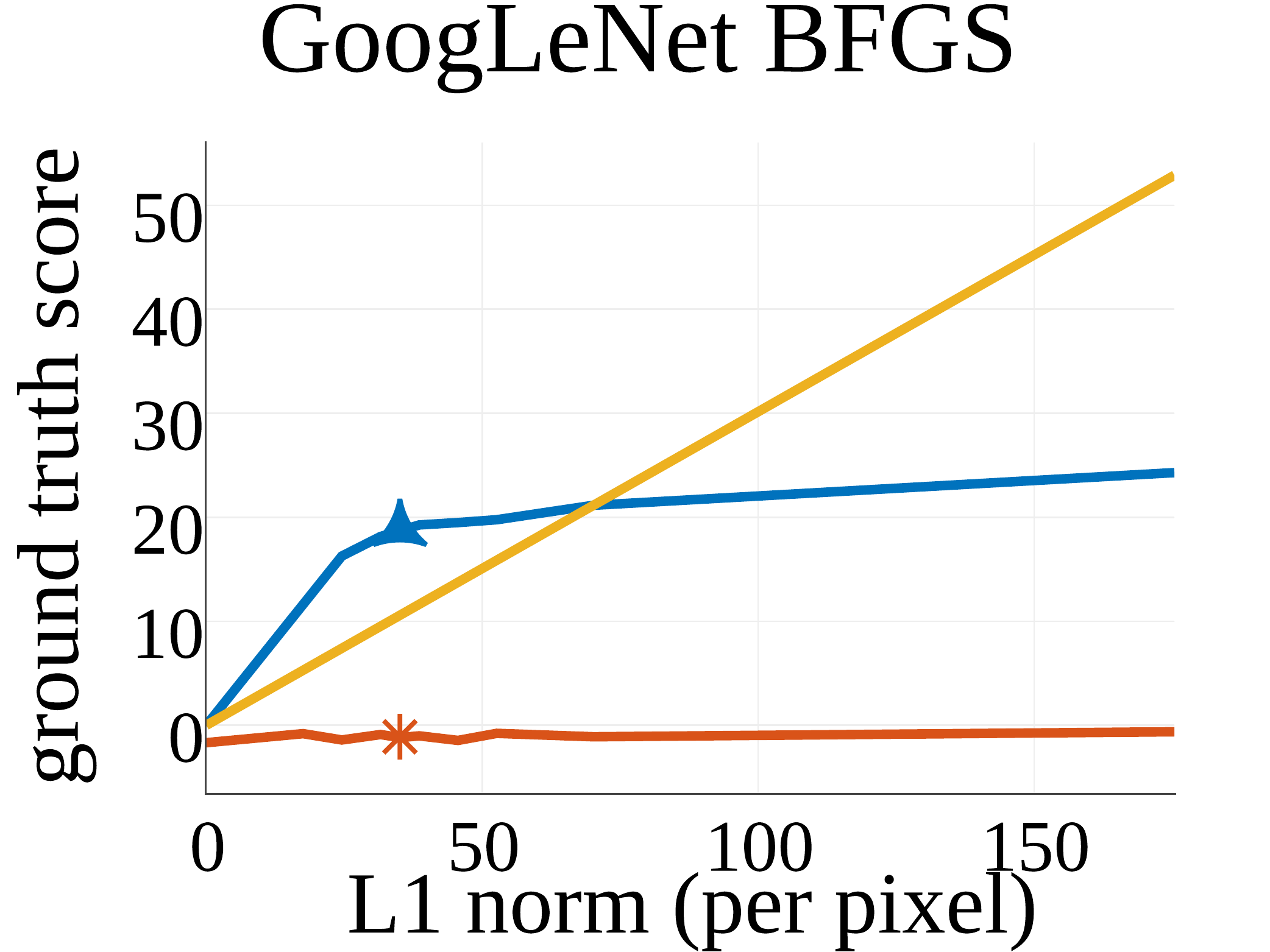}} &
			\subfloat{\includegraphics[width=.305\textwidth]{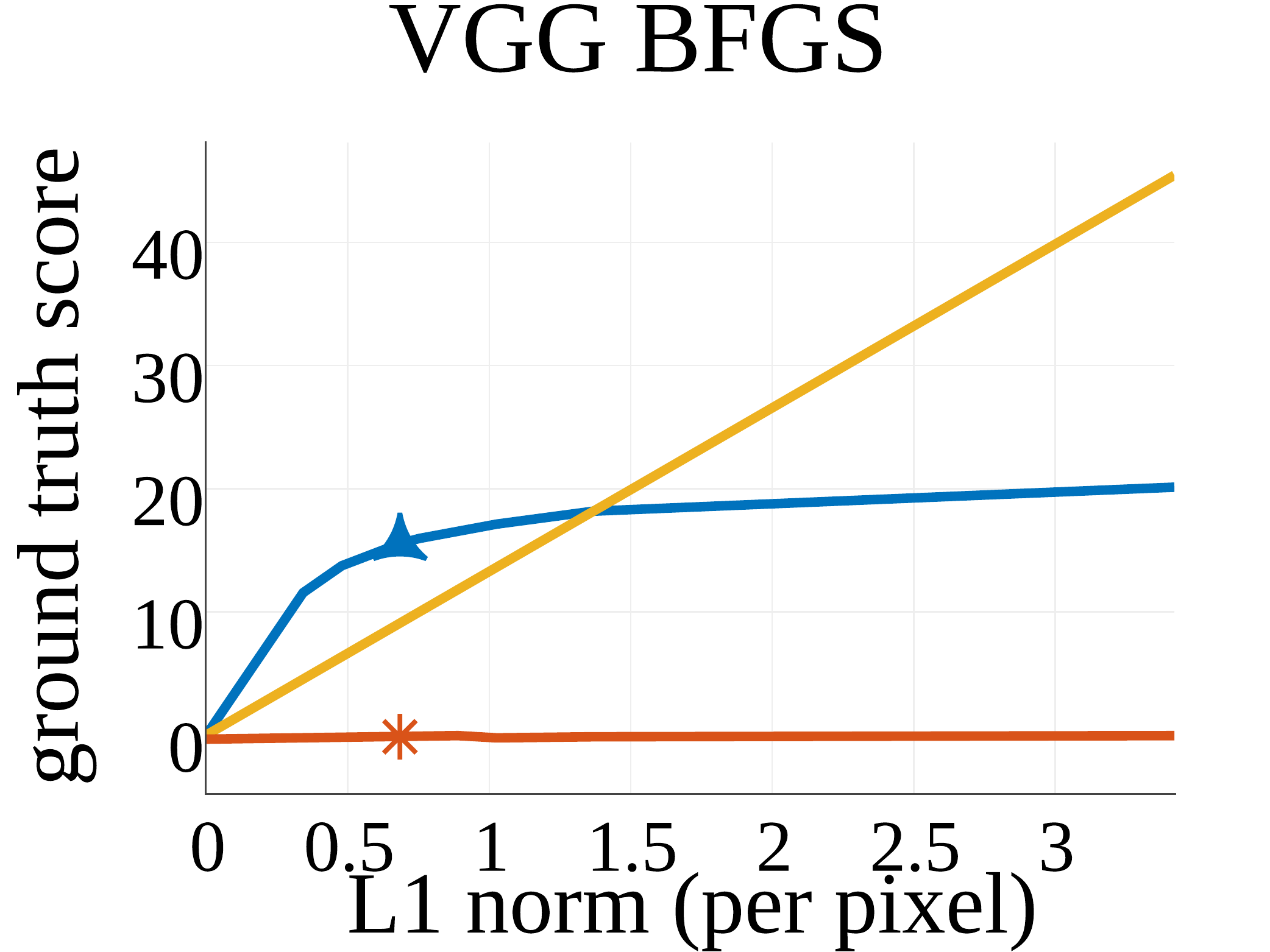}} 	\vspace{-1ex}\\
			\subfloat{\includegraphics[width=.305\textwidth]{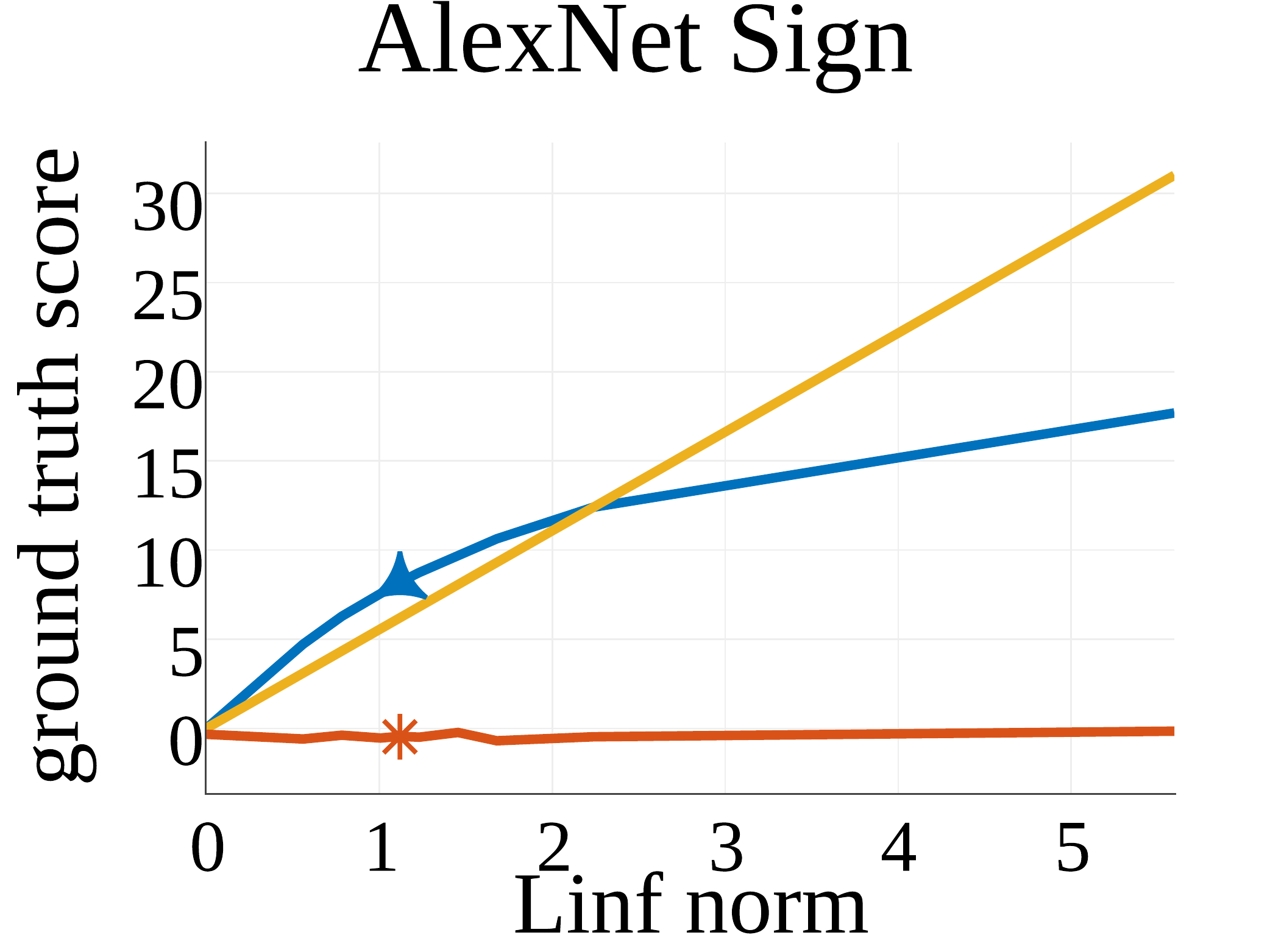}} &
			\subfloat{\includegraphics[width=.305\textwidth]{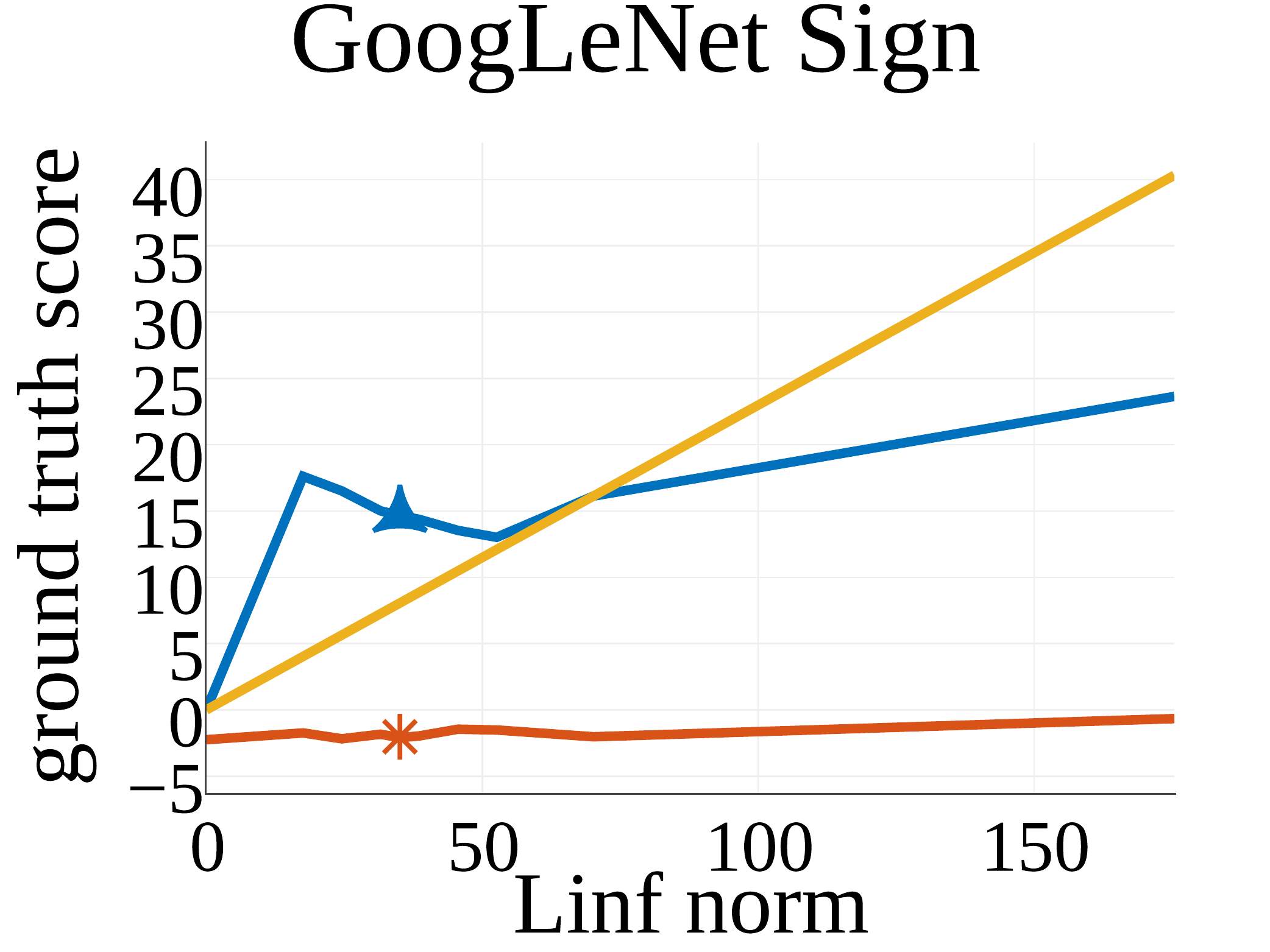}} &
			\subfloat{\includegraphics[width=.305\textwidth]{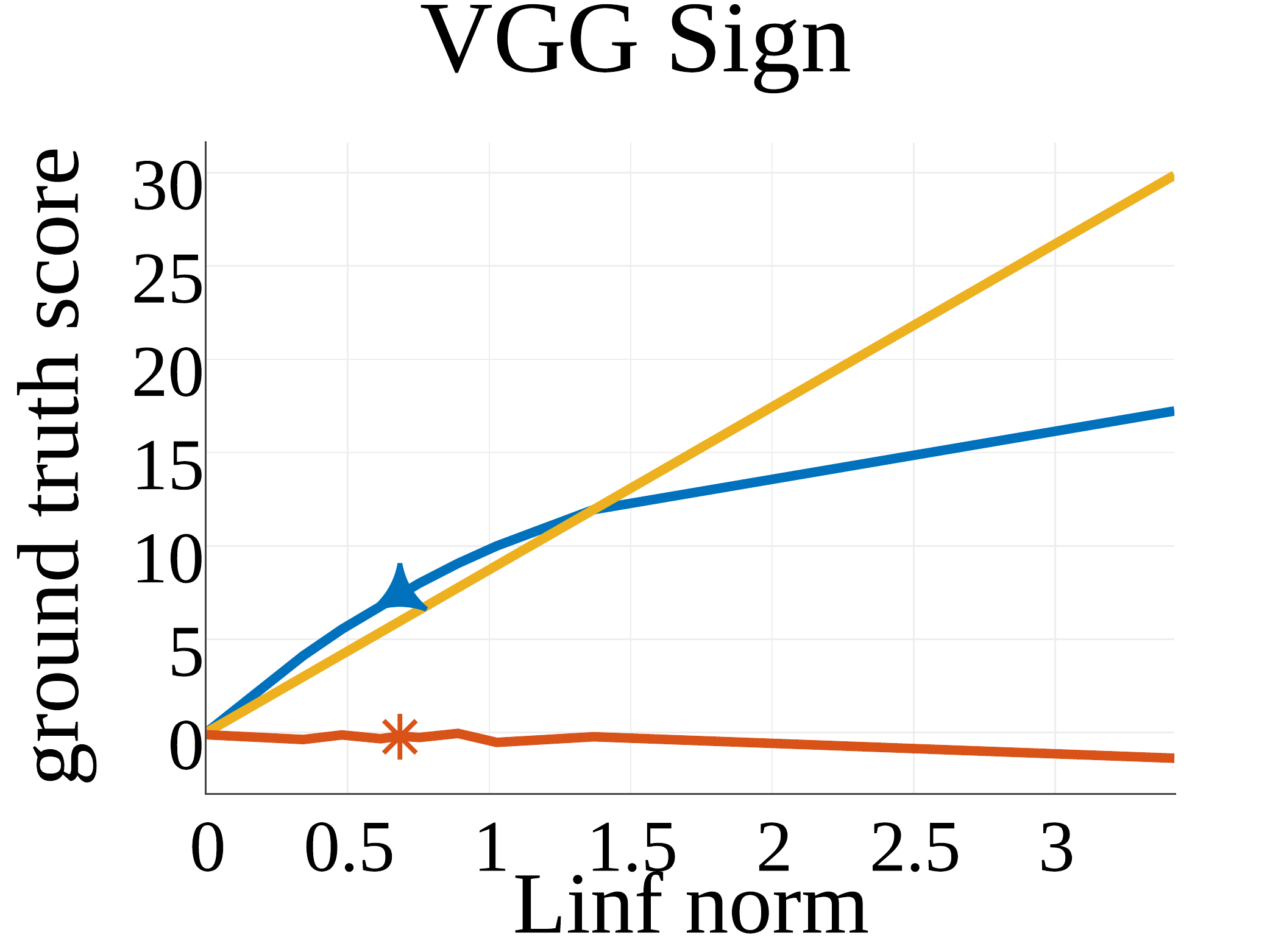}} \\
		\end{tabular}
	}
	\caption
	{
				\textit{Local Linearity of the CNNs.} 
		 Classification score of the ground truth object category for the image in Fig.~\ref{fig:l1example2}, when varying the $L_1$ norm  per pixel of \emph{BFGS} and $L_\infty$ for \emph{Sign}.
		 }
	\label{fig:linearexample75}
\end{figure*}
\begin{figure*}[t!]
	\centering
	\vspace{-2ex}
	\raisebox{0\height}{\subfloat{\includegraphics[width=0.55\textwidth]{linearity_example/linearexample_legend}}}
	{\renewcommand{\arraystretch}{0.1}
		\setlength{\tabcolsep}{2pt}
		\begin{tabular}{ccc}
			\subfloat{\includegraphics[width=.305\textwidth]{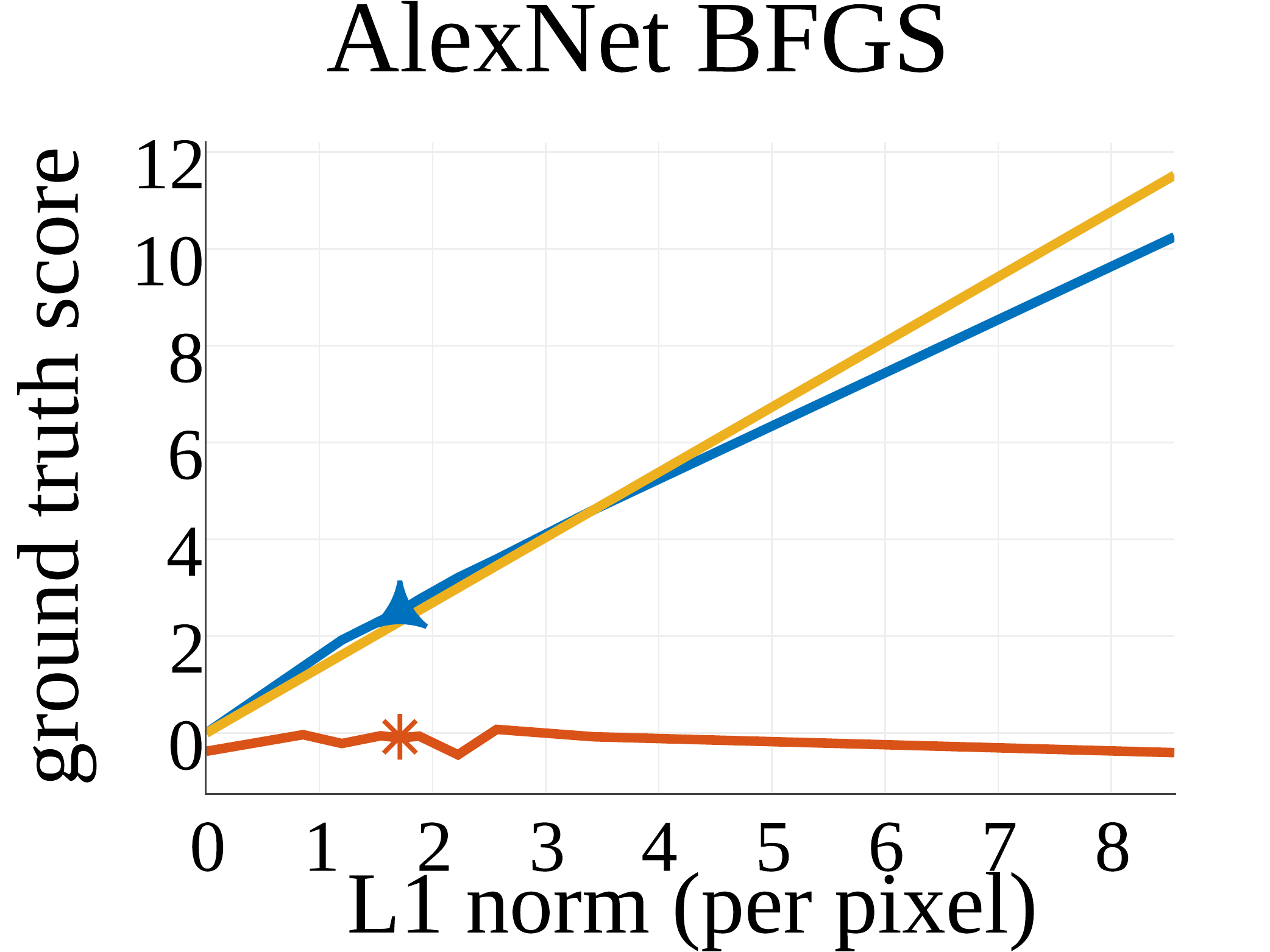}} &
			\subfloat{\includegraphics[width=.305\textwidth]{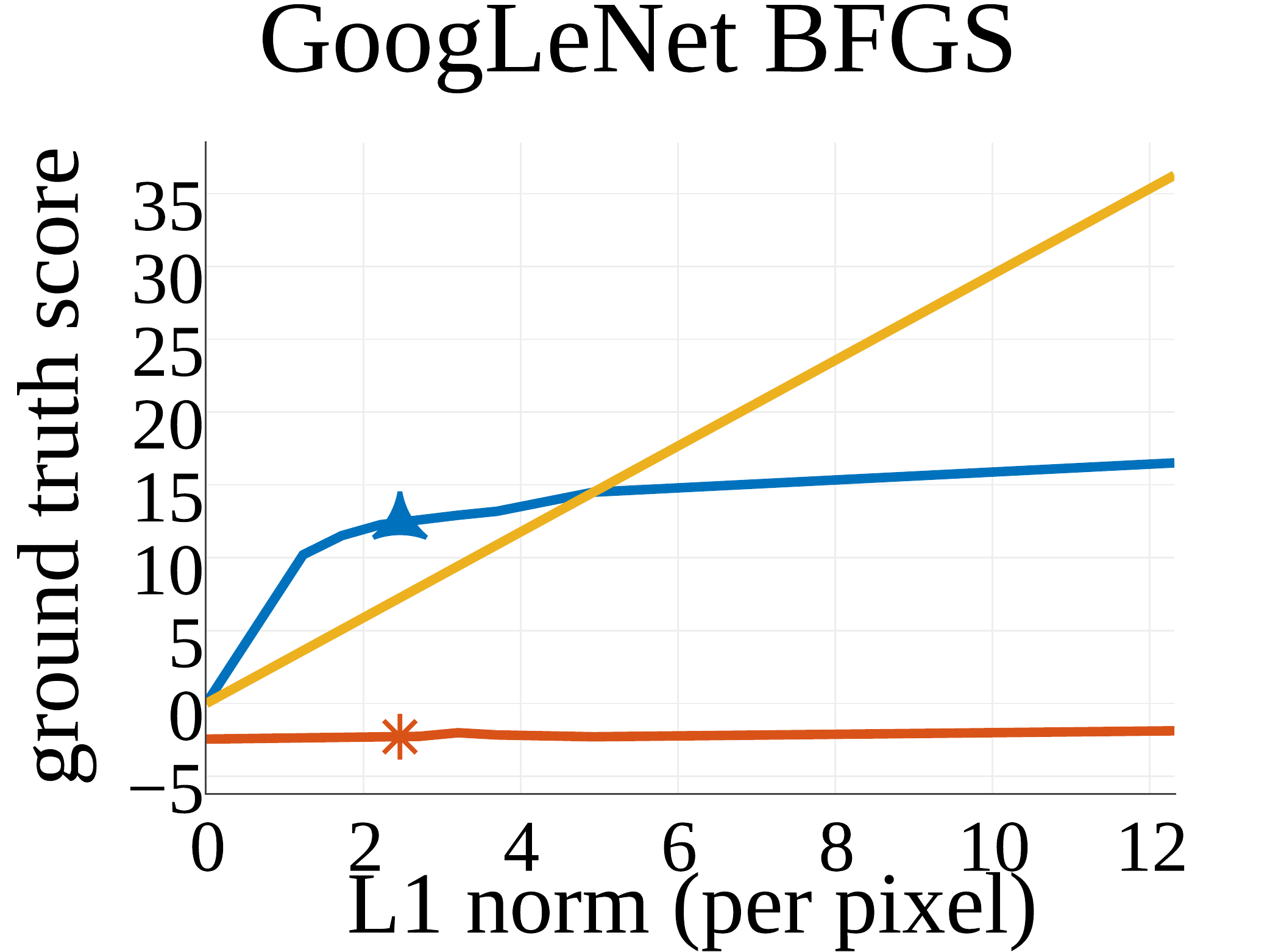}} &
			\subfloat{\includegraphics[width=.305\textwidth]{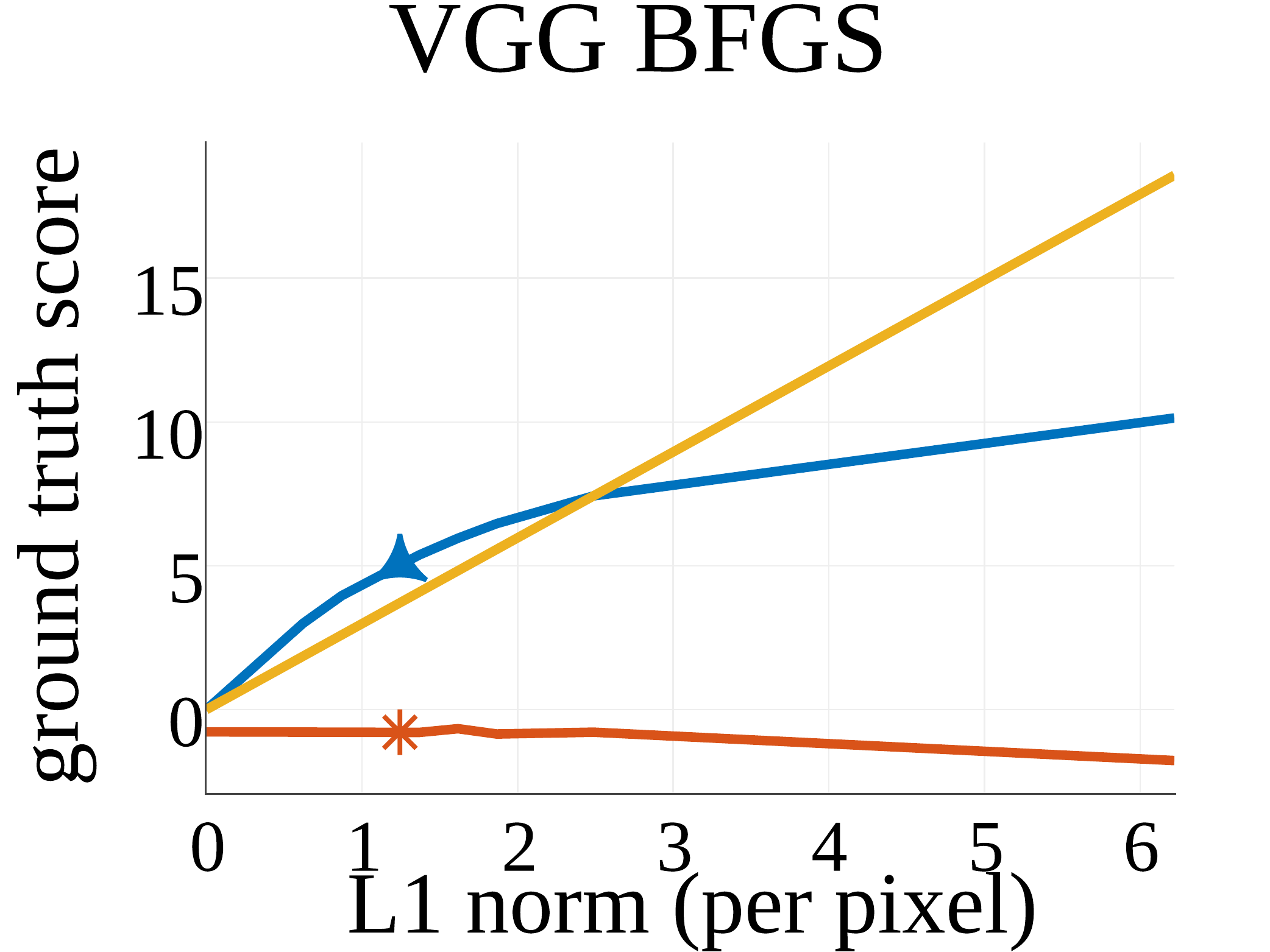}} 	\vspace{-1ex}\\
			\subfloat{\includegraphics[width=.305\textwidth]{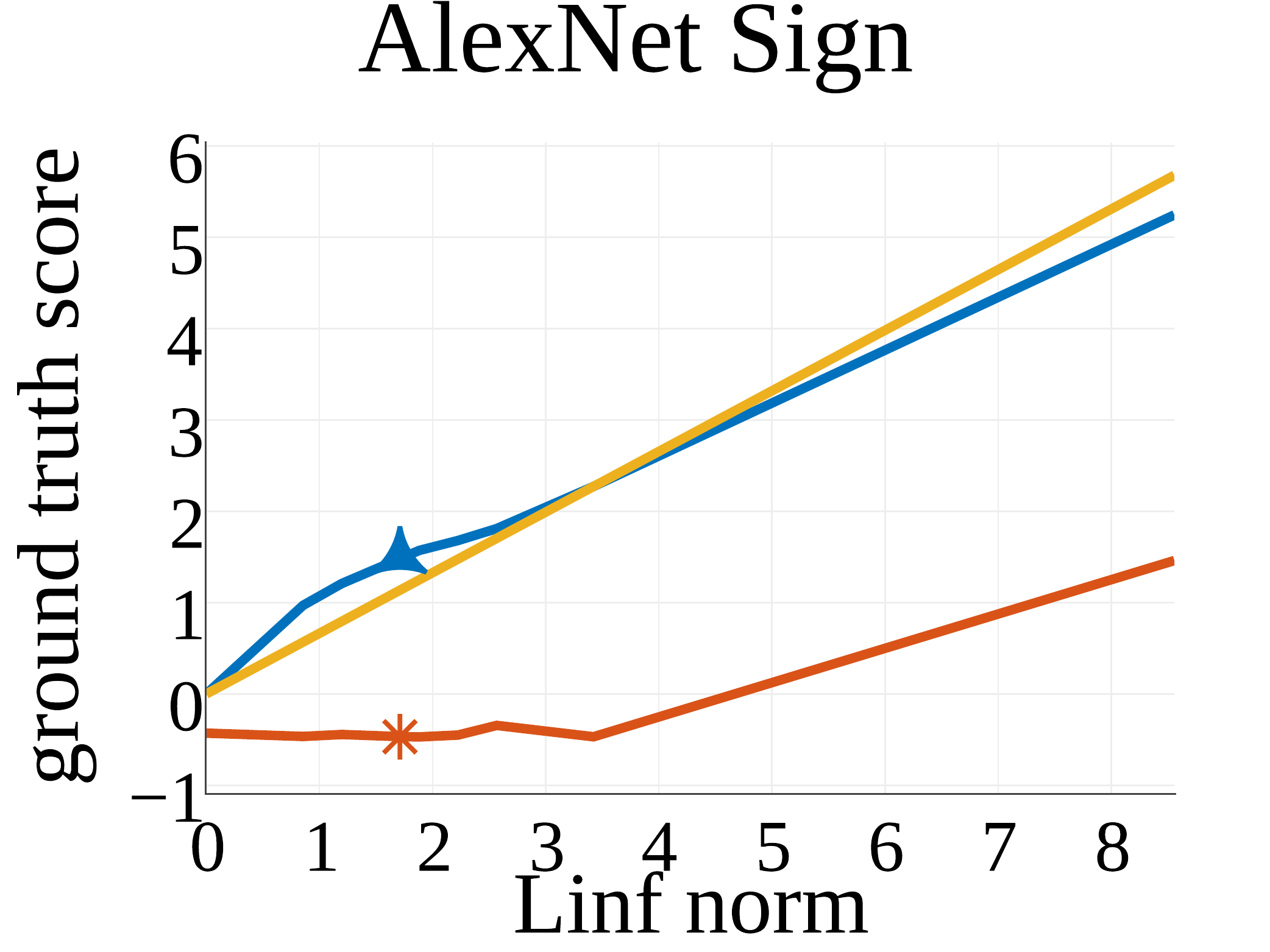}} &
			\subfloat{\includegraphics[width=.305\textwidth]{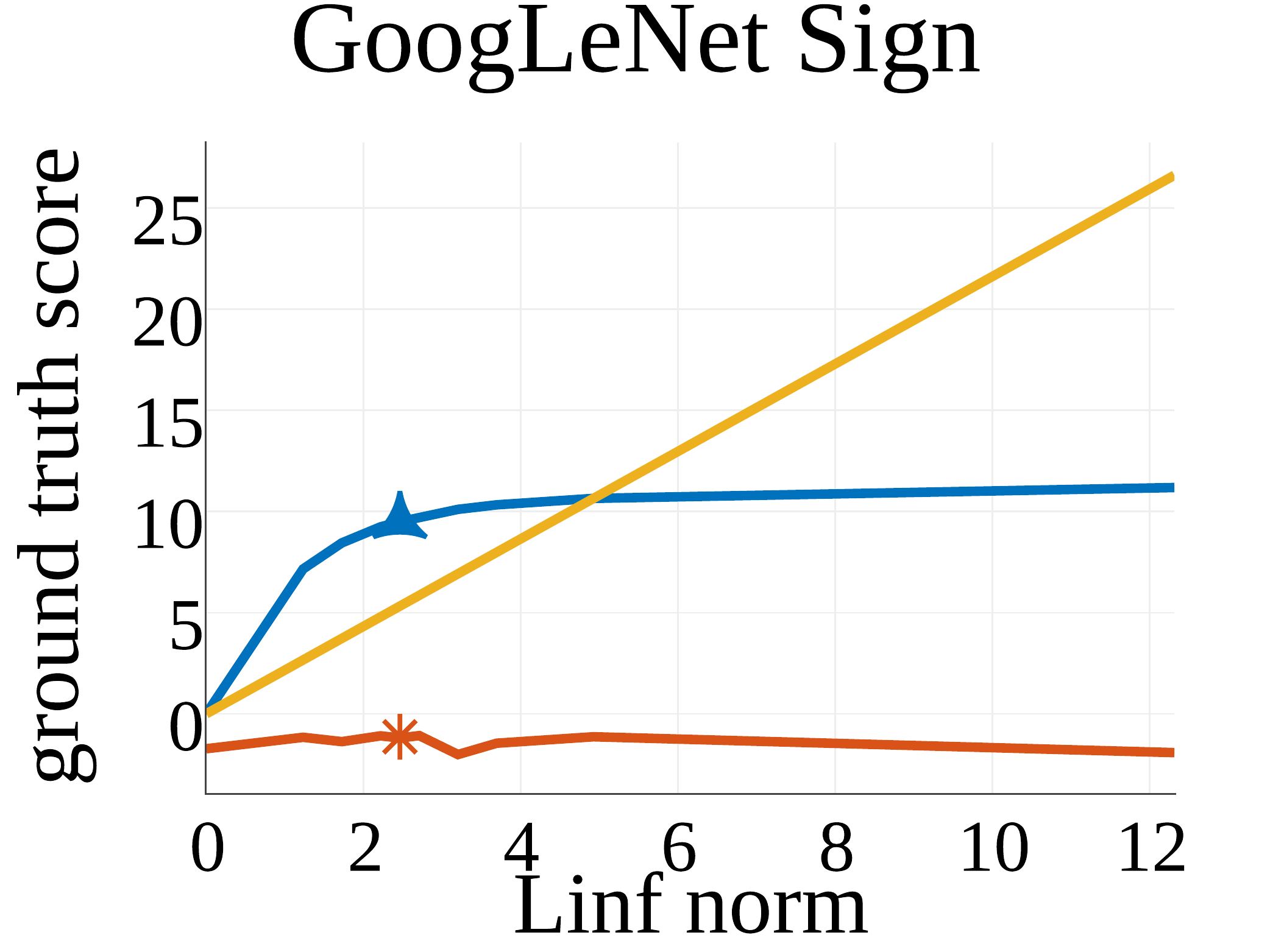}} &
			\subfloat{\includegraphics[width=.305\textwidth]{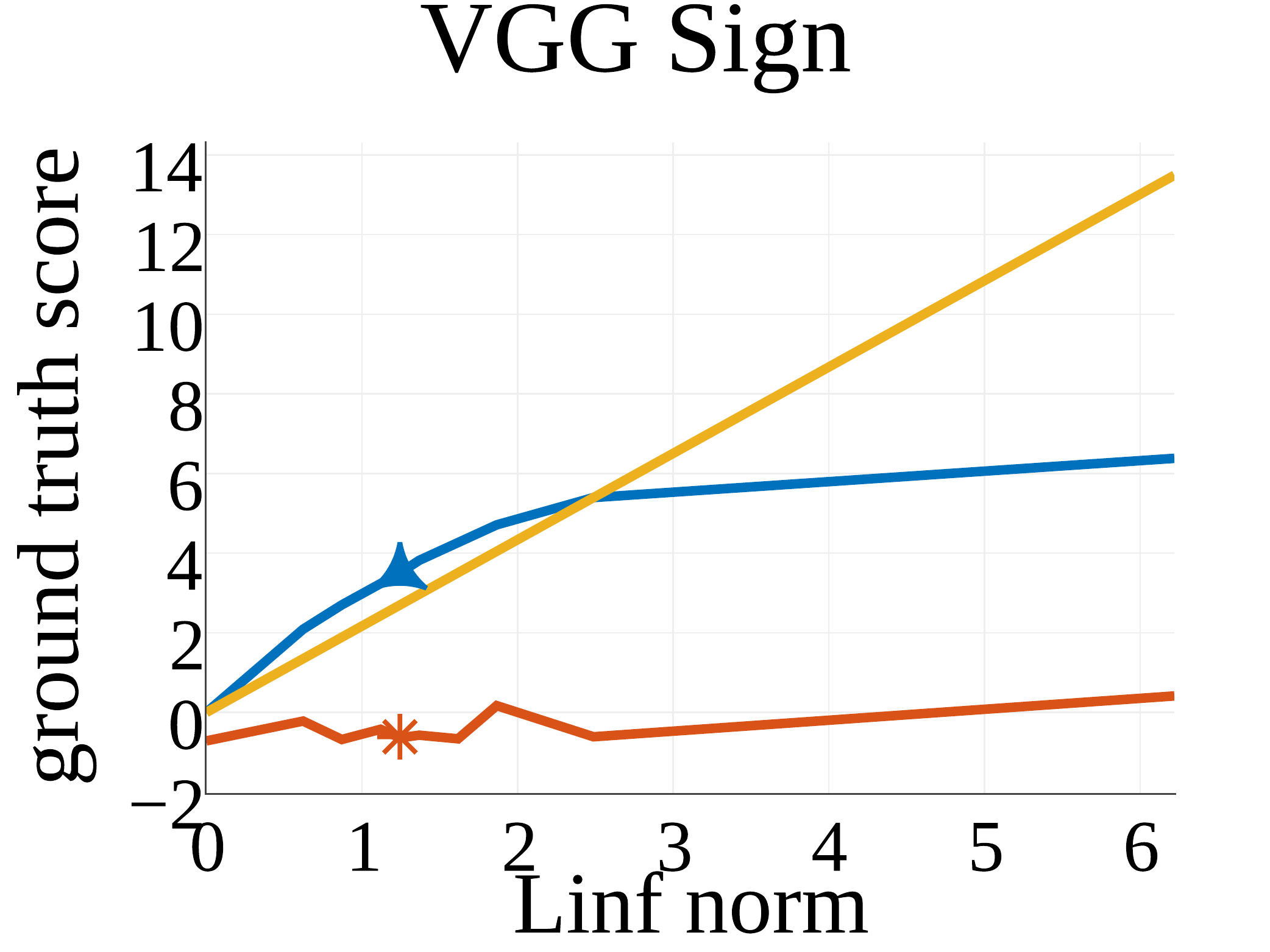}} \\
		\end{tabular}
	}
	\caption
	{
	\textit{Local Linearity of the CNNs.} 
		 Classification score of the ground truth object category for the image in Fig.~\ref{fig:l1example3}, when varying the $L_1$ norm  per pixel of \emph{BFGS} and $L_\infty$ for \emph{Sign}.
		}
	\label{fig:linearexample174}
\end{figure*}

{\bf CNNs act Locally Linearly on the Positions of a Recognized Object \;}
We test if CNNs act as a linear classifier by evaluating whether $f(\Mat{x}+\Mat{\epsilon}^\star)=f(\Mat{x})+f(\Mat{\epsilon}^\star)$ holds in practice. We evaluate  $f(\Mat{\epsilon}^\star) $ and $f(\Mat{x}+\Mat{\epsilon}^\star)-f(\Mat{x})$ for each image, varying the norm of $\Mat{\epsilon}^\star$, for the ground truth object category. We remove from the CNNs the last soft-max layer, because this layer may distort the results of evaluating the linearity, and the accuracy of the CNN  at test time is the same (the soft-max preserves the ranking of the classification scores). In Fig.~\ref{fig:removenoise}b, we show an example for \emph{BFGS} and AlexNet CNN for the image of Fig.~\ref{fig:examples}. We indicate the point at which the perturbation produces the misclassification, $\Mat{\epsilon}^\star$.  We show more examples for \emph{BFGS} and \emph{Sign} in Fig.~\ref{fig:linearexample18},~\ref{fig:linearexample75},~\ref{fig:linearexample174} in~\ref{app:Results}.  We can see that   $f(\Mat{\epsilon}^\star)$ is completely unrelated to $f(\Mat{x}+\Mat{\epsilon}^\star) - f(\Mat{x})$, but there is some local linearity in the vicinity of the image without perturbation, $\Mat{x}$, in the direction of $\Mat{\epsilon}^\star$. Thus, the hypothesis that the CNNs behave as a linear classifier needs to be reviewed in order to reconcile the hypothesis with these results. We introduce the next hypothesis that is able to explain all the previous results. 

\begin{hypothesis}
\label{hyp:linearity}
CNNs are locally linear to changes on the regions of the image that contain objects recognized by the CNN, otherwise the CNN may act non-linearly. 
\end{hypothesis}

We use the expression ``locally linear'' to denote that the linearity of the CNN only holds for perturbations of $\Mat{x}$ that are small,~\ie images that are in the vicinity of $\Mat{x}$. Also, note that the local linearity is only for perturbations on the image regions of $\Mat{x}$ that cause the CNN to recognize an object. The Hypothesis~\ref{hyp:linearity} leaves open whether the CNN behaves locally linearly or not in image regions that do not cause the CNN to detect an object, as we have not empirically analyzed what type of objects that are not detected by the CNN may produce a locally linear behavior of the CNN.

As in previous works, the hypothesis suggests that  adversarial examples are a consequence of CNNs acting as a high-dimensional linear classifier. In our hypothesis this is because the CNN acts as a linear classifier in the vicinity of the images with recognized objects. Also, our hypothesis predicts the result in Fig.~\ref{fig:removenoise}b about why  $f(\Mat{\epsilon}^\star)$ is not equal to $f( \Mat{x}+\Mat{\epsilon}^\star) - f(\Mat{x})$,  because $f(\Mat{\epsilon}^\star)$ is  in the non-linear region of the CNN since  $\Mat{\epsilon}^\star$  is not an image of an object that is recognized by the CNN. 
Finally, our hypothesis explains why the perturbation is always aligned with the classifier on the position of the recognized object, as in a different position the CNN may not behave linearly. 

We provide quantitative results in Fig.~\ref{fig:removenoise}c to show that the local linearity hypothesis reasonably holds in the vicinity of $\Mat{x}$. We approximate the alignment of the perturbation with an hypothetical linear classifier,~\ie $\Mat{w}^\prime\Mat{\epsilon} $,  and we evaluate how much the classification score given by this hypothetical linear classifier deviates from the real classification score, $f(\Mat{x}+\Mat{\epsilon})-f(\Mat{x})$. To calculate $\Mat{w}^\prime\Mat{\epsilon} $ we trace a line between $f(\Mat{x})$ and $f(\Mat{x}+2\Mat{\epsilon}^\star)$, which gives  $\Mat{w}^\prime\Mat{\epsilon} $ in the direction of $\Mat{\epsilon}^\star$  (see the yellow line in Fig.~\ref{fig:removenoise}c). If the CNN would approximately act as a linear classifier between $\Mat{x}$ and $\Mat{x}+2\Mat{\epsilon}^\star$, then the error between $\Mat{w}^\prime\Mat{\epsilon}^\star $ and  $f(\Mat{x}+\Mat{\epsilon}^\star)-f(\Mat{x})$ would be $0$. In Fig.~\ref{fig:linearity2}c we report the cumulative histogram of the number of images with an error derived from the Hypothesis~\ref{hyp:linearity} smaller than a threshold, and we compare it with the error that yields calculating $f(\Mat{x}+\Mat{\epsilon}^\star)-f(\Mat{x})$ as  $f(\Mat{\epsilon}^\star)$, which comes from the assumption that the CNN is always a linear classifier. We can see that the error derived from the Hypothesis~\ref{hyp:linearity} is orders of magnitude smaller than the error from the hypothesis that the CNN is always a linear classifier.

Note that the Hypothesis~\ref{hyp:linearity} can be motivated from the non-linearities induced by the linear rectifier units (ReLUs)~\citep{Krizhevsky_NIPS_2012}, which are used in all CNNs we tested. ReLUs are applied to the neural responses after the convolutional layers of the CNNs. They have a linear behavior only for the active neural responses, otherwise they output a constant value equal to $0$. Thus, if the perturbation is added in a region where the neural responses are active, the CNN tends to behave more linearly than when the neural responses are not active.  When we increase the norm of the  perturbation,  the effect of the perturbation to the final classification score ($f(\Mat{x}+\Mat{\epsilon}^\star)-f(\Mat{x})$) stops increasing at the same linear pace because the number of ReLUs that return a $0$ value is higher than before increasing the norm of the perturbation. This can be seen in Fig.~\ref{fig:removenoise}b, and  Fig.~\ref{fig:linearexample18},~\ref{fig:linearexample75},~\ref{fig:linearexample174} in~\ref{app:Results}, when we increase the norm of the perturbation.

\section{Foveation-based Mechanisms Alleviate Adversarial Examples}

We define a foveation as a transformation of the image that selects a region in which the CNN is applied, discarding the information from the other regions.
The foveation provides an input to the CNN that always has the same size, even if the size of the selected region varies. Thus, the foveation may produce a change of the position and scale of the objects in the input of the CNN. It is well-known that the representations learned by the CNNs are robust to changes of scale and position of the objects, and we assume that the transformation produced by the foveation does not negatively affect the performance of the CNN. Also, we assume that the foveation mechanism does not introduce non-linearities,~\eg~it does not modify pixel values from the original image, and the interpolations for re-sizing the image are linear. Without loss of generality, we use as the foveation mechanism a crop of a region that includes most of the object or the whole object. Below, the details of several foveation mechanism we use are introduced.

Let $T(\Mat{x})$ be the image after the foveation. Recall  the local linearity hypothesis introduced in Hypothesis~\ref{hyp:linearity}, in the previous section, and recall the linearity of $T(\cdot)$. If the perturbation $\Mat{\epsilon}$ is applied in the same position of an object recognized by the CNN, we have that hypothetically, $f(T(\Mat{x}+\Mat{\epsilon}^\star))$ is equal to $\Mat{w}^\prime T(\Mat{x})+\Mat{w}^\prime T(\Mat{\epsilon}^\star)$ for small perturbations. Since the representations learned by the CNNs are robust to changes of scale and position of the objects produced by the foveation, the term without the perturbation after the foveation, $\Mat{w}^\prime T(\Mat{x})$, is not expected to have a lower accuracy than before, $\Mat{w}^\prime \Mat{x}$. In addition,  since $T(\cdot)$ may remove clutter from the image, applying the foveation could even improve the accuracy when there is not the perturbation, if $T(\Mat{x})$ does not remove any part of the target object. Yet, the main reason why foveations may improve the performance of CNNs for adversarial examples is due to the hypothesis we now introduce:

\begin{hypothesis}
\label{hyp:foveation}
The aforementioned robustness of the CNNs to changes of scale and position of the objects does not generalize to the perturbations.
\end{hypothesis}

Thus, the foveation mechanism reduces the alignment between the classifier and the perturbation, $\Mat{w}^\prime T(\Mat{\epsilon}^\star)$, but does not negatively affect the alignment between the classifier and the image without perturbation, $\Mat{w}^\prime T(\Mat{x})$. Note that CNNs are trained with objects at many different positions and scales, that produce representations robust to the transformations of the object, such as the transformation produced by $T(\cdot)$. However, objects are visually very different from the adversarial perturbations, and the robustness of CNNs to transformations of objects does not generalize to the perturbations.

In the following, we show experimental evidence that supports the Hypothesis~\ref{hyp:foveation}, and show that foveations can be used to greatly alleviate the effect of the adversarial examples.

\input{tex/setupexample}

\subsection{Foveation Mechanisms Used in Experiments} 
A foveation may increase the accuracy of the CNNs by removing clutter rather than alleviating the effect of the adversarial perturbation. To analyze the effect of clutter, we use two different experimental set-ups. 

\begin{figure*}[t!]
	\centering
	\vspace{-2ex}

	\begin{tabular}{c}
		\subfloat{\includegraphics[width=.5\textwidth]{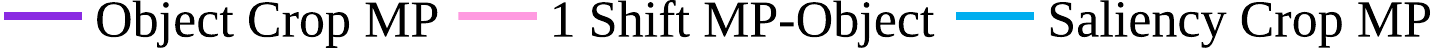}} \\
		\subfloat{\includegraphics[width=0.5\textwidth]{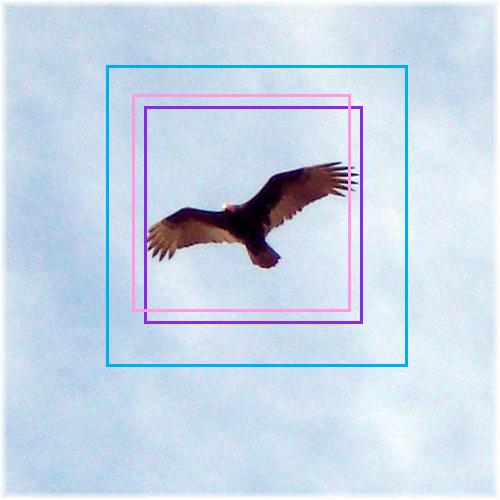}} \\
	\end{tabular}
	
	\caption{\textit{Example of Different Foveations.} \emph{Object Crop MP} is the purple bounding box and \emph{$1$ Shift MP-Object} is the pink bounding box. The blue bounding box corresponds to a crop from \emph{Saliency Crop MP}. Note that the center crop of \emph{$10$ Shift MP-Object} is exactly the same as \emph{Object Crop MP}.
	}
	\label{fig:cropqual}
\end{figure*}

{\bf {\emph{MP} set-up (with clutter)}  \;} In the first set-up, we use the adversarial example used in the previous section (see Fig.~\ref{fig:setupexamples}a and b). We use the following foveations:

{- \emph{Object Crop MP:}} We use as foveation mechanism the crop of the target object using the ground truth bounding box. If there are multiple bounding boxes in the image, because there are multiple target objects, we crop each of them, and average the classification scores.  Note that this foveation mechanism does not remove any part of the object, and it removes most of the clutter. See the purple bounding box in Fig.~\ref{fig:cropqual} in~\ref{app:Fixation}. The rest of the foveations we use do not guarantee that all the clutter is removed, and also, they may remove part of the target object.

{- \emph{Saliency Crop MP:}} This foveation is based on using a state-of-the-art saliency model to select $3$ regions to crop from the most salient locations of the image, depicted in Fig.~\ref{fig:setupexamples}c and the cyan bounding box in Fig.~\ref{fig:cropqual} in~\ref{app:Fixation}. The crops are generated selecting three centroids of the saliency map. We use the SALICON saliency map, which extracts the saliency map using the same CNNs we test~\citep{Salicon_2015}. We observed that these saliency maps are robust to the adversarial perturbations, since the adversarial examples are generated to produce misclassification for object recognition, but not to affect the saliency prediction. The classification accuracy is computed by averaging the confidences from the multiple crops.

{- \emph{${10}$ Crop MP:}} Another foveation strategy we evaluate is the $10$ crops that is implemented in the Caffe library to boost the accuracy of the CNNs~\citep{Jia_arXiv_2014}. The crops are done on large regions of the image, which in most cases the target object is contained in the crops (each crop discards about $21\%$ of the area of the image). The crops are always of the same size, and there are $5$ of them ($4$ clamped at each corner of the image, and one in the center of the image). The images resulting from these $5$ crops are flipped, which makes a total of $10$ crops. 

{- \emph{${3}$ Crop MP:}} This is the same as \emph{${10}$ Crop MP}  but only with $3$ random crops selected among the $10$ crops. It can be used for a fair comparison between \emph{Saliency Crop MP} with \emph{$10$ Crop MP}, and to evaluate how much improvement comes from averaging multiple foveations.

{\bf{\emph{MP-Object} set-up (without clutter) \;}} In the second set-up, we generate the adversarial example for the image produced after cropping the target object with the ground truth bounding box (denoted as \emph{MP-Object}, see Fig.~\ref{fig:setupexamples}d and e, and Fig.~\ref{fig:objectmpexamples}). Note that \emph{MP-Object} is the adversarial example for an image  with almost no clutter. The foveations for this set-up are:

{- \emph{Embedded MP-Object:}} It consists on embedding the image of the cropped object with the perturbation (\emph{MP-Object}) to the full image. Note that this foveation adds the clutter back to the image.

{- \emph{${10}$ Shift MP-Object:}} It is based on the $10$ crops of Caffe we introduced before for \emph{$10$ Crop MP}. In this case, each of the $10$ crops is a shifted version of the target object, that removes part of the object and uses part of the background to do the crop (the foveation shifts a bit to the background, and it yields a cropped region with $12\%$ of background, as shown in Fig.~\ref{fig:setupexamples}f and Fig.~\ref{fig:cropqual} in Sec.~\ref{app:Fixation}). We set the size of the crops such that they do not modify the original scale of the object. 

{- \emph{${1}$ Shift MP-Object:}} It consists on selecting $1$ random crop from the $10$ shifts of \emph{$10$ Shift MP-Object}.

\subsection{Results} 

We now show that a foveation can produce significant improvements of the CNN accuracy with adversarial examples, and also that the impact of removing clutter or averaging the predictions of multiple foveations is lower than the effect of our Hypothesis~\ref{hyp:foveation}.

\begin{figure*}[t!]
	\centering
	\vspace{-2ex}
	\raisebox{0\height}{\subfloat{\includegraphics[width=0.9\textwidth]{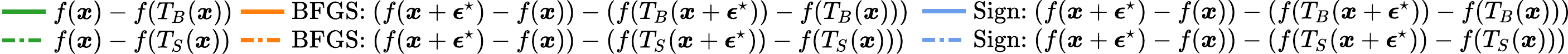}}}
	{\renewcommand{\arraystretch}{0.1}%
		\setlength{\tabcolsep}{2pt}
		\begin{tabular}{ccc}
			\subfloat{\includegraphics[width=.305\textwidth]{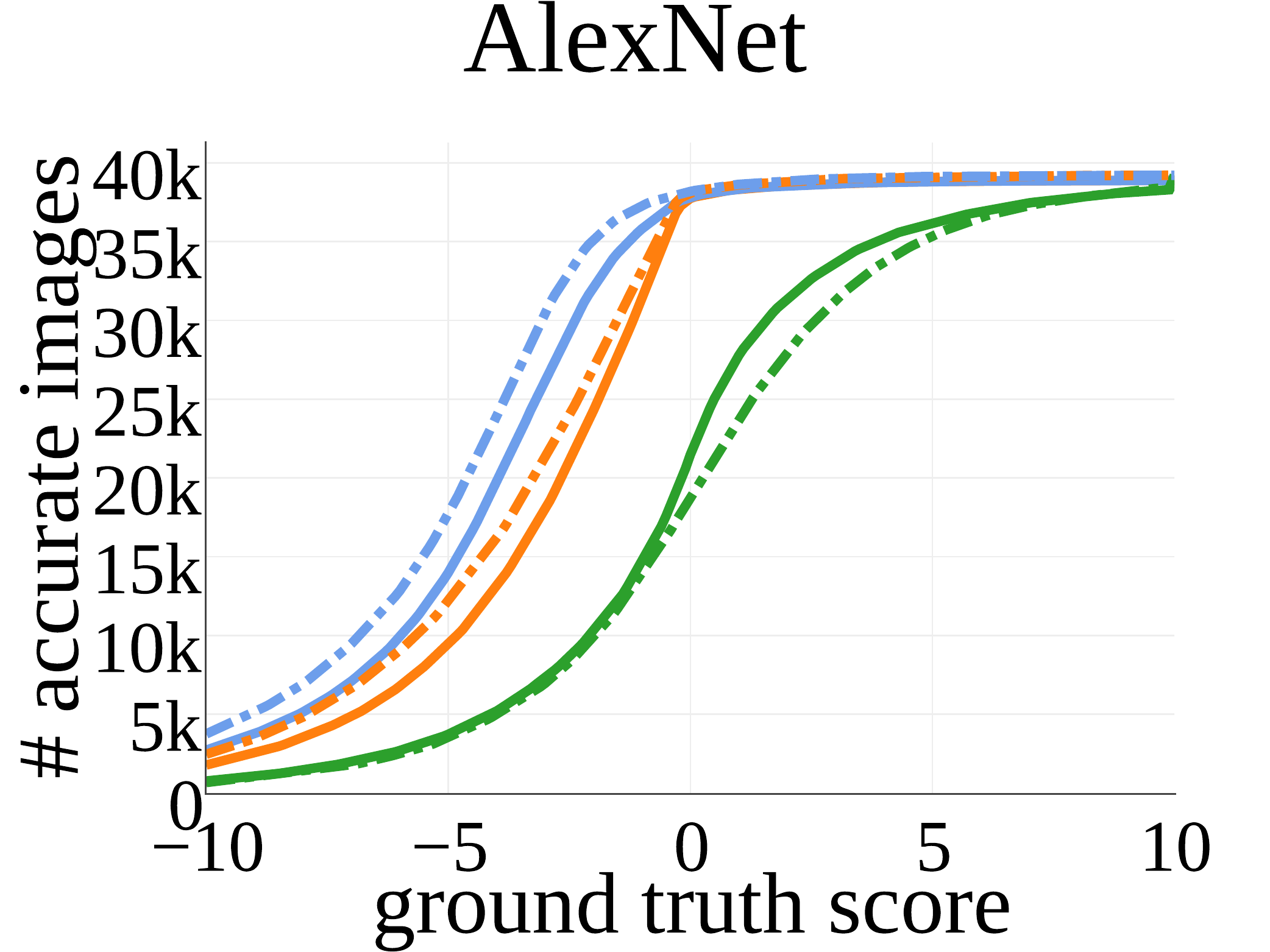}} &
			\subfloat{\includegraphics[width=.305\textwidth]{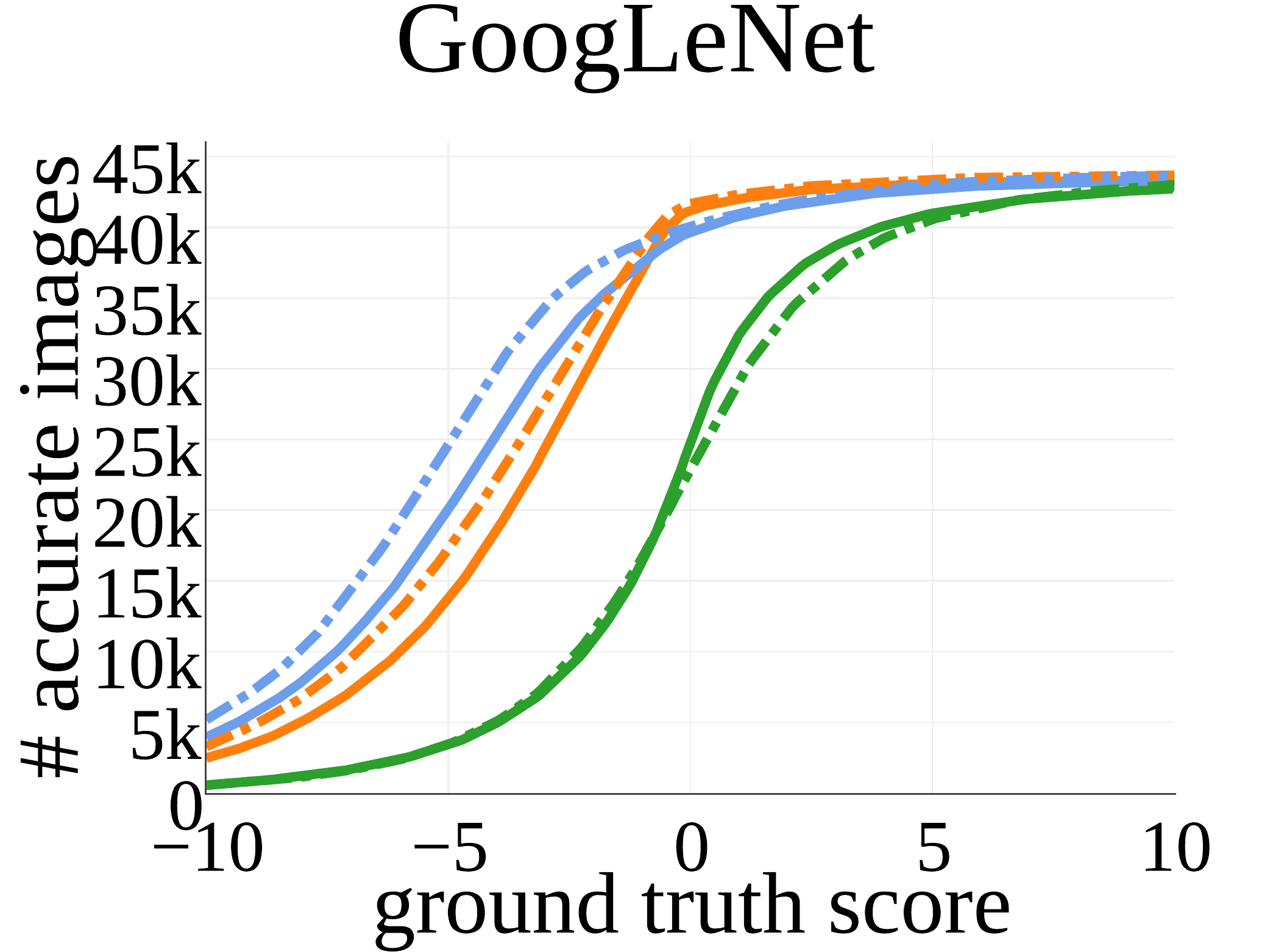}} &
			\subfloat{\includegraphics[width=.305\textwidth]{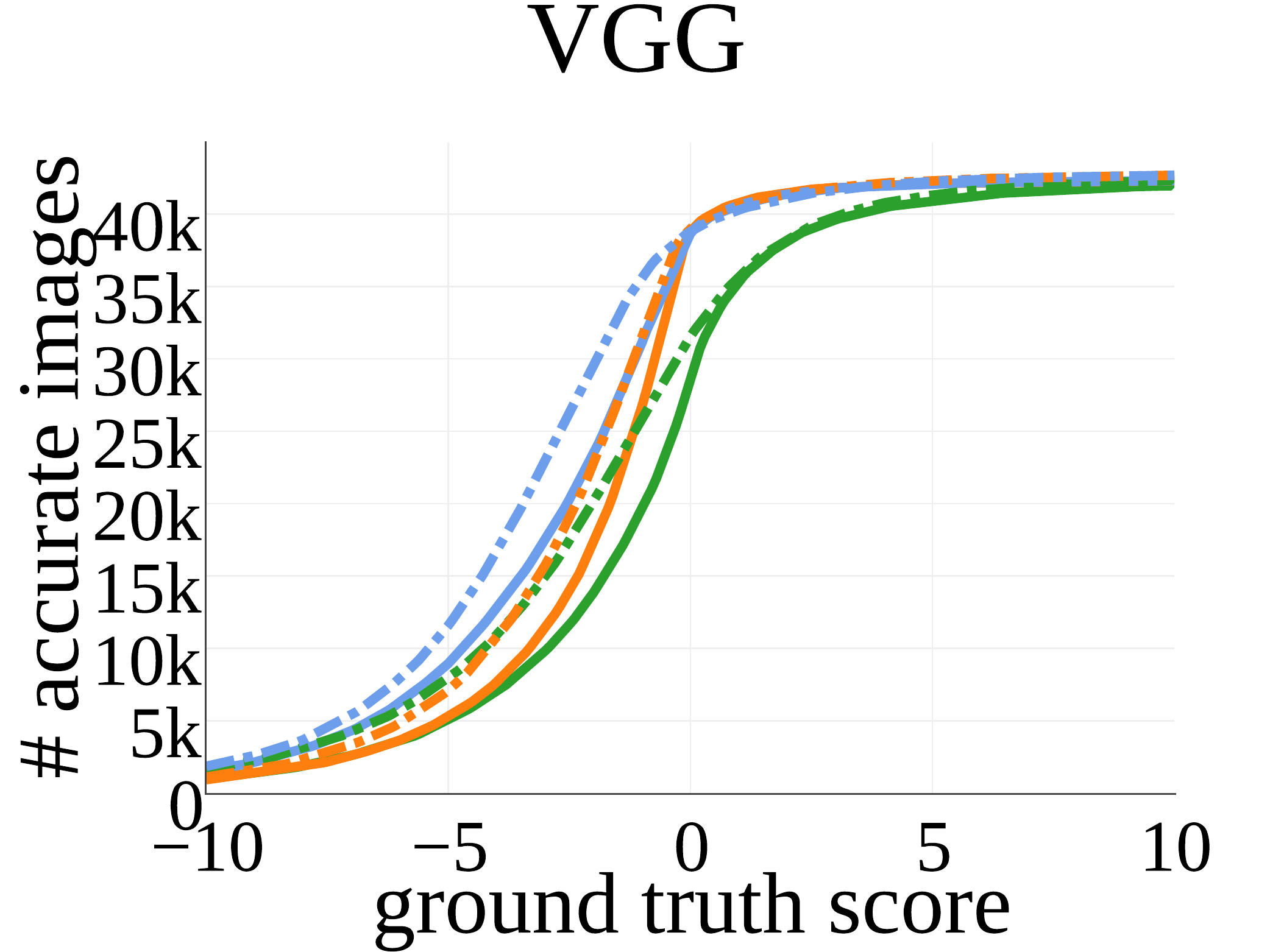}} 	\vspace{-1ex}\\
		\end{tabular}
	}
	\caption
	{
		\textit{Effect of the Foveation to the Perturbations.} 
Cumulative histogram of the number of images with a change of classification score smaller than the indicated in the horizontal axis. $T_B(\cdot)$ is the foveation with \emph{Object Crop MP}, and $T_S(\cdot)$ is the foveation with \emph{$1$ Shift MP-Object}.
		Only the images that are correctly classified by the CNN are included.
		}
	\label{fig:Optimal}
\end{figure*}

{\bf{Evaluation of the Hypothesis~\ref{hyp:foveation}} \;} We now test the hypothesis by analyzing the effect of the foveation to the classification scores of the adversarial perturbation and of the image without the perturbation. Namely,  we evaluate  $f(\Mat{x})-f(T(\Mat{x}))$ (how the classification score varies after the foveation for the image without perturbation), and  $\left( f(\Mat{x}+\Mat{\epsilon}^\star)-f(\Mat{x}) \right) - \left( f(T(\Mat{x}+\Mat{\epsilon}^\star))-f(T(\Mat{x}))  \right) $ (how the classification score varies after the foveation for the adversarial perturbation).  We do so for the classification score for the ground truth object category  (without the last soft-max layer, as in Sec.~\ref{sec:exp}), and for both set-ups. In Fig.~\ref{fig:Optimal}, we show the cumulative histogram of the number of images with a change of the classification score smaller than the indicated in the horizontal axis in the figure. We can see that the alignment between the perturbation and the classifier decreases after the foveation, and that the term of the image without perturbation is not affected as much as the term of the perturbation (\emph{BFGS} and \emph{Sign} are compared independently), which strongly supports Hypothesis~\ref{hyp:foveation}.

{\bf{Accuracy of the adversarial examples after the foveation } \;} Table~\ref{tbl:clutterMag}  shows the accuracy  before and after the foveation for the adversarial examples with the perturbation with minimum norm. In all cases, we can see that the accuracy after the foveation is almost the same as the accuracy without the adversarial perturbation, suggesting that foveations are a powerful mechanism to alleviate the adversarial examples (it improves the accuracy from $0\%$ to more than $70\%$ in all cases). \emph{Object Crop MP} produces the biggest improvement over all foveation mechanisms, probably because it is the only foveation mechanism that guarantees that  parts of the target object are not removed, and at the same time it removes the clutter. Note that the accuracy for the adversarial examples is not exactly $0\%$ because in some images the minimum norm of the perturbation is out of the range of the line search of the algorithm to calculate \emph{BFGS} and \emph{Sign} perturbations.

{\bf{How much does removing clutter or average the result of multiple foveations improve the accuracy? } \;} 
In Table~\ref{tbl:clutter} in Sec.~\ref{app:Fixation} we show the accuracy after the foveation when there is no adversarial perturbation. We can see that the foveation slightly  improves the accuracy in the absence of adversarial perturbation (about $5\%$), when we crop the bounding box  or average over multiple crops.  This suggests that removing clutter and averaging multiple crops always improve the accuracy, independently of the Hypothesis~\ref{hyp:foveation}. However, note that the improvement from removing clutter can not be the main reason of the improvement of the accuracy by the foveation when there is an adversarial perturbation, since in the set-up of \emph{MP-Object}, in which there is no clutter, the accuracy significantly improves after the foveation, and this improvement is almost the same as for the foveations that remove clutter in the \emph{MP} set-up. Also, we can see that the improvement of the accuracy from averaging multiple foveations is relatively smaller by comparing the accuracy of  \emph{$1$} and  \emph{$10$ Shift MP-Object}, and the accuracy of \emph{$1$} and  \emph{$10$ Crop MP}.

{\bf{How much does the accuracy decrease when the norm of the adversarial perturbation is increased?} \;} In Fig.~\ref{fig:clutter2} in~\ref{app:Fixation} we report the accuracy using different values of the norm of the perturbation before the foveation. When the perturbations are imperceptible or almost imperceptible, all the foveation mechanisms we introduce improve the accuracy between $30\%$ to $40\%$ in both set-ups, which further confirms that foveations considerably alleviate adversarial examples. Note that this result can not be directly compared with Table~\ref{tbl:clutterMag},  because here the norm of the perturbation is fixed for all images, while in Table~\ref{tbl:clutterMag} uses the minimum norm for each image.  Fig.~\ref{fig:clutter2} also shows that when we increase the value of the norm of the perturbation, the accuracy decreases. This is because large perturbations bring the CNN to the non-linear region (Sec.~\ref{sec:exp}), and our hypothesis of the foveations assumes that the CNN is working in the linear region. Also, when we increase the norm of the perturbation, CNNs fail because the perturbation is occluding the object rather than acting as an adversarial perturbation. We report the increase of the norm of the adversarial perturbation after the foveation in Table~\ref{tbl:normcomparison} in~\ref{app:Fixation}. Note that after the foveation the norm of the perturbation has to substantially increase to produce misclassification (about $5$ times for \emph{BFGS}, and between $5$ to $8$ times for \emph{Sign}).

\begin{table}[t!]

\centering
		\captionof{table}{\emph{Evaluation of the Foveation Mechanisms.} Quantitative results of the top-$5$ accuracy with minimum perturbation. See Table \ref{tbl:clutterMagAll} in~\ref{app:Fixation} for the results with \emph{Sign}.}
		\vspace{-2ex}

		{\scriptsize

		\begin{tabular}{ccccccccc}
		
			\toprule
			
			& \multicolumn{8}{c}{{BFGS Minimum Perturbation }} \\
			\cmidrule(lr){2-9} & \multicolumn{2}{c}{Before Foveation} & \multicolumn{2}{c}{After Foveation} & \multicolumn{2}{c}{Before Foveation} & \multicolumn{2}{c}{After Foveation} \\
			
			\cmidrule(lr){2-3} \cmidrule(lr){4-5} \cmidrule(lr){6-7} \cmidrule(lr){8-9}

			& {w/o} & {}  & {Object Crop} & {Saliency Crop} & {w/o} & {} & {1 Shift} & {Embedded}\\
			
			& {MP} & {MP}  & {MP} & {MP} & {MP-Object} & {MP-Object} & {MP-Object} & {MP-Object}\\
			\cmidrule(lr){2-3} \cmidrule(lr){4-5} \cmidrule(lr){6-7} \cmidrule(lr){8-9}
			
			ALX & 0.7841 & 0.0048 & 0.7804 & 0.7076 & 0.8192 & 0.0175 & 0.7683 & 0.7514 \\
			GNT & 0.8736 & 0.0104 & 0.8313 & 0.7875 & 0.8939 & 0.0284 & 0.8329 & 0.8418 \\
			VGG & 0.8536 & 0.0055 & 0.8258 & 0.8162 & 0.9122 & 0.0308 & 0.8151 & 0.7806 \\					
%
%
%
%
%
%
			\bottomrule
			
			\label{tbl:clutterMag}

		\end{tabular}}

\end{table}

{\bf{Is there an adversarial perturbation for a CNN that already uses foveations?} \;}  Table~\ref{tbl:clutterMag}   shows that this is the case, as the accuracy of the adversarial examples \emph{MP} is very similar to the adversarial example for the cropped object, \emph{MP-Object}.  However, our results also show that we can apply a foveation mechanism different from the foveation mechanism used to generate the adversarial perturbation, and improve again the accuracy. For example,  \emph{$1$ Shift MP-Object} and  \emph{Embedded MP-Object} substantially improve the accuracy of the adversarial perturbation calculated to affect the crop of the target object (\emph{MP-Object}).

Previous works to alleviate adversarial perturbations are based on training a CNN that learns to classify adversarial examples generated at the training phase~\citep{gu2014towards,GoodfellowSS_CoRR_2014}. Analogously to the foveations, new adversarial examples can be generated for the CNN that has been re-trained with adversarial examples. Yet, with the foveations, at testing phase a different foveation mechanism from the one used to generate the adversarial example could be exploited, as we have previously shown, but  we can not re-train the CNN at testing phase. Thus, foveations are a promising tool to design object recognition systems robust to adversarial examples.

\section{Conclusions}

Adversarial examples are a consequence of CNNs acting as a high-dimensional  linear classifier in the vicinity of the images with  objects recognized by the CNN. Also, the transformation of the image produced by a foveation decreases the effect of the adversarial perturbation to  the classification scores, because the robustness of the CNNs to transformations of objects does not generalize to the perturbations.

This suggests that a system similar to~\cite{Mnih_RNN_2014}, which integrates information from multiple fixations for object recognition, may be more robust to the phenomenon of the adversarial examples. Note that this system, which is based on integrating information from several image regions, is more similar to human vision than current CNN architectures. Yet, a remaining puzzle is the robustness of human perception to the perturbations. Some hints towards an answer could be that humans fixate their eyes on salient regions,  and the eccentricity dependent resolution of the retina help eliminate the background outside the eye fixation.

\subsubsection*{Acknowledgments}

This work was supported by the Singapore Defense Innovative Research Program 9014100596, Ministry of Education Academic Research Fund Tier 2 MOE2014-T2-1-144, and also, by the Center for Brains, Minds and Machines (CBMM), funded by NSF STC award CCF-1231216.

\bibliographystyle{iclr2016_conference}
\bibliography{tex/references}

\begin{thebibliography}{14}
\providecommand{\natexlab}[1]{#1}
\providecommand{\url}[1]{\texttt{#1}}
\expandafter\ifx\csname urlstyle\endcsname\relax
  \providecommand{\doi}[1]{doi: #1}\else
  \providecommand{\doi}{doi: \begingroup \urlstyle{rm}\Url}\fi

\bibitem[Deng et~al.(2009)Deng, Dong, Socher, Li, Li, and
  Fei-Fei]{Deng_CVPR_2009}
Deng, Jia, Dong, Wei, Socher, Richard, Li, Li-Jia, Li, Kai, and Fei-Fei, Li.
\newblock Imagenet: A large-scale hierarchical image database.
\newblock In \emph{CVPR}, 2009.

\bibitem[Fawzi et~al.(2015)Fawzi, Fawzi, and Frossard]{fawzi15}
Fawzi, Alhussein, Fawzi, Omar, and Frossard, Pascal.
\newblock Fundamental limits on adversarial robustness.
\newblock In \emph{ICML}, 2015.

\bibitem[Goodfellow et~al.(2015)Goodfellow, Shlens, and
  Szegedy]{GoodfellowSS_CoRR_2014}
Goodfellow, Ian~J., Shlens, Jonathon, and Szegedy, Christian.
\newblock Explaining and harnessing adversarial examples.
\newblock In \emph{ICLR}, 2015.

\bibitem[Gu \& Rigazio(2015)Gu and Rigazio]{gu2014towards}
Gu, Shixiang and Rigazio, Luca.
\newblock Towards deep neural network architectures robust to adversarial
  examples.
\newblock In \emph{ICLR}, 2015.

\bibitem[Huang et~al.(2015)Huang, Shen, Boix, and Zhao]{Salicon_2015}
Huang, Xun, Shen, Chengyao, Boix, Xavier, and Zhao, Qi.
\newblock {SALICON}: Reducing the semantic gap in saliency prediction by
  adapting neural networks.
\newblock In \emph{ICCV}, 2015.

\bibitem[Jia et~al.(2014)Jia, Shelhamer, Donahue, Karayev, Long, Girshick,
  Guadarrama, and Darrell]{Jia_arXiv_2014}
Jia, Yangqing, Shelhamer, Evan, Donahue, Jeff, Karayev, Sergey, Long, Jonathan,
  Girshick, Ross, Guadarrama, Sergio, and Darrell, Trevor.
\newblock Caffe: Convolutional architecture for fast feature embedding.
\newblock In \emph{MM}, 2014.

\bibitem[Krizhevsky(2009)]{Krizhevsky2009cifar}
Krizhevsky, Alex.
\newblock Learning multiple layers of features from tiny images.
\newblock \emph{Tech. Rep}, 2009.

\bibitem[Krizhevsky et~al.(2012)Krizhevsky, Sutskever, and
  Hinton]{Krizhevsky_NIPS_2012}
Krizhevsky, Alex, Sutskever, Ilya, and Hinton, Geoffrey~E.
\newblock Imagenet classification with deep convolutional neural networks.
\newblock In \emph{NIPS}, 2012.

\bibitem[LeCun et~al.(1998)LeCun, Bottou, Bengio, and
  Haffner]{Lecun_PIEEE_1998}
LeCun, Yann, Bottou, L{\'e}on, Bengio, Yoshua, and Haffner, Patrick.
\newblock Gradient-based learning applied to document recognition.
\newblock \emph{Proceedings of the IEEE}, 86\penalty0 (11):\penalty0
  2278--2324, 1998.

\bibitem[Mnih et~al.(2014)Mnih, Heess, Graves, and Kavukcuoglu]{Mnih_RNN_2014}
Mnih, Volodymyr, Heess, Nicolas, Graves, Alex, and Kavukcuoglu, Koray.
\newblock Recurrent models of visual attention.
\newblock In \emph{NIPS}, 2014.

\bibitem[Russakovsky et~al.(2015)Russakovsky, Deng, Su, Krause, Satheesh, Ma,
  Huang, Karpathy, Khosla, Bernstein, Berg, and
  Fei{-}Fei]{Russakovsky_CoRR_2014}
Russakovsky, Olga, Deng, Jia, Su, Hao, Krause, Jonathan, Satheesh, Sanjeev, Ma,
  Sean, Huang, Zhiheng, Karpathy, Andrej, Khosla, Aditya, Bernstein,
  Michael~S., Berg, Alexander~C., and Fei{-}Fei, Li.
\newblock Imagenet large scale visual recognition challenge.
\newblock \emph{IJCV}, 2015.

\bibitem[Simonyan \& Zisserman(2015)Simonyan and Zisserman]{Simonyan_CoRR_2014}
Simonyan, Karen and Zisserman, Andrew.
\newblock Very deep convolutional networks for large-scale image recognition.
\newblock In \emph{ICLR}, 2015.

\bibitem[Szegedy et~al.(2014)Szegedy, Zaremba, Sutskever, Bruna, Erhan,
  Goodfellow, and Fergus]{SzegedyZSBEGF_CoRR_2013}
Szegedy, Christian, Zaremba, Wojciech, Sutskever, Ilya, Bruna, Joan, Erhan,
  Dumitru, Goodfellow, Ian~J., and Fergus, Rob.
\newblock Intriguing properties of neural networks.
\newblock In \emph{ICLR}, 2014.

\bibitem[Szegedy et~al.(2015)Szegedy, Liu, Jia, Sermanet, Reed, Anguelov,
  Erhan, Vanhoucke, and Rabinovich]{Szegedy_CoRR_2014}
Szegedy, Christian, Liu, Wei, Jia, Yangqing, Sermanet, Pierre, Reed, Scott,
  Anguelov, Dragomir, Erhan, Dumitru, Vanhoucke, Vincent, and Rabinovich,
  Andrew.
\newblock Going deeper with convolutions.
\newblock In \emph{CVPR}, 2015.

\end{thebibliography}

\newpage

\section*{\LARGE\sc Supplementary Material}

\renewcommand{\thesection}{\Alph{section}}
\setcounter{section}{0}

\section{Results}

\subsection{The Linearity of CNNs for ImageNet}
\label{app:Results}

\begin{figure*}[t!]
	\centering
	\vspace{-2ex}
	\raisebox{0\height}{\subfloat{\includegraphics[width=0.8\textwidth]{sec1/cross_l1_legend}}}
	\vspace{-2ex}
	{\renewcommand{\arraystretch}{0.1}
		\setlength{\tabcolsep}{2pt}
		\begin{tabular}{ccc}
			\subfloat{\includegraphics[width=.305\textwidth]{sec1/alx_cross_l1_new}} &
			\subfloat{\includegraphics[width=.305\textwidth]{sec1/gnt_cross_l1_new}} &
			\subfloat{\includegraphics[width=.305\textwidth]{sec1/vgg_cross_l1_new}} \\
			\subfloat{\includegraphics[width=.305\textwidth]{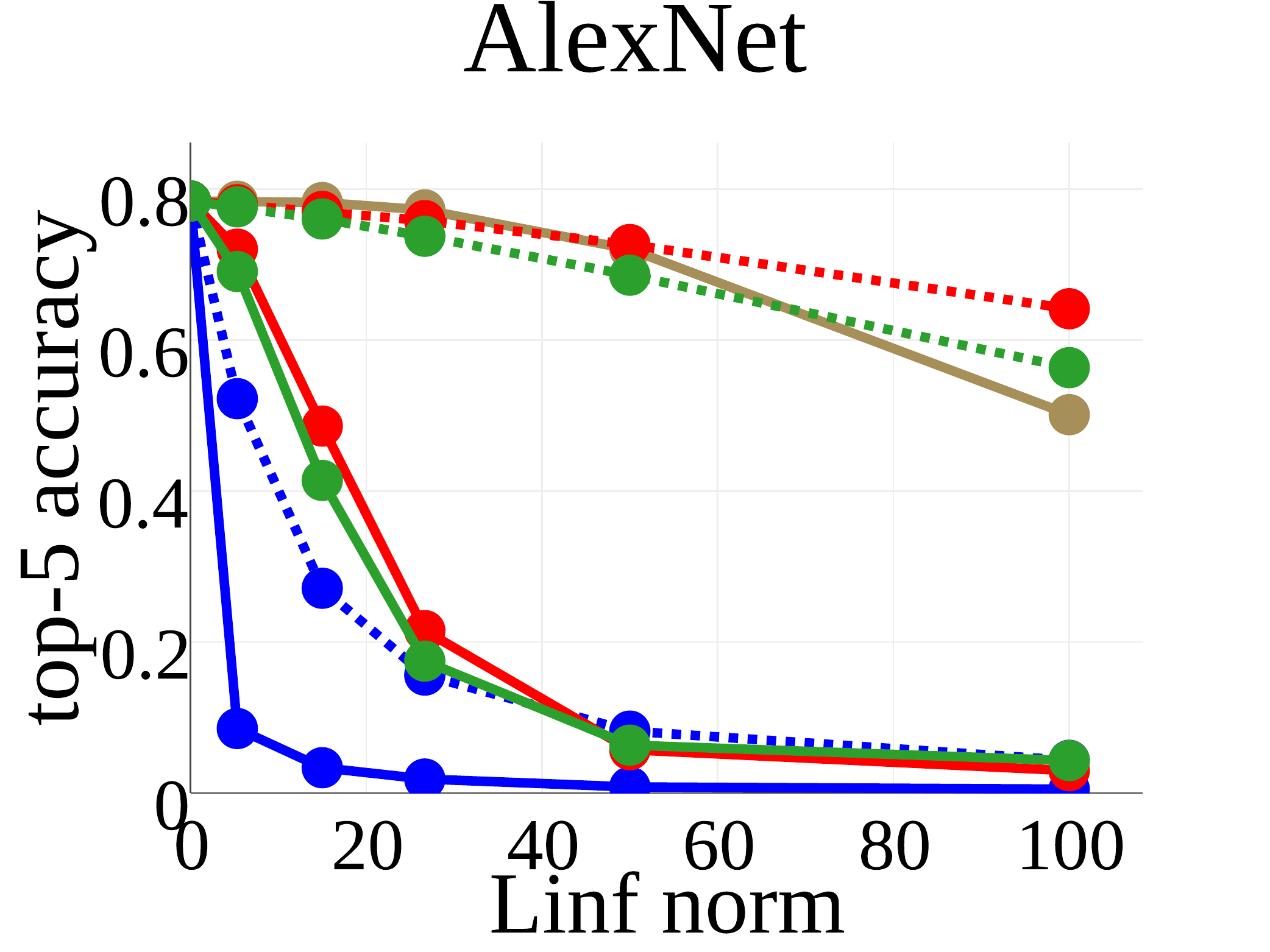}} &
			\subfloat{\includegraphics[width=.305\textwidth]{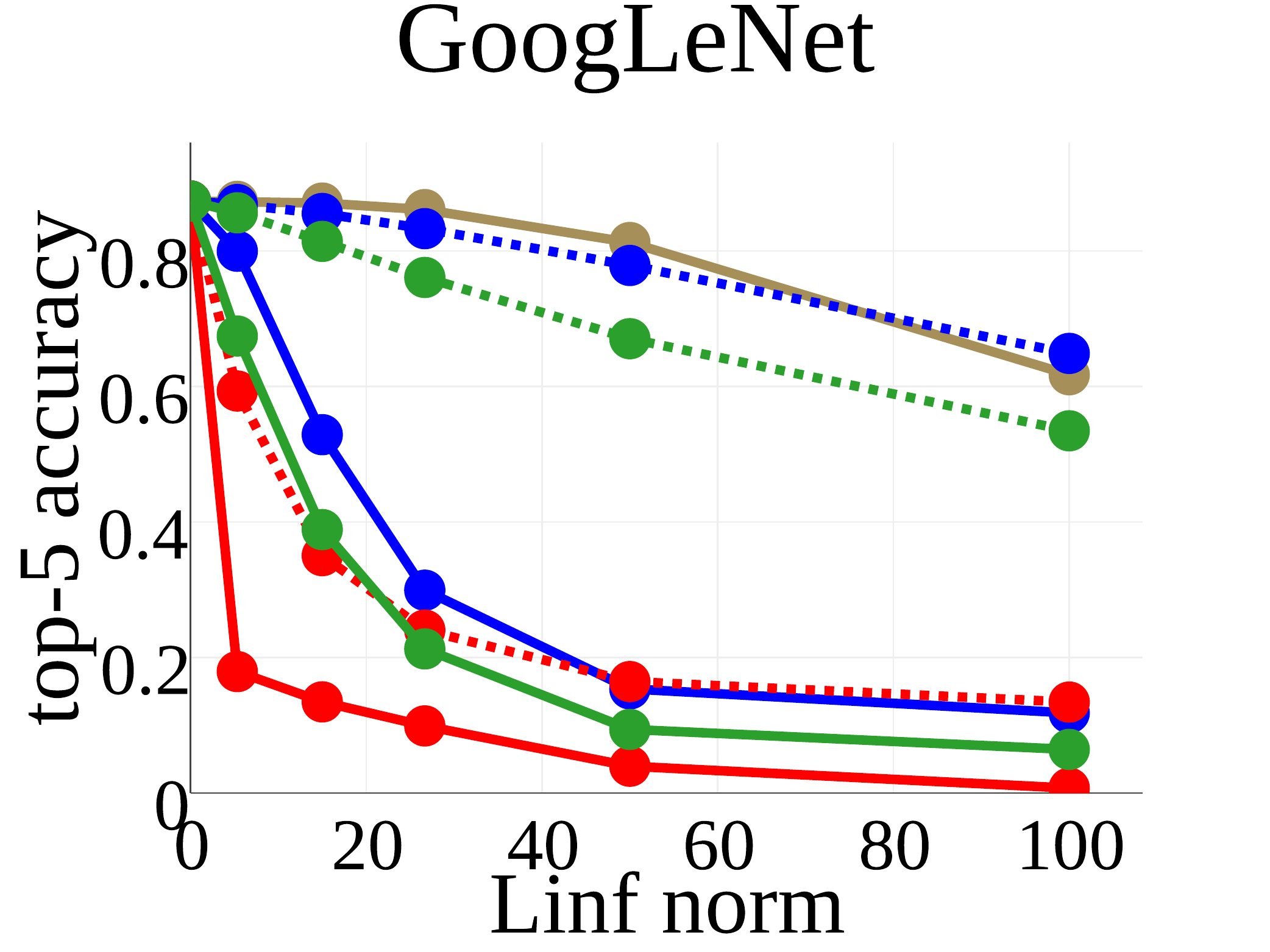}} &
			\subfloat{\includegraphics[width=.305\textwidth]{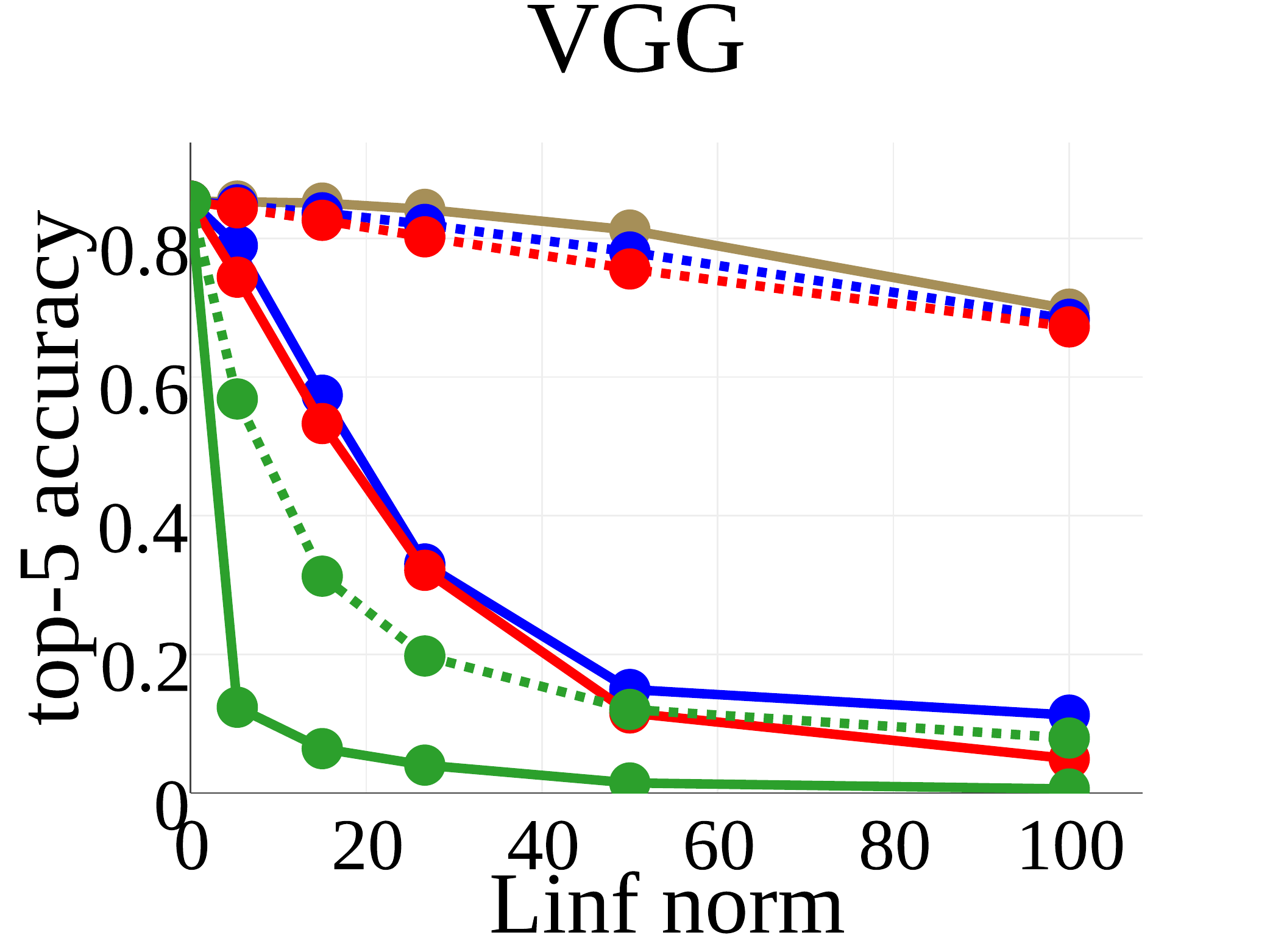}} \\
		\end{tabular}
	}
	\caption
	{		
	\textit{Accuracy when Varying the $L_\infty$ Norm of the Perturbation.} Accuracy for the three CNNs we evaluate. We denote the perturbation as $X$ $Y$, where $X$ is the network that generated the perturbation --- ALX (AlexNet), GNT (GoogLeNet), VGG --- and $Y$ indicates the \emph{BFGS} or \emph{Sign}. 
	}
	\label{fig:cross2}
\end{figure*}

{\bf Review of the Properties of Adversarial Examples in ImageNet \;}
In Fig.~\ref{fig:cross2}, we extend the results of Fig.~\ref{fig:cross} in the paper with the results for \emph{Sign perturbation} using $L_\infty$ norm. We observe the same tendency than in Fig.~\ref{fig:cross}.

\begin{figure*}[t!]
	\centering
	\vspace{-2ex}
	\raisebox{0\height}{\subfloat{\includegraphics[width=0.7\textwidth]{sec2/crop_noise_legend}}}
	{\renewcommand{\arraystretch}{0.1}%
		\setlength{\tabcolsep}{2pt}
		\begin{tabular}{ccc}
			\subfloat{\includegraphics[width=.305\textwidth]{sec2/alx_bfgs_masknoise_new_nsyn}} &
			\subfloat{\includegraphics[width=.305\textwidth]{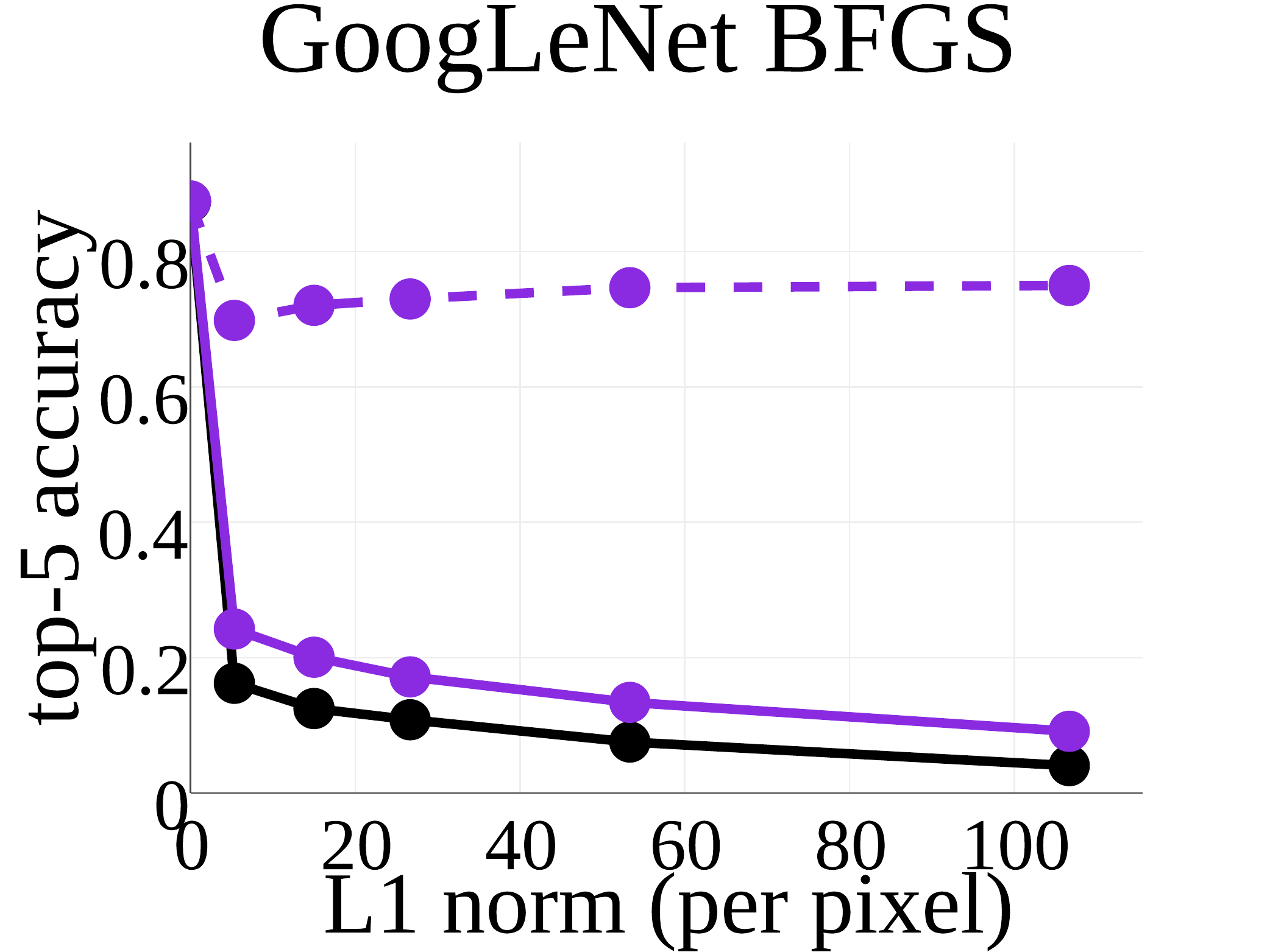}} &
			\subfloat{\includegraphics[width=.305\textwidth]{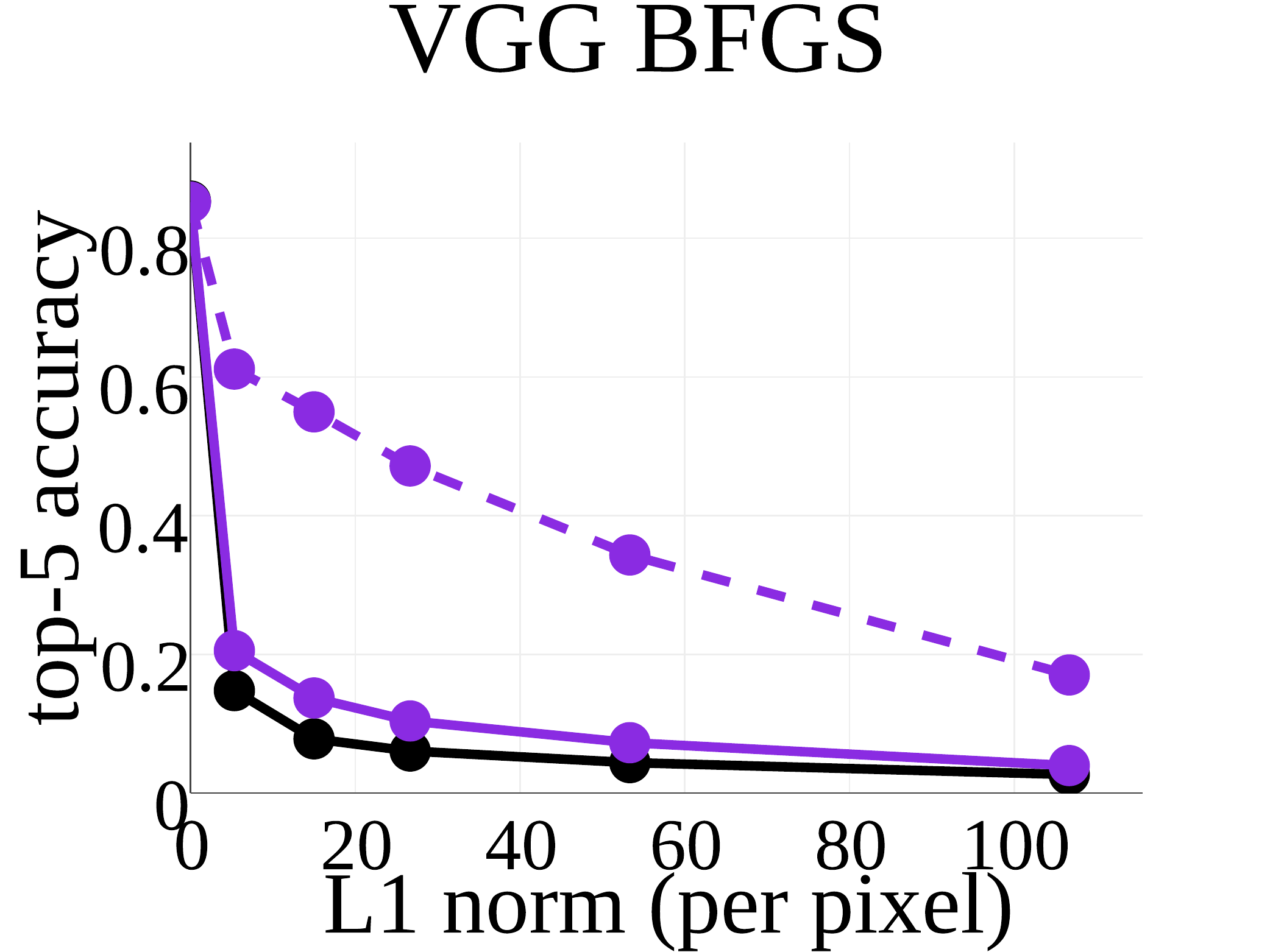}} \\
			\subfloat{\includegraphics[width=.305\textwidth]{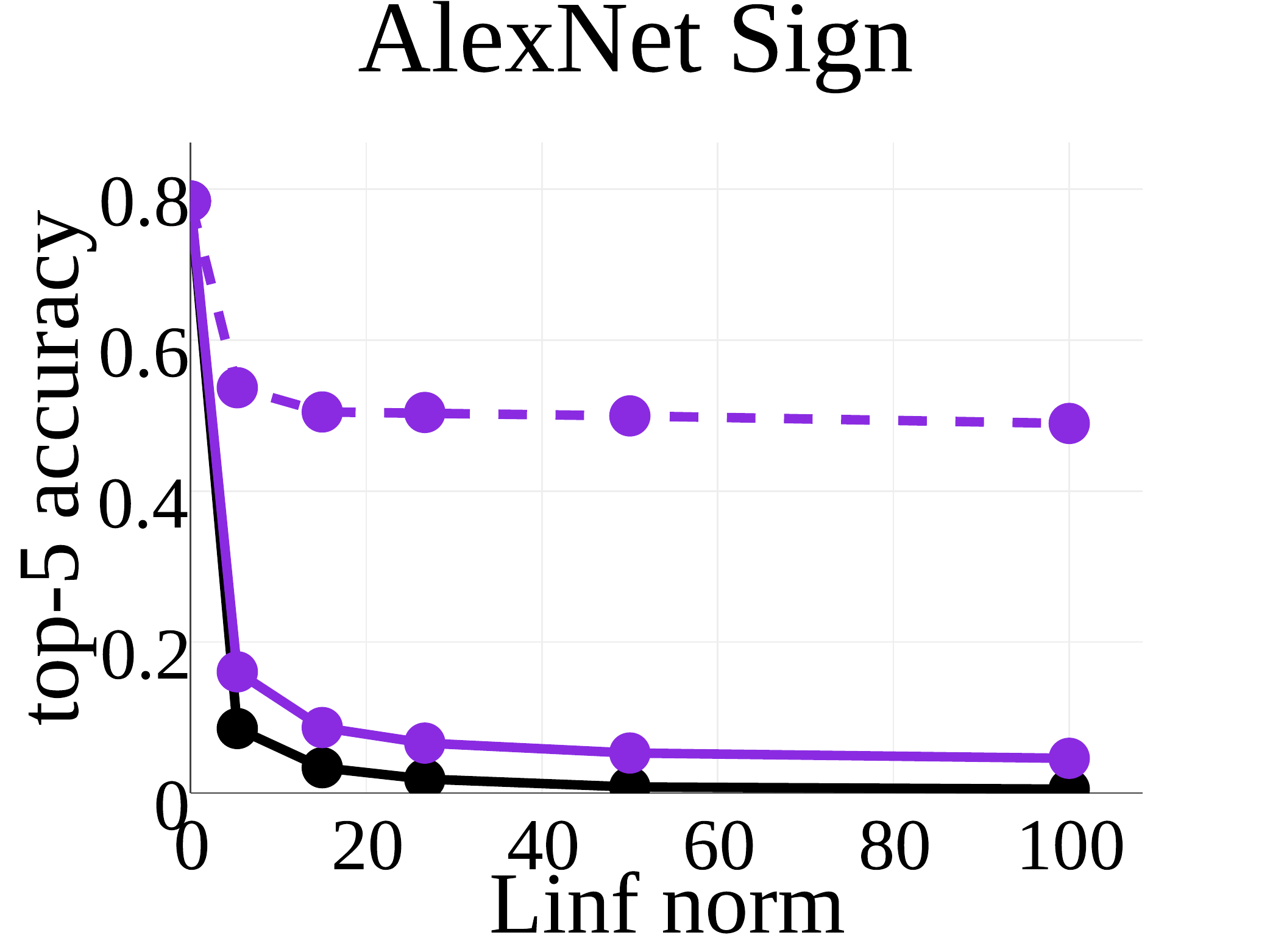}} &
			\subfloat{\includegraphics[width=.305\textwidth]{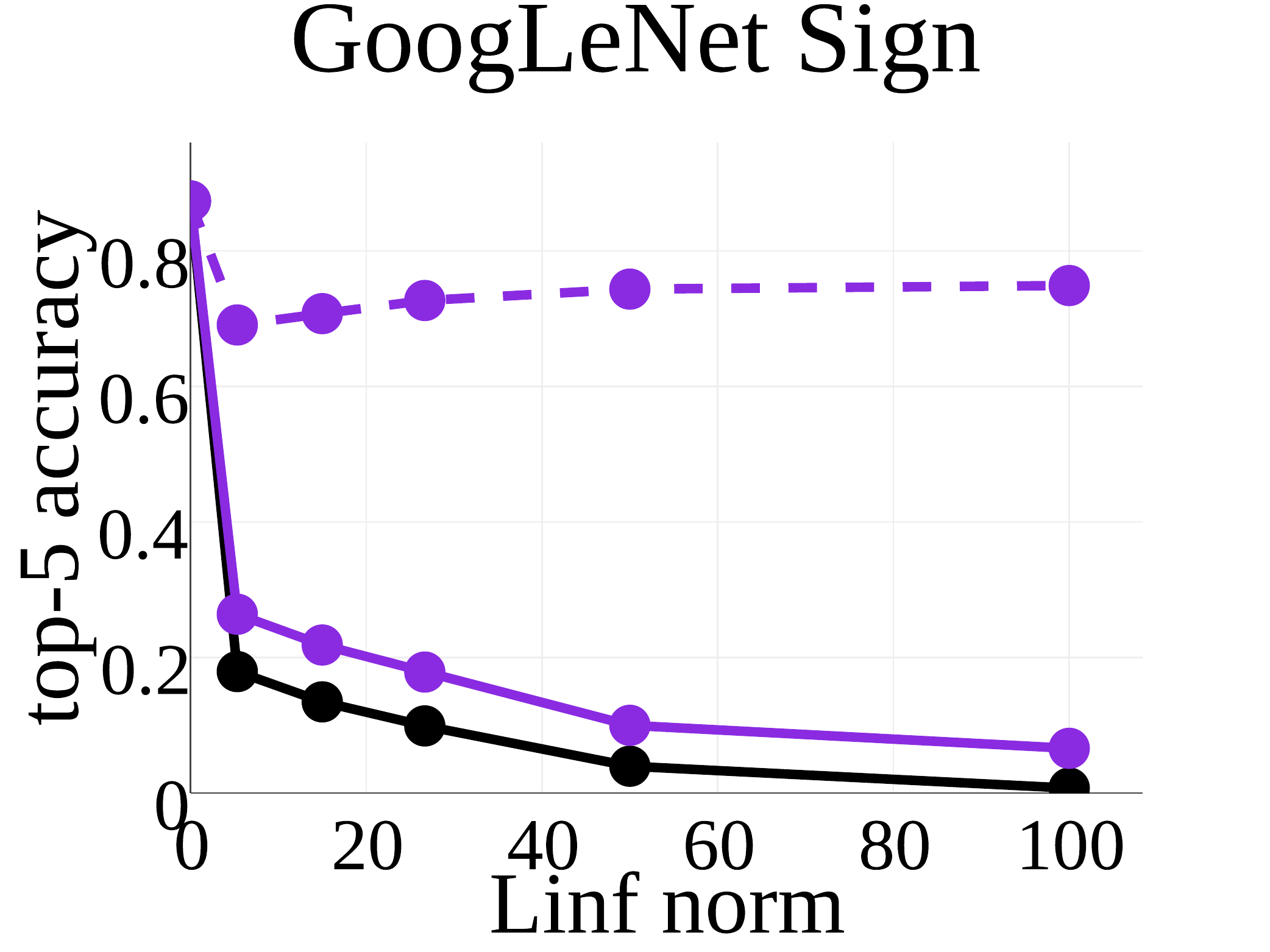}} &
			\subfloat{\includegraphics[width=.305\textwidth]{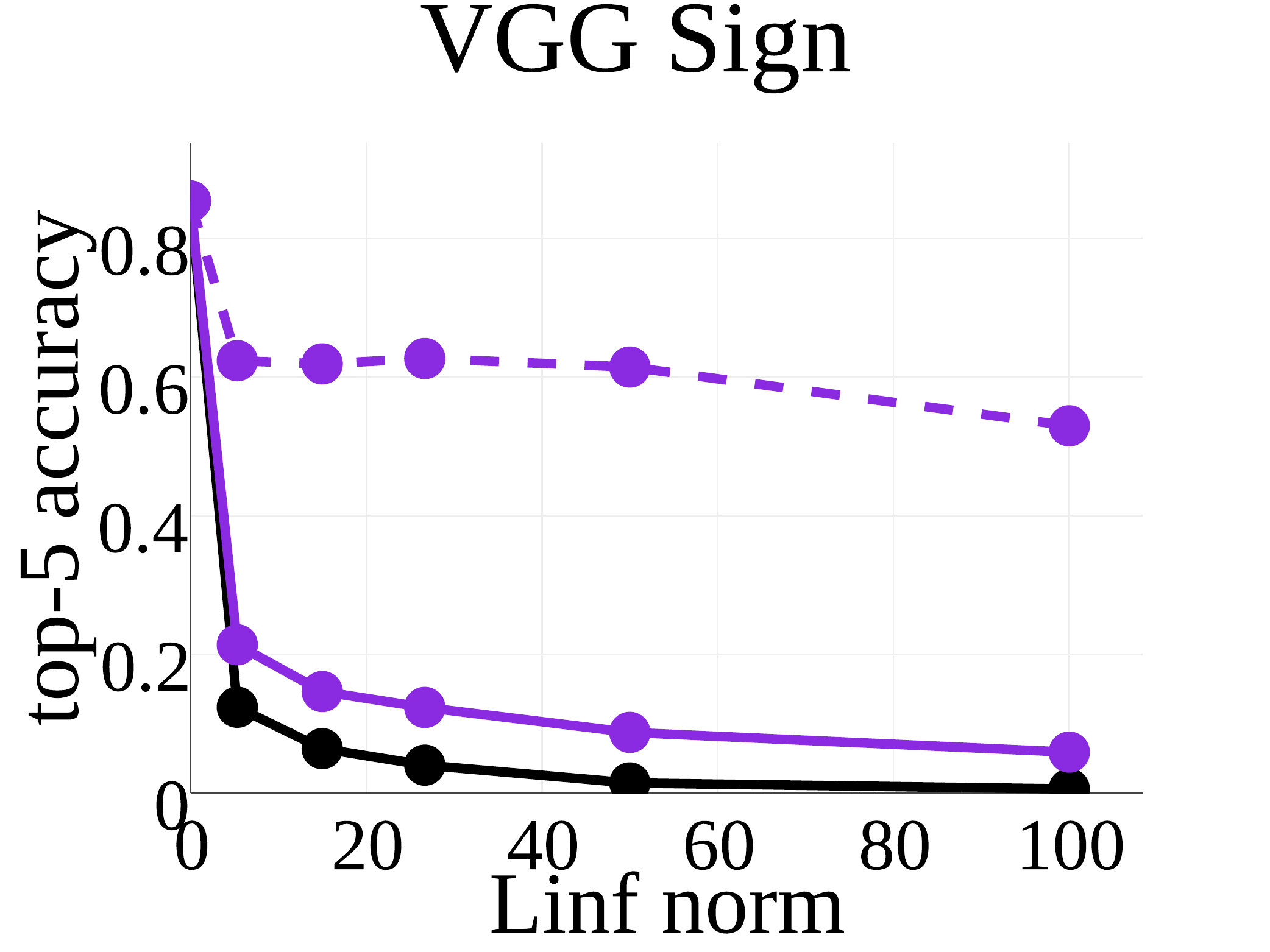}} 
		\end{tabular}
	}
	\caption
	{ 
        \textit{Accuracy of the Masked Perturbations.} 		Accuracy for the three CNNs we evaluate, when varying the norm of \emph{BFGS} (first row) and \emph{Sign} (second row).    
		}
	\label{fig:removenoise2}
\end{figure*}

{\bf Role of the Target Object Position in the Perturbation \;} 
In Fig.~\ref{fig:removenoise2}  we report the accuracy performance of the different CNN architectures for \emph{BFGS}, and \emph{Sign} using $L_\infty$ norm,  when changing the norm of the perturbation that is inside the mask, or the inverted mask, and the result for the perturbation applied in the full image (denoted as \emph{minimum perturbation (MP)}). 
We make the same observations as Fig.~\ref{fig:removenoise}a in the paper.

{\bf CNNs act Locally Linear on the Positions of a Recognized Object \;}
 In Fig.~\ref{fig:linearexample18},~\ref{fig:linearexample75},~\ref{fig:linearexample174}  we show more  examples as the one in Fig.~\ref{fig:removenoise}b in the paper, and for \emph{BFGS} and \emph{Sign} perturbations, for the images in  Fig.~\ref{fig:l1example},~\ref{fig:l1example2} and ~\ref{fig:l1example3}, in Sec.~\ref{app:Figures}, respectively. We can see that   $f(\Mat{\epsilon}^\star)$ is completely unrelated to $f(\Mat{x}+\Mat{\epsilon}^\star) - f(\Mat{x})$, and we need to review the hypothesis that the CNNs behave too linearly in order to reconcile the hypothesis with these results.

\begin{figure*}[t!]
	\centering
	\vspace{-2ex}
	\raisebox{0\height}{\subfloat{\includegraphics[width=0.15\textwidth]{sec4/linearity_legend}}}
	{\renewcommand{\arraystretch}{0.1}
		\setlength{\tabcolsep}{2pt}
		\begin{tabular}{ccc}
			\subfloat{\includegraphics[width=.305\textwidth]{sec4/alx_bfgs_linearity}} &
			\subfloat{\includegraphics[width=.305\textwidth]{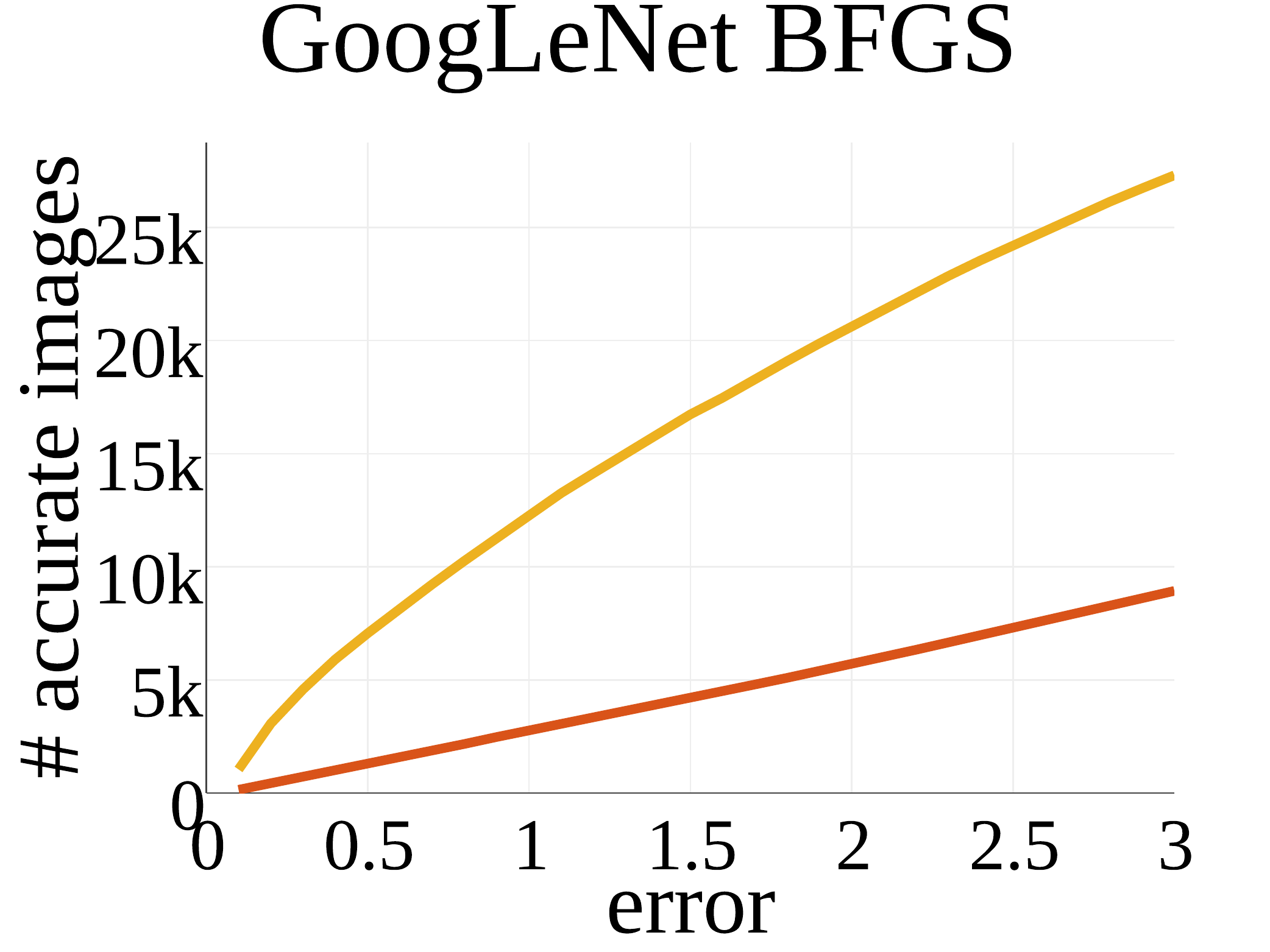}} &
			\subfloat{\includegraphics[width=.305\textwidth]{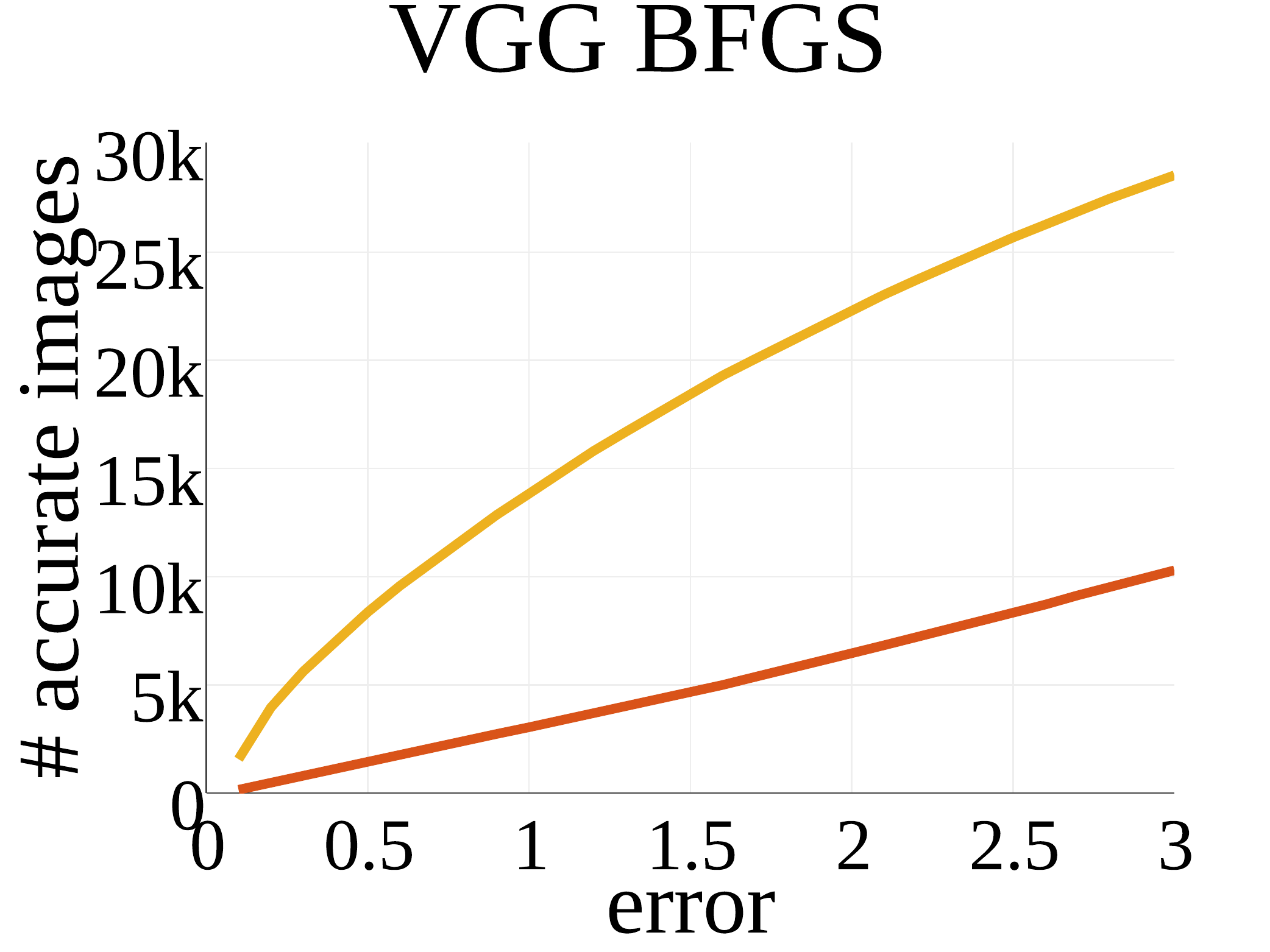}} 	\vspace{-1ex}\\
			\subfloat{\includegraphics[width=.305\textwidth]{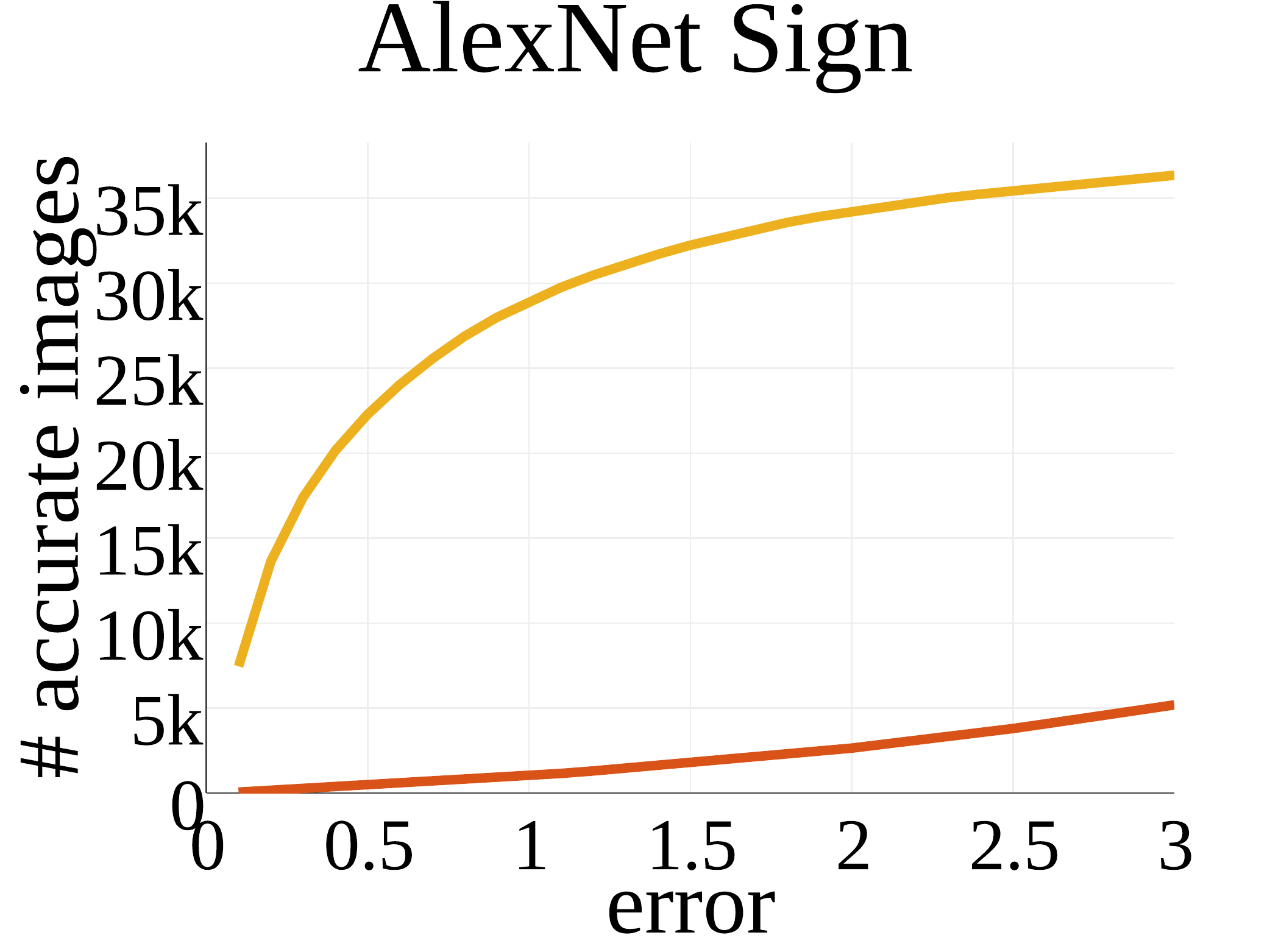}} &
			\subfloat{\includegraphics[width=.305\textwidth]{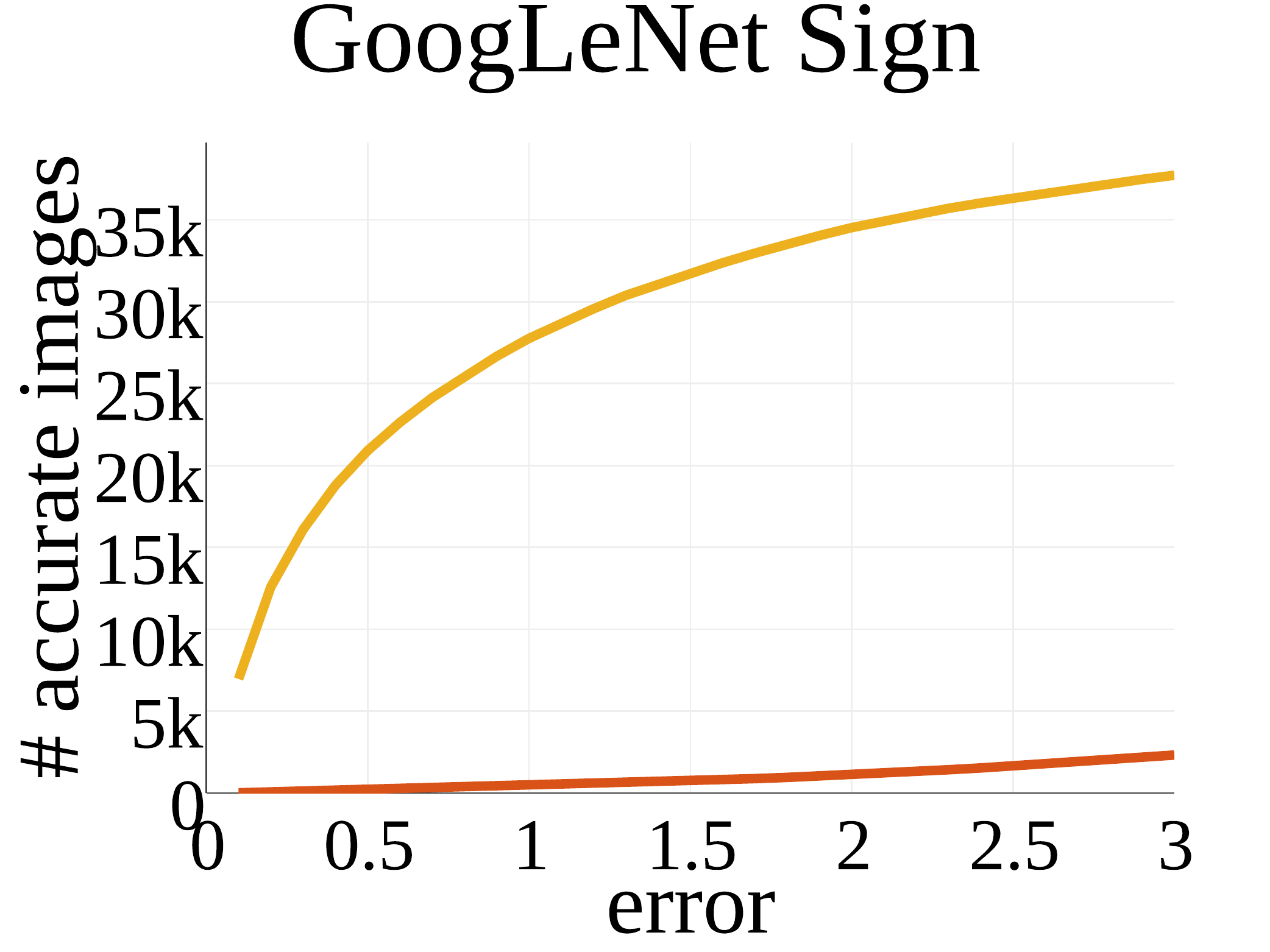}} &
			\subfloat{\includegraphics[width=.305\textwidth]{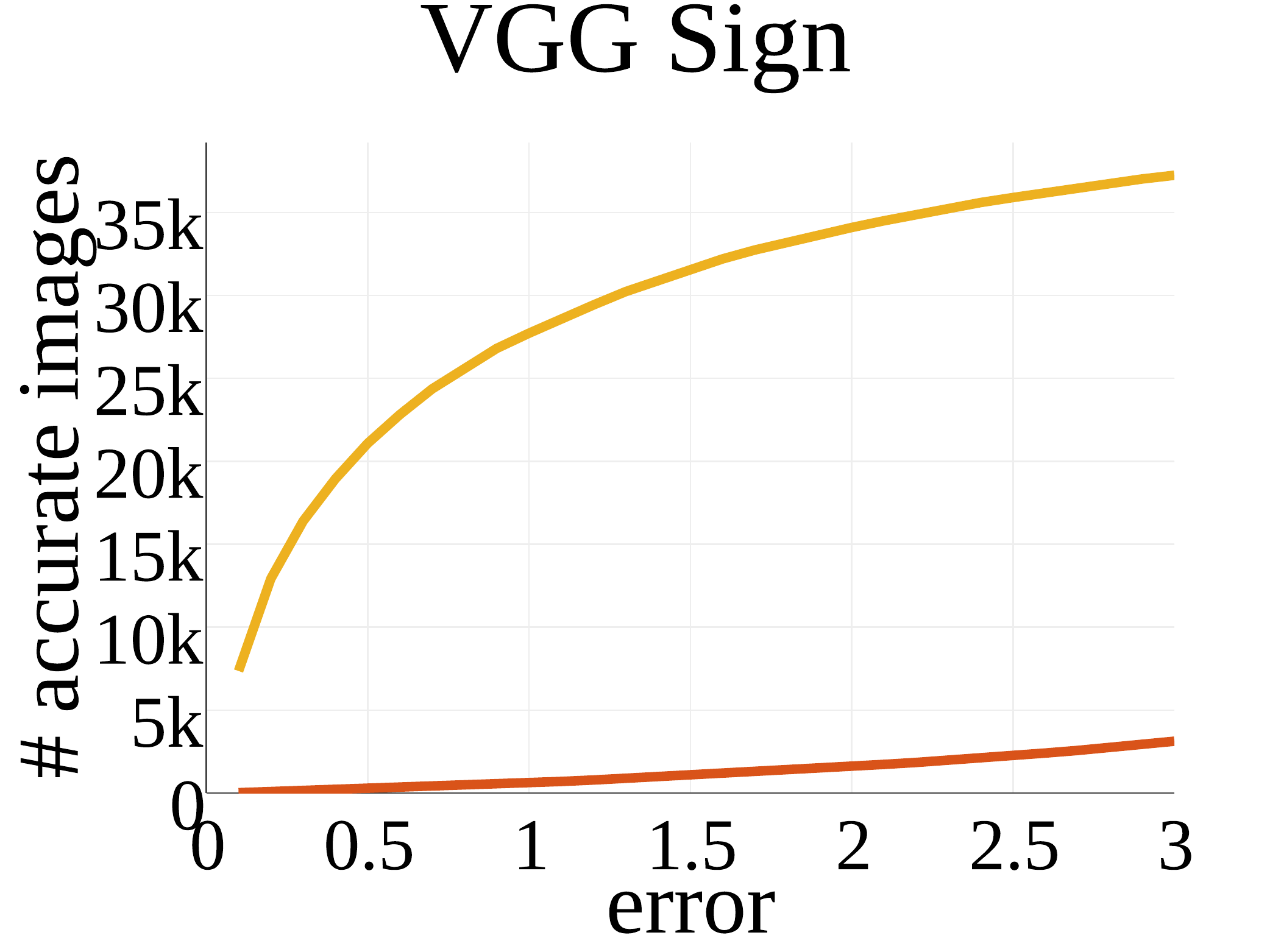}} \\
		\end{tabular}
	}
	\caption
	{
		\textit{Error of the Linearity Hypothesis.} Cumulative histogram of the number of images with an $L_1$ error smaller than the value in the horizontal axis. Only the images that are correctly classified by the CNN are included.}
	\label{fig:linearity2}
\end{figure*}

In Fig.~\ref{fig:linearity2}, we approximate the alignment of the perturbation with the hypothetical linear classifier,~\ie $\Mat{w}^\prime\Mat{\epsilon} $,  and we evaluate how much this hypothetical linear classifier deviates from the real classification score, $f(\Mat{x}+\Mat{\epsilon})-f(\Mat{x})$, for all evaluated CNNs and perturbations. As observed in Fig.~\ref{fig:removenoise}c in the paper for AlexNet using \emph{BFGS}, we can see that the error derived from our hypothesis is much smaller than from the hypothesis that the CNN is always a linear classifier.

\subsection{Foveation-based mechanisms alleviate adversarial examples}
\label{app:Fixation}

{\bf Accuracy of the adversarial examples after the foveation \;} Table~\ref{tbl:clutterMagAll}  shows the accuracy  before and after the foveation for the adversarial examples with the perturbation with minimum norm using \emph{BFGS} and \emph{Sign} perturbations. We can extract the same conclusions as in Table~\ref{tbl:clutterMag} in the paper.

{\bf{How much does removing clutter or average the result of multiple foveations improve the accuracy?} \;} 
In Table~\ref{tbl:clutter} we show the accuracy after the foveation when there is no adversarial perturbation, and when the adversarial perturbation has norm equal to $5.3$. As we have explained in the paper, we can see that the impact of removing clutter or averaging multiple foveations is much smaller than the impact of Hypothesis~\ref{hyp:foveation}.  

\begin{figure*}[t!]
	\centering
	\vspace{-2ex}
	\raisebox{0\height}{\subfloat{\includegraphics[width=0.65\textwidth]{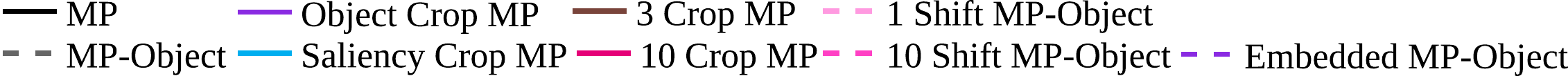}}}
	{\renewcommand{\arraystretch}{0.1}
		\setlength{\tabcolsep}{2pt}
		\begin{tabular}{ccc}
			\subfloat{\includegraphics[width=.305\textwidth]{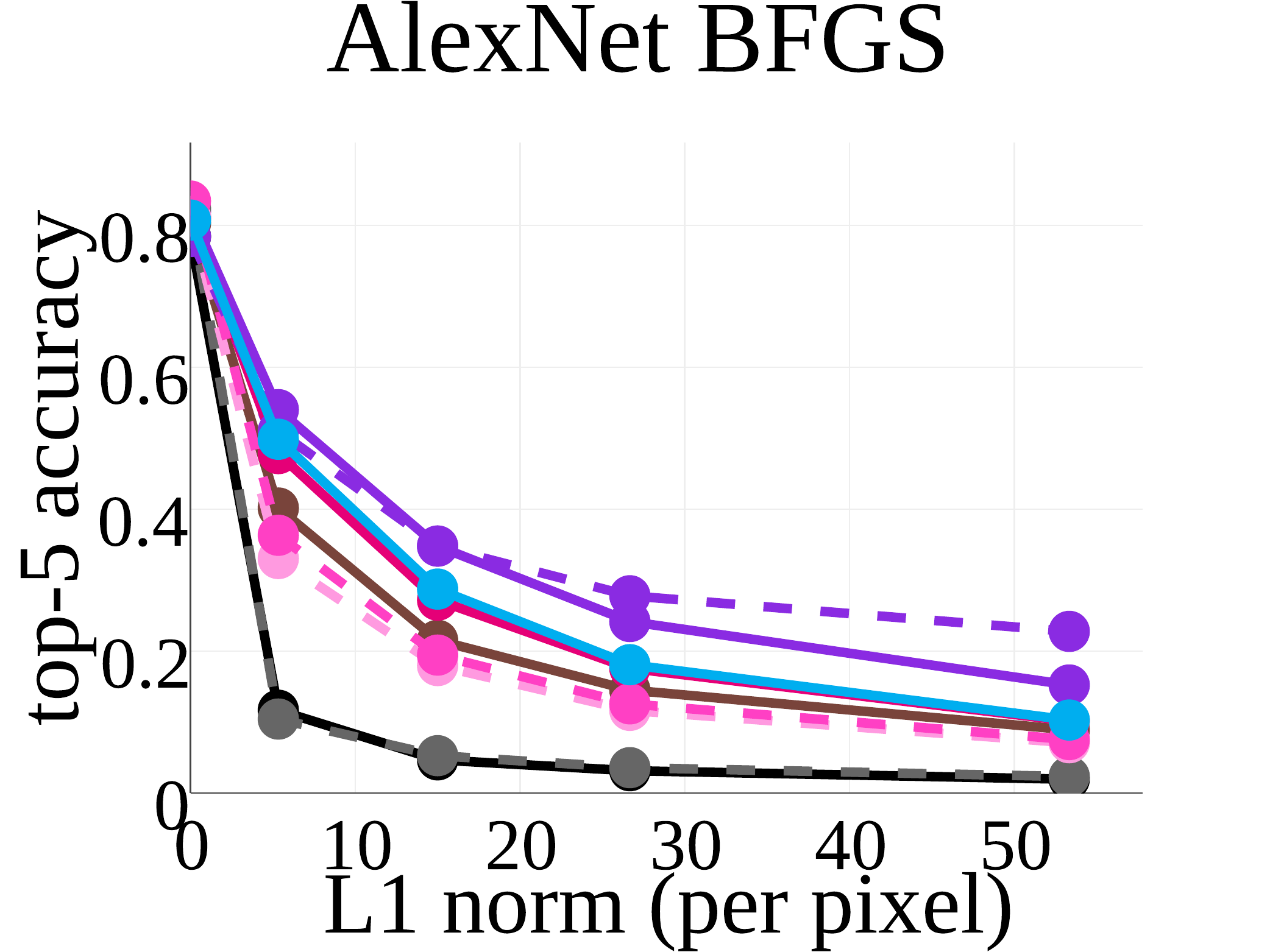}} &
			\subfloat{\includegraphics[width=.305\textwidth]{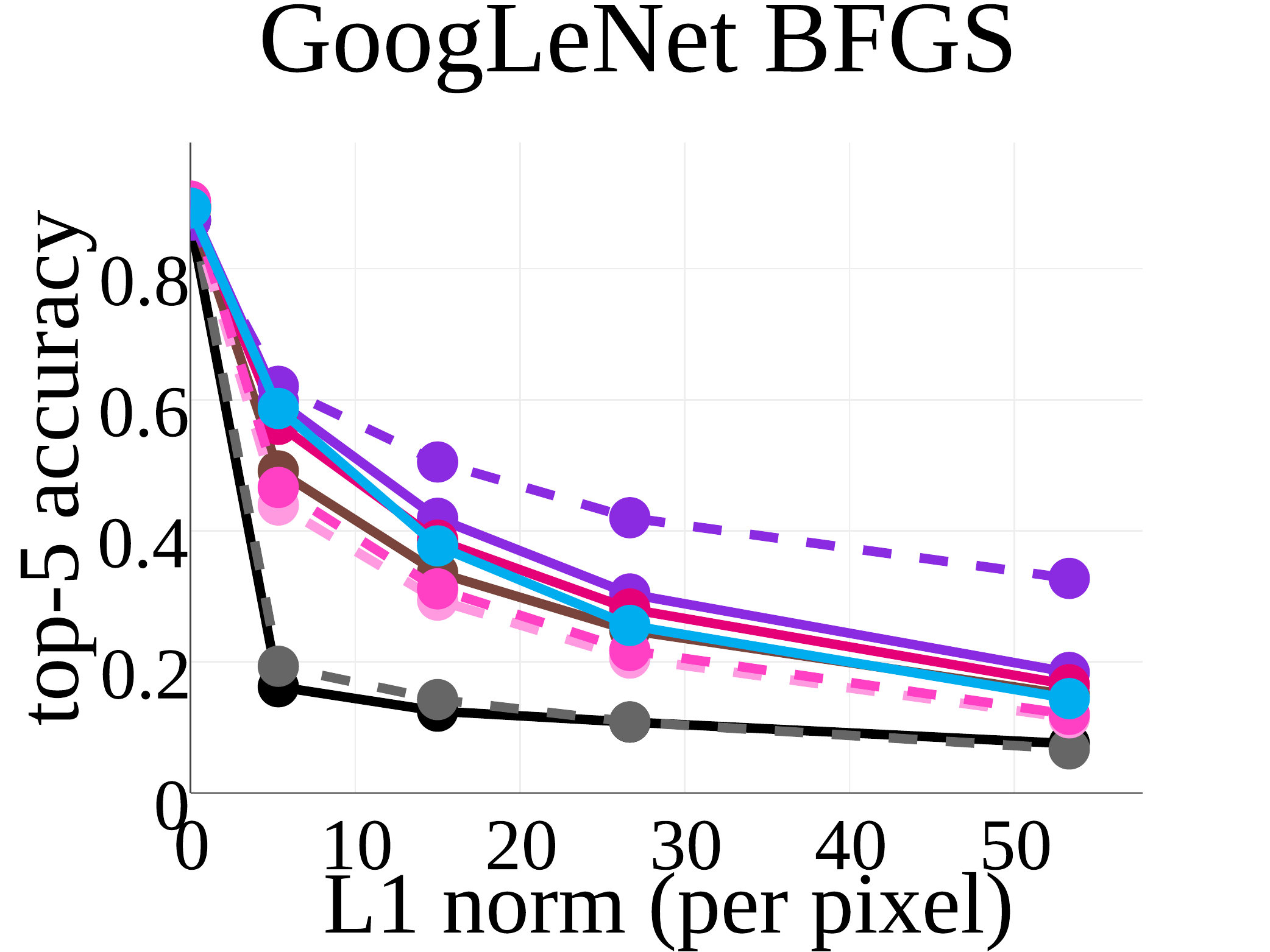}} &
			\subfloat{\includegraphics[width=.305\textwidth]{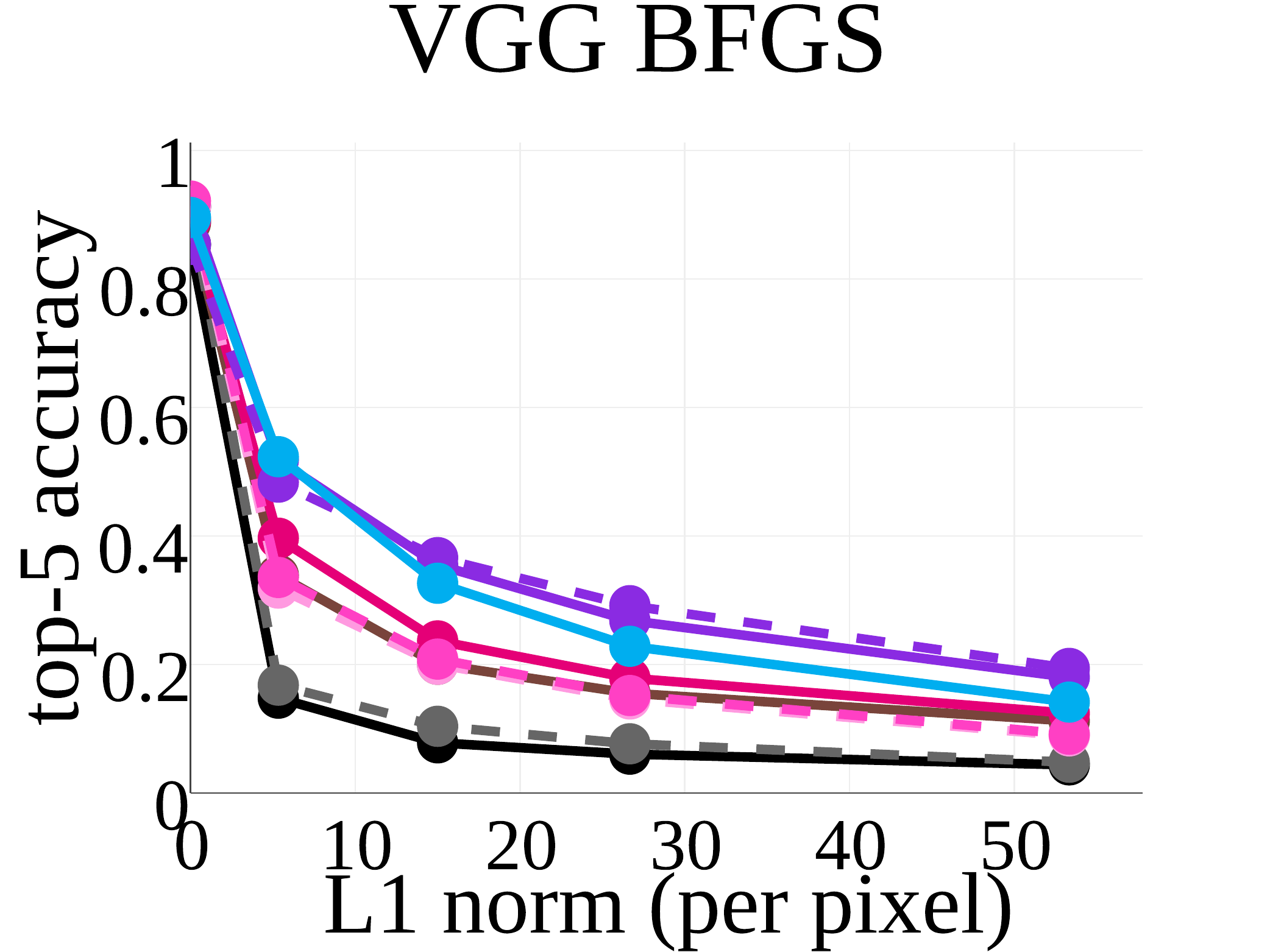}} \\
			\subfloat{\includegraphics[width=.305\textwidth]{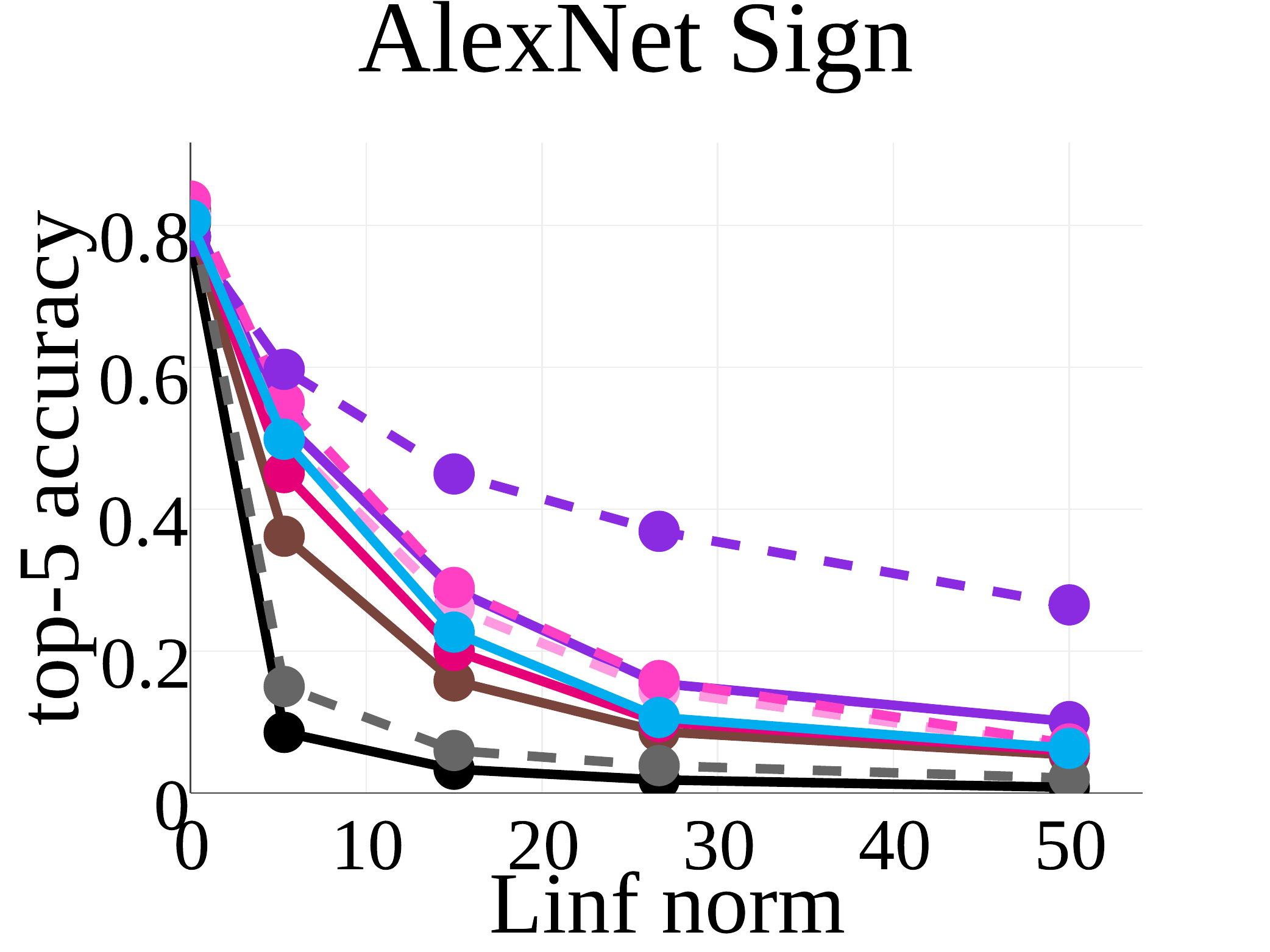}} &
			\subfloat{\includegraphics[width=.305\textwidth]{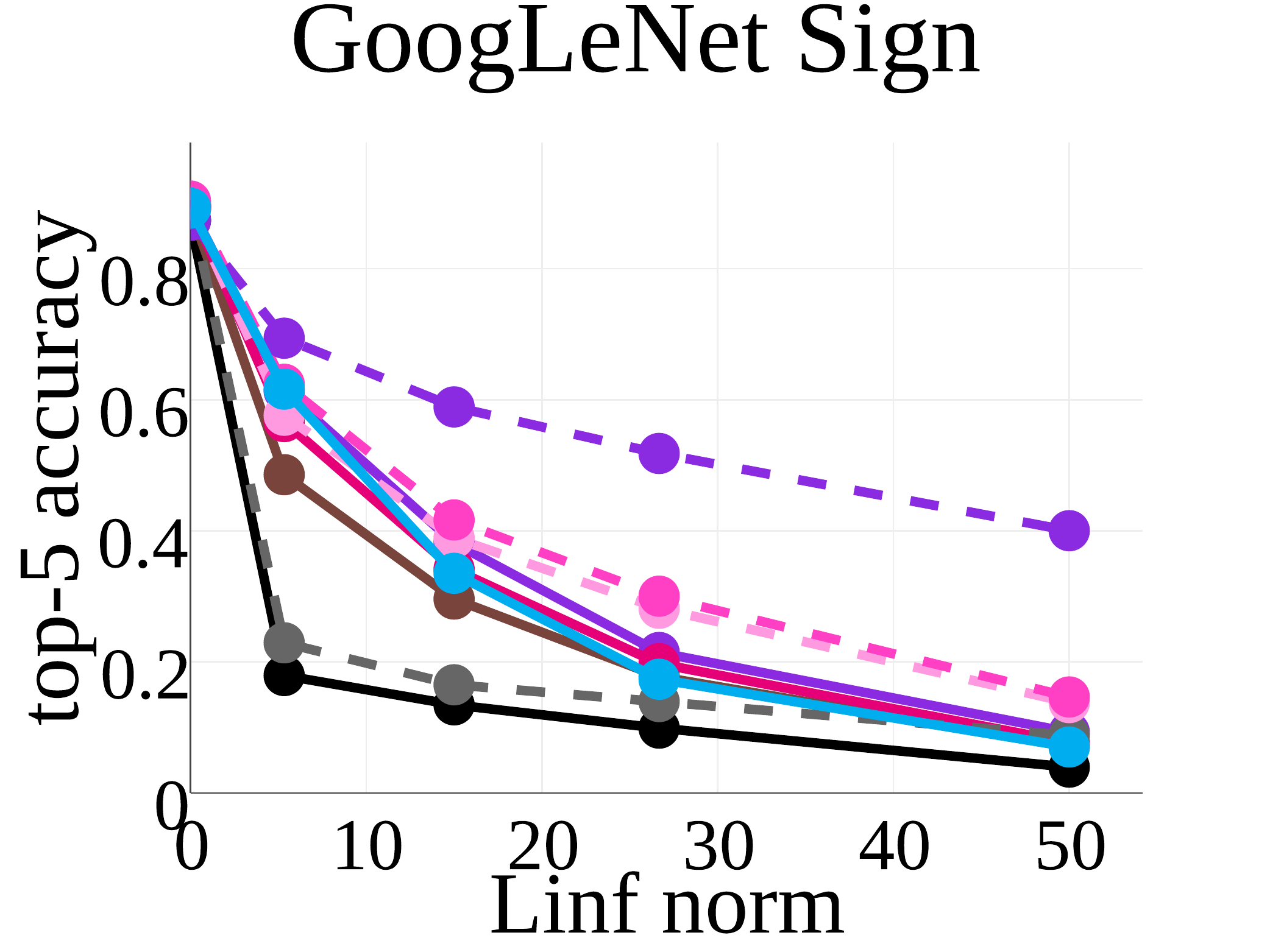}} &
			\subfloat{\includegraphics[width=.305\textwidth]{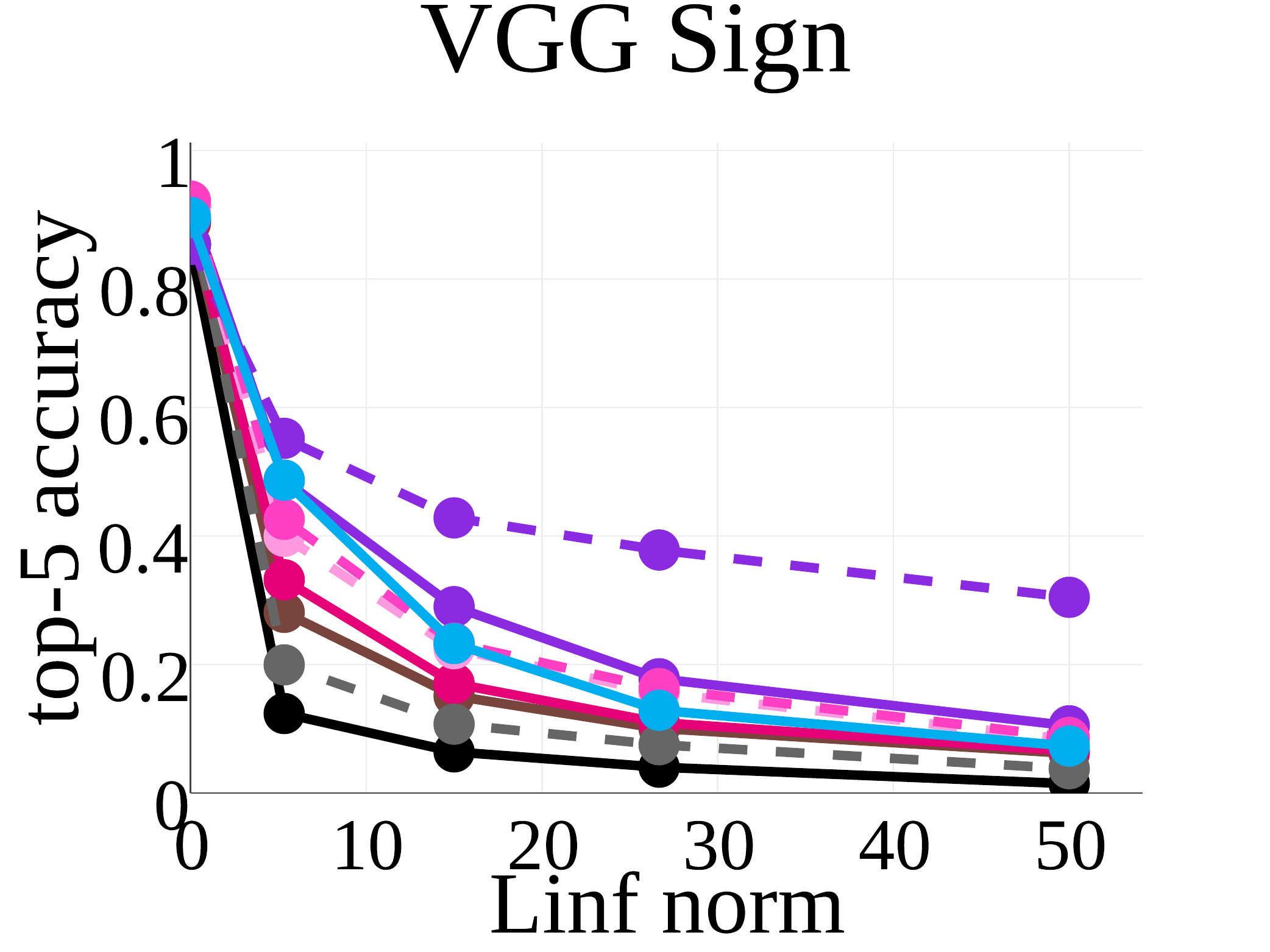}} \\
		\end{tabular}
	}
\caption{\textit{Accuracy of the Foveations.} Accuracy for the three CNNs we evaluate, when varying the $L_1$ norm of \emph{BFGS} and varying the $L_\infty$ norm of \emph{Sign}. 
}
				
	\label{fig:clutter2}
\end{figure*}

{\bf {How much does the accuracy decrease when the norm of the adversarial perturbation is increased?} \;} In Fig.~\ref{fig:clutter2} we report the accuracy using different values of the norm of the perturbation before the foveation. As explained in the paper, when we increase the value of the norm of the perturbation, the accuracy decreases because the CNN may enter to the non-linear region due to the occlusions caused by the perturbation. 

Finally, in Table~\ref{tbl:normcomparison}, we report the increase of the norm of the adversarial perturbation after the foveation, which is  about $5$ times for \emph{BFGS}, and between $5$ to $8$ times for \emph{Sign}.

\clearpage


\input{tex/objectmp_example}

\clearpage
\begin{table}[t!]

\centering
		\captionof{table}{\emph{Evaluation of the Foveation Mechanisms.} Quantitative results of the top-$5$ accuracy with minimum perturbation.}
		\vspace{-2ex}

		{\scriptsize

		\begin{tabular}{ccccccccc}
		
			\toprule
			
			& \multicolumn{8}{c}{{BFGS Minimum Perturbation }} \\
			\cmidrule(lr){2-9} & \multicolumn{2}{c}{Before Foveation} & \multicolumn{2}{c}{After Foveation} & \multicolumn{2}{c}{Before Foveation} & \multicolumn{2}{c}{After Foveation} \\
			
			\cmidrule(lr){2-3} \cmidrule(lr){4-5} \cmidrule(lr){6-7} \cmidrule(lr){8-9}

			& {w/o} & {}  & {Object Crop} & {Saliency Crop} & {w/o} & {} & {1 Shift} & {Embedded}\\
			
			& {MP} & {MP}  & {MP} & {MP} & {MP-Object} & {MP-Object} & {MP-Object} & {MP-Object}\\
			\cmidrule(lr){2-3} \cmidrule(lr){4-5} \cmidrule(lr){6-7} \cmidrule(lr){8-9}
			
			ALX & 0.7841 & 0.0048 & 0.7804 & 0.7076 & 0.8192 & 0.0175 & 0.7683 & 0.7514 \\
			GNT & 0.8736 & 0.0104 & 0.8313 & 0.7875 & 0.8939 & 0.0284 & 0.8329 & 0.8418 \\
			VGG & 0.8536 & 0.0055 & 0.8258 & 0.8162 & 0.9122 & 0.0308 & 0.8151 & 0.7806 \\					
			\toprule
			
			& \multicolumn{8}{c}{{Sign Minimum Perturbation}} \\
			\cmidrule(lr){2-9} & \multicolumn{2}{c}{Before Foveation} & \multicolumn{2}{c}{After Foveation} & \multicolumn{2}{c}{Before Foveation} & \multicolumn{2}{c}{After Foveation} \\
			
			\cmidrule(lr){2-3} \cmidrule(lr){4-5} \cmidrule(lr){6-7} \cmidrule(lr){8-9}

			& {w/o} & {}  & {Object Crop} & {Saliency Crop} & {w/o} & {} & {1 Shift} & {Embedded}\\
			
			& {MP} & {MP}  & {MP} & {MP} & {MP-Object} & {MP-Object} & {MP-Object} & {MP-Object}\\
			
			\cmidrule(lr){2-3} \cmidrule(lr){4-5} \cmidrule(lr){6-7} \cmidrule(lr){8-9}
			
			ALX & 0.7841 & 0 & 0.7210 & 0.6656 & 0.8192 & 0.0022 & 0.7066 & 0.7084 \\
			GNT & 0.8736 & 0.0006 & 0.7285 & 0.6995 & 0.8939 & 0.0051 & 0.7330 & 0.7764  \\
			VGG & 0.8536 & 0.0007 & 0.6856 & 0.7304 & 0.9122 & 0.0065 & 0.6397 & 0.6766  \\
			\bottomrule
			
			\label{tbl:clutterMagAll}

		\end{tabular}}

\end{table}

\begin{table}[t!]
\centering
		\captionof{table}{\emph{Evaluation of the Foveation Mechanisms.} Quantitative results of the top-$5$ accuracy in \fig~\ref{fig:clutter2}.}
		\vspace{-2ex}

		{\scriptsize 

		\begin{tabular}{cccccccccc}
			\toprule
			& \multicolumn{9}{c}{{No Perturbation ($L_{1}=0$)}} \\
           \cmidrule(lr){2-10} & {Before Foveation} & \multicolumn{4}{c}{After Foveation} & {Before Foveation} & \multicolumn{3}{c}{After Foveation} \\
            
           \cmidrule(lr){2-2} \cmidrule(lr){3-6} \cmidrule(lr){7-7} \cmidrule(lr){8-10}

			 & {}  & {Object Crop} & {10 Crop} & {3 Crop} & {Saliency Crop} & {} & {10 Shift} & {1 Shift} & {Embedded}\\
			
			 & {MP}  & {MP} & {MP} & {MP} & {MP} & {MP-Object} & {MP-Object} & {MP-Object} & {MP-Object}\\
			 
            \cmidrule(lr){2-2} \cmidrule(lr){3-6} \cmidrule(lr){7-7} \cmidrule(lr){8-10}

			ALX & 0.7841 & 0.8192 & 0.8111 & 0.8026 & 0.8073 & 0.8192 & 0.8341 & 0.8123 & 0.7841 \\
			GNT & 0.8736 & 0.8939 & 0.8951 & 0.8903 & 0.8922 & 0.8939 & 0.9030 & 0.8910 & 0.8736 \\
			VGG & 0.8536 & 0.9122 & 0.8912 & 0.8872 & 0.8957 & 0.9122 & 0.9212 & 0.9132 & 0.8536 \\					
			\toprule
		
			& \multicolumn{9}{c}{{BFGS Perturbation ($L_{1}=5.3$)}} \\
			\cmidrule(lr){2-10} & {Before Foveation} & \multicolumn{4}{c}{After Foveation} & {Before Foveation} & \multicolumn{3}{c}{After Foveation} \\
			
			\cmidrule(lr){2-2} \cmidrule(lr){3-6} \cmidrule(lr){7-7} \cmidrule(lr){8-10}
			
			 & {}  & {Object Crop} & {10 Crop} & {3 Crop} & {Saliency Crop} & {} & {10 Shift} & {1 Shift} & {Embedded}\\
			
			 & {MP}  & {MP} & {MP} & {MP} & {MP} & {MP-Object} & {MP-Object} & {MP-Object} & {MP-Object}\\

            \cmidrule(lr){2-2} \cmidrule(lr){3-6} \cmidrule(lr){7-7} \cmidrule(lr){8-10}
			 
			ALX & 0.1166 & 0.5401 & 0.4782 & 0.4016 & 0.4985 & 0.1043 & 0.3632 & 0.3304 & 0.5100 \\
			GNT & 0.1622 & 0.5972 & 0.5623 & 0.4913 & 0.5866 & 0.1934 & 0.4662 & 0.4394 & 0.6204 \\
			VGG & 0.1477 & 0.5180 & 0.3966 & 0.3391 & 0.5234 & 0.1679 & 0.3355 & 0.3190 & 0.4838 \\
			
			\toprule
			& \multicolumn{9}{c}{{Sign Perturbation ($L_{\infty}=5.3$)}} \\
			\cmidrule(lr){2-10} & {Before Foveation} & \multicolumn{4}{c}{After Foveation} & {Before Foveation} & \multicolumn{3}{c}{After Foveation} \\
			\cmidrule(lr){2-2} \cmidrule(lr){3-6} \cmidrule(lr){7-7} \cmidrule(lr){8-10}
			
			 & {}  & {Object Crop} & {10 Crop} & {3 Crop} & {Saliency Crop} & {} & {10 Shift} & {1 Shift} & {Embedded}\\
			
			 & {MP}  & {MP} & {MP} & {MP} & {MP} & {MP-Object} & {MP-Object} & {MP-Object} & {MP-Object}\\

            \cmidrule(lr){2-2} \cmidrule(lr){3-6} \cmidrule(lr){7-7} \cmidrule(lr){8-10}
			
			ALX & 0.0855 & 0.5225 & 0.4515 & 0.3619 & 0.4986 & 0.1501 & 0.5506 & 0.4989 & 0.5970 \\
			GNT & 0.1793 & 0.6133 & 0.5666 & 0.4856 & 0.6161 & 0.2292 & 0.6235 & 0.5763 & 0.6938 \\
			VGG & 0.1238 & 0.4867 & 0.3319 & 0.2812 & 0.4869 & 0.1995 & 0.4259 & 0.3995 & 0.5520 \\
			\bottomrule
			
			\label{tbl:clutter}

        \end{tabular}}

\end{table}

\begin{table}
	\centering
	\caption{\emph{Evaluation of the Foveation Mechanisms.} Increase factor of the norm of the perturbation to produce misclassification after the foveation mechanisms. Only the images that are correctly classified before and after the foveation are included.}
	{\scriptsize 
	\begin{tabular}{ccccccc}
		\toprule
		& \multicolumn{2}{c}{{AlexNet}} & \multicolumn{2}{c}{{GoogLeNet}} & \multicolumn{2}{c}{{VGG}} \\
		\cmidrule(lr){2-3}\cmidrule(lr){4-5}\cmidrule(lr){6-7} ratio & Sign & BFGS & Sign & BFGS & Sign & BFGS \\
		\midrule
		Object Crop MP / MP & 6.2484 & 14.4476 & 6.3432 & 25.8569 & 5.0216 & 17.1506 \\
		10 Shift MP-Object / MP-Object & 4.9085 & 9.1912 & 5.6342 & 19.4249 & 2.9630 & 8.5588 \\
		1 Shift MP-Object / MP-Object & 4.5336 & 8.4306 & 5.1256 & 18.1737 & 2.9676 & 8.8038 \\
		\bottomrule
		\label{tbl:normcomparison}
	\end{tabular}}
\end{table}

\clearpage


\section{Generation of the Adversarial Perturbation}
\label{app:NoiseGeneration}

Before introducing the generation of the perturbation, we introduce a more specific mathematical notation, that later will serve to clarify the details about the perturbation generation.
Recall that $\Mat{x} \in \mathbb{R}^{N}$ is an image of size $N$ pixels. This image contains an object whose object category is of class label $\ell \in \{1\ldots k\}$, where $k$ is the number of object categories (\eg~$k=1000$ in ImageNet). $f:\mathbb{R}^{N} \rightarrow \mathbb{R}^k$ is the mapping  from the input image to the classification scores returned by the CNN. Note that $f(\Mat{x})\in \mathbb{R}^k$ is a vector that contains the classification scores for all object categories. We use $e(f(\Mat{x}),\ell)$ to denote a function to evaluate the error of the classification scores, where $\ell$ is the label of the ground truth object category. In ImageNet, this error function is based on the top-$5$ scores~\citep{Russakovsky_CoRR_2014}. Thus, $e(f(\Mat{x}),l)$ returns $0$ when $\ell$ corresponds to one of the five highest scores in $f(\Mat{x})$, otherwise $e(f(\Mat{x}),l)$ returns $1$.

We use $\Mat{\epsilon}\in \mathbb{R}^N$ to denote the perturbation image that produce a misclassification when added to the input image,~\ie~$e(f(\Mat{x}+\Mat{\epsilon}),l)=1$. The image $\Mat{x}+\Mat{\epsilon}$ is the \emph{adversarial example}.
The set of all perturbation images that produce a misclassification of an image can be grouped together.
We use $ \mathcal{E}_{\Mat{x}} \subset \mathbb{R}^N$ to denote such set, $\mathcal{E}_{\Mat{x}}  = \left\{ \Mat{\epsilon} \in \mathbb{R}^N |\;e(f(\Mat{x}+\Mat{\epsilon}),l)=1 \right\}$. Then, we define $\Mat{\epsilon}^\star$ as the perturbation with minimal norm to produce misclassification of the image,~\ie~$\Mat{\epsilon}^\star = \arg\min_{\Mat{\epsilon}\in \mathcal{E}_{\Mat{x}}} \|\Mat{\epsilon}\|$. Observe that the optimal perturbation depends on the norm we choose to minimize. We will analyze the $L_1$ and $L_\infty$ norms, since the perturbations we analyze optimize one of these two norms.

{\bf BFGS perturbation.}
As in~\citep{SzegedyZSBEGF_CoRR_2013}, we approximate the perturbation $\Mat{\epsilon}^\star$ by using a box-constrained L-BFGS \footnote{\url{http://www.cs.ubc.ca/~schmidtm/Software/minFunc.html}}. 
It consists on minimizing the $L_1$ norm of the perturbation, $\|\Mat{\epsilon}\|_1$, that produces a misclassification of the image using the top-$5$ accuracy criteria,~\ie~$e(f(\Mat{x}+\Mat{\epsilon}),\ell)=1$. To do so, we minimize the $L_1$ norm plus a loss function, which we denote as $\text{loss}(\Mat{\epsilon},\{\Mat{x},\ell\})$. Let $\eta$ be a constant that weights the $L_1$ norm with respect to the loss function.

Since the minimization should produce a misclassification, the loss function is based on the accuracy. Thus, the loss could be equal to $(1-e(f(\Mat{x}+\Mat{\epsilon}),\ell))$. Since directly minimizing the top-$5$ accuracy is difficult, because the derivative of  
 $e(f(\Mat{x}+\Mat{\epsilon})$ is $0$ for any $\epsilon$ except in the boundary of producing a misclassification, we use a hinge loss. Thus, $\text{loss}(\Mat{\epsilon},\{\Mat{x},\ell\})$ is equal to $0$ when the image is misclassified (which corresponds to the final objective), otherwise the loss function takes the value of the classification score of the class that we aim to misclassify, which can be expressed as $I(\ell)^T f(\Mat{x}+\Mat{\epsilon})$ where $I(\ell)\in\mathbb{R}^k$ is an indicator vector that is $1$ in the entry corresponding to the class $\ell$ and $0$ otherwise.

The minimization of $\eta\|\Mat{\epsilon}\|_1+\text{loss}(\Mat{\epsilon},\{\Mat{x},\ell\})$ is done with L-BFGS. This uses the gradient of the loss with respect to $\Mat{\epsilon}$, which can be computed using the back-propagation routines used during training of the CNN. 
In order to further minimize the norm returned by L-BFGS, we do a line search of the norm given the perturbation by L-BFGS,~\ie~$\tilde{\epsilon} = \alpha \epsilon / \|\epsilon\|$, where $\alpha$ is a scalar factor. 

In the experiments, we set $\eta=10^{-6}$ because using this constant L-BFGS could find a perturbation that produces a misclassification in the majority of the images, except for approximately the $0.5\%$ of the images. In all the CNNs tested, these images had a classification score higher than $0.9$. To obtain a perturbation that produces a misclassification in these images, we apply the line-search method with the perturbation returned after stopping L-BFGS after one iteration.

{\bf Sign perturbation.}
It was introduced in~\cite{GoodfellowSS_CoRR_2014}.  The perturbation of the \emph{Sign perturbation} is generated using $ \text{sign}( \nabla \text{loss}(\Mat{\epsilon},\{\Mat{x},\ell\})  )$, which can be computed using back-propagation. Then, we use the line-search method to minimize the norm of the perturbation to misclassify all images.

\clearpage


\section{Perceptibility of the Adversarial Perturbation}
\label{app:Figures}

In the following pages we show examples of the perceptibility of the perturbations of the adversarial examples.

In Fig.~\ref{fig:l1example},~\ref{fig:l1example2} and ~\ref{fig:l1example3}, we show examples of the same adversarial perturbation varying the $L_1$ norm per pixel, and  the $L_\infty$ norm, in Fig.~\ref{fig:linfexample},~\ref{fig:linfexample2},~\ref{fig:linfexample3}. We can see that from a certain factor, the perturbation becomes clearly perceptible, and it occludes  parts of the target object. This helps us approximately determine at what point the adversarial perturbations become perceptible. For \emph{BFGS},  we can say that when the $L_1$ norm per pixel is higher than $15$ the perturbation becomes visible, and for $L_{\infty}$ the threshold is $100$. This difference is because the density of \emph{BFGS} is not the same through all image, it is mainly in the same position of the object, as shown in Fig.~\ref{fig:examples}. For \emph{Sign}, the threshold for both norms to make the perturbation visible is about $15$, and it is the same for both norms  because this perturbation is spread evenly through all the image. We use these values for the analysis.

\newpage
\input{tex/l1example}
\input{tex/linfexample}

\input{tex/l1example2}
\clearpage

\input{tex/linfexample2}

\input{tex/l1example3}
\input{tex/linfexample3}

\end{document}